\documentclass[sigconf]{acmart}
\AtBeginDocument{%
  }

\copyrightyear{2025}
\acmYear{2025}
\setcopyright{none}
\setcctype{by}
\acmConference[MM '25]{Proceedings of the 33rd ACM International Conference on Multimedia}{October 27--31, 2025}{Dublin, Ireland}
\acmBooktitle{Proceedings of the 33rd ACM International Conference on Multimedia (MM '25), October 27--31, 2025, Dublin, Ireland}\acmDOI{10.1145/3746027.3754779}
\acmISBN{979-8-4007-2035-2/2025/10}

\usepackage{textcmds}
\usepackage{adjustbox}
\usepackage{multirow}
\usepackage{algorithm}
\usepackage{algpseudocode} 
\usepackage{booktabs} % To thicken table lines
\usepackage{siunitx}
\usepackage{float}
\usepackage{subcaption}
\usepackage{tikz} % Add this in your preamble
\usepackage{cuted}

\usepackage{balance}
\usepackage{colortbl}

\begin{document}

%%
%% The "title" command has an optional parameter,
%% allowing the author to define a "short title" to be used in page headers.
\title{Look Beyond: Two-Stage Scene View Generation via Panorama and Video Diffusion}

%%
%% The "author" command and its associated commands are used to define
%% the authors and their affiliations.
%% Of note is the shared affiliation of the first two authors, and the
%% "authornote" and "authornotemark" commands
%% used to denote shared contribution to the research.
\author{Xueyang Kang}
\authornote{Equal contribution}
\affiliation{
  \institution{University of Melbourne}
  \city{Melbourne}
  \country{Australia}
}
\email{xueyangk@student.unimelb.edu.au}

\author{Zhengkang Xiang}
\authornotemark[1]  % Repeats the same author note as above
\affiliation{
  \institution{University of Melbourne}
  \city{Melbourne}
  \country{Australia}
}
\email{zhengkangx@student.unimelb.edu.au}

\author{Zezheng Zhang}
\affiliation{
  \institution{University of Melbourne}
  \city{Melbourne}
  \country{Australia}
}
\email{zezheng.zhang1@student.unimelb.edu.au}

\author{Kourosh Khoshelham}
\affiliation{
  \institution{University of Melbourne}
  \city{Melbourne}
  \country{Australia}
}
\email{k.k@unimelb.edu.au}

\ccsdesc[100]{Computing methodologies~Artificial intelligence~Computer vision~Computer vision representations~Hierarchical representations}

\keywords{Novel View Synthesis, Panorama 360 Image, Video Diffusion, Spatial Diffusion, Diffusion Transformer}
%% A "teaser" image appears between the author and affiliation
%% information and the body of the document, and typically spans the
%% page

% \received{20 February 2007}
% \received[revised]{12 March 2009}
% \received[accepted]{5 June 2009}
% \twocolumn[{%
% \renewcommand\twocolumn[1][]{#1}%
% \maketitle
\begin{teaserfigure}
    % \vspace{-2.0em}
\includegraphics[width=\linewidth, trim=10 0 20 0, clip]{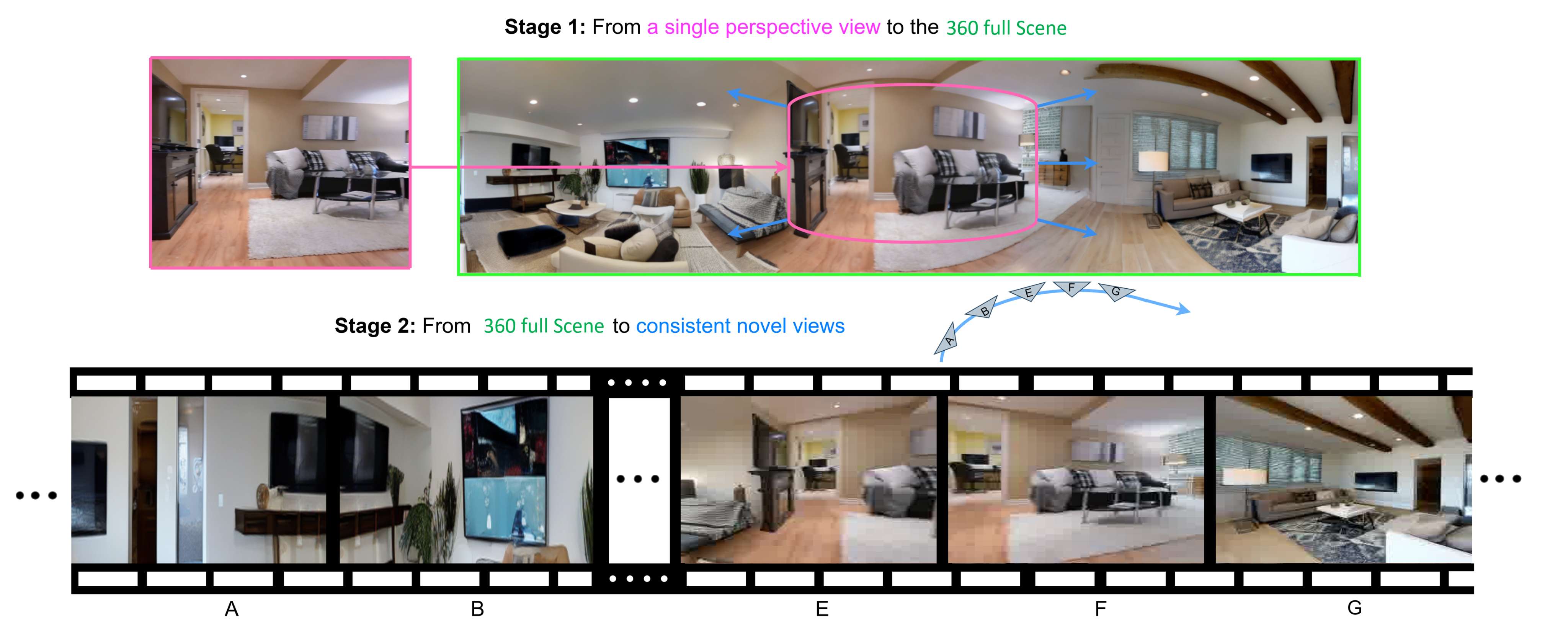}
\vspace{-2.2em}
\caption{Two-Stage Scene View Diffusion with Camera Control. Stage 1 expands a single view into a $360^{\circ}$ panorama (green box). Stage 2 generates consistent video frames by interpolating view frames (A–G) from the panorama, enabling smooth transitions across perspectives. Pink and blue arrows indicate input view and view synthesis motion directions, respectively.}
\label{fig:teaser}
\vspace{1em}
% }]
\Description{Teaser image.}
\end{teaserfigure}

\begin{abstract}
Novel view synthesis (NVS) from a single image is highly ill-posed due to large unobserved regions, especially for views that deviate significantly from the input. While existing methods focus on consistency between the source and generated views, they often fail to maintain coherence and correct view alignment across long-range or looped trajectories. We propose a model that addresses this by decomposing single-view NVS into a 360-degree scene extrapolation followed by novel view interpolation. This design ensures long-term view and scene consistency by conditioning on keyframes extracted and warped from a generated panoramic representation. In the first stage, a panorama diffusion model learns the scene prior from the input perspective image. Perspective keyframes are then sampled and warped from the panorama and used as anchor frames in a pre-trained video diffusion model, which generates novel views through a proposed spatial noise diffusion process. Compared to the prior work, our method produces globally consistent novel views—even in loop-closure scenarios, while enabling flexible camera control. Experiments on diverse scene datasets demonstrate that our approach outperforms existing methods in generating coherent views along user-defined trajectories. Our implementation is available at \textcolor{blue}{\texttt{\href{https://github.com/YiGuYT/LookBeyond}{https://github.com/YiGuYT/LookBeyond}}}.
% through neighboring positions or zoom-in, 
 %The generated videos are further used on downstream 3D reconstruction tasks, to demonstrate the superior spatial coherence of our model. 
 % \vspace{-1.4em}
\end{abstract}    
%%
%% This command processes the author and affiliation and title
%% information and builds the first part of the formatted document.
\maketitle

% \medskip\\
\section{Introduction}
Novel view synthesis (NVS) enables immersive 3D scene exploration from limited inputs, benefiting applications in mixed reality, robotics, and gaming. Generating novel views from a single image remains challenging due to occlusions and large unobserved areas, especially for looped or distant viewpoints that require plausible hallucination with multi-view consistency.

Diffusion-based models like Zero123 \cite{shi2023zero123++} and its scene-level extension \cite{tseng2023consistent} address this using epipolar geometry. PhotoNVS \cite{yu2023long} introduces autoregressive view generation, while CAT3D \cite{gao2024cat3d} and MultiDiff \cite{muller2024multidiff} enable parallel multi-view synthesis with geometric conditioning. Yet, achieving global consistency across wide baselines remains difficult.

Recent hybrid approaches integrate generative models with 3D priors or language guidance. WonderJourney \cite{yu2024wonderjourney} uses LLMs and VLMs for scene extrapolation, ViewCrafter \cite{yu2024viewcrafter} combines 3D Gaussian Splatting (GS) with video diffusion refinement, and VistaDream \cite{wang2024vistadream} applies RGB-D inpainting with GS scaffolds. These methods perform well but rely on dense training and accurate 3D inputs.
To address these challenges, we propose a novel two-stage diffusion framework that decomposes long-range view synthesis into: panoramic scene generation and trajectory-aware view interpolation. Our method first reconstructs a 360-degree panoramic scene from a single image, then synthesizes consistent novel views by conditioning a video diffusion model on anchor frames extracted from the panorama. This enables coherent long-term navigation, including loop closures and scene diversity. In principle, the approach can be scaled by iteratively outpainting new panoramic scenes and generating videos, extending the navigable environment indefinitely. Our main technical contributions are three-fold:

\begin{itemize}
\item A $360^{\circ}$ panorama-guided view synthesis approach that reconstructs a complete panoramic scene from a single image, allowing reliable keyframe extraction and warping for both neighboring and walk-in viewpoints. This panoramic representation serves as a geometric prior, overcoming the long-term view consistency limitations of existing single-view video generation methods.

\item A trajectory-aware video diffusion model that simultaneously synthesizes and inpaints multiple novel views, conditioned on extracted keyframes, inpainting masks, and camera motion. Our model incorporates Plücker embedding raymaps to encode concatenated camera trajectory poses.

\item A spatial sampling strategy that enforces long-term loop consistency. By conditioning the diffusion process on all keyframes mapped and warped from the panoramic scene, our method maintains visual coherence across views and significantly outperforms state-of-the-art NVS methods in both view consistency and correct novel view alignment.
\end{itemize}

\section{Related Work}
\noindent\textbf{Panoramic Image Generation.} Generating 360-degree panoramic images has garnered interest in immersive virtual and augmented reality applications. Previous works adapt text-to-image (T2I) models to produce panoramas from text prompts \cite{liu2024panofree, wang2024customizing, zhang2024taming} or through progressive outpainting from narrow views \cite{wu2023panodiffusion}. The generation of conditioning in the layout of the room \cite{Xu_2021_CVPR} improves the physical plausibility, while downstream applications include 3D reconstruction and the completion of the scene \cite{minfenli2024GenRC, yang2024layerpano3dlayered3dpanorama}.

\noindent\textbf{Video Diffusion.} Video diffusion models have become effective tools for generating diverse video content and synthesizing novel views \cite{voleti2024sv3d, kwak2024vivid, chen2024v3d}, with applications in 3D reconstruction \cite{melas20243d}. Text-to-Video approaches \cite{khachatryan2023text2video, wu2023tune, ho2022video, ho2022imagen} use prompts for guided generation, as seen in WonderJourney \cite{yu2024wonderjourney} and Structure Diffusion \cite{esser2023structure}. Models such as Lumiere \cite{bar2024lumiere}, CogVideoX \cite{yang2024cogvideox}, and VideoCrafter2 \cite{chen2024videocrafter2} have advanced text-guided generation but struggle with scene consistency and flexible camera control. To address this, models like CameraCtrl \cite{he2024cameractrl}, Direct-a-Video \cite{yang2024direct}, Collaborative Video Diffusion \cite{kuang2024cvd}, and VD3D \cite{bahmani2024vd3d} incorporate camera motion awareness, yet still face challenges under large camera shifts such as 360-degree pans. Recent works improve control by conditioning camera parameters or depth-warped references \cite{kang2025multi, gao2024cat3d, zhang2023controlvideo, muller2024multidiff, yu2024viewcrafter}. Others approach video generation as an interpolation between frames to ensure temporal coherence \cite{niklaus2017video, bao2019depth, liu2019deep, huang2022real, singer2022make, shi2022video, danier2023ldmvfi, wang2024generative}.

\noindent\textbf{3D Scene Generation.} Progress in 3D scene generation includes text-driven scene synthesis \cite{fridman2024scenescape, tang2024diffuscene, seo2023ditto} and landscape videos from single images \cite{chai2023persistent}. Methods also generate 3D mesh assets from text or images \cite{wu2024unique3d, xu2024instantmesh, ju2024diffindscene, Liu2023MeshDiffusion, stan2023ldm3d, lin2023magic3d, tang2024diffuscene, poole2022dreamfusion}. SceneCraft \cite{yang2024scenecraft} uses 3D semantic layouts to generate proxy images for NeRF-based scene learning. LT3SD \cite{meng2024lt3sd} further improves synthesis via a latent tree structure for generating complete or partial indoor meshes at arbitrary scale.

\begin{figure*}[!tbp]
    \centering
    \includegraphics[width=\textwidth,  trim=4pt 4pt 21pt 0pt, clip]{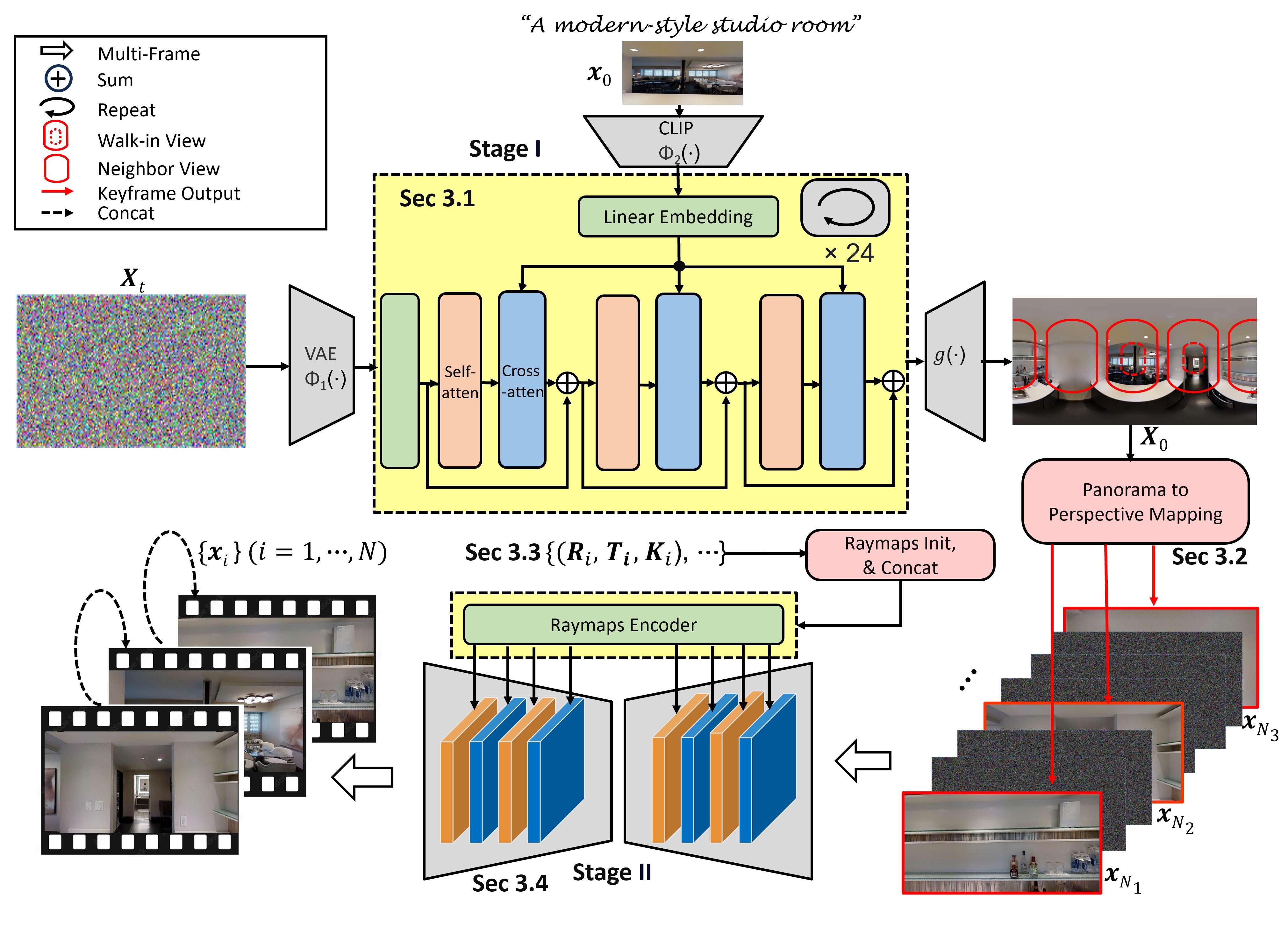}
    \vspace{-2.6em}
    \caption{The process comprises two stages. Stage I uses a Diffusion Transformer (DiT) to outpaint a $360^{\circ}$ panorama from an initial noisy image, guided by view-region rotation consistency loss and CLIP-based image or text prompt $\mathbf{x}_0$ embeddings. The panorama $X_0$ is first mapped to multiple perspective views. In Stage II, keyframes are constructed either from adjacent views (red windows) derived from the panorama or by simulating a walk-in direction (dashed red window) from an inpainted source view. These keyframes $(x_{N_k}^{}, x_{N_{k+1}}^{})$ are then used for video frame interpolation via a pre-trained video diffusion model, conditioned on concatenated raymap features initialized from camera poses ${\mathbf{R}_i, \mathbf{T}_i}$ and intrinsics $\mathbf{K}_i$. Yellow blocks contain learnable layers and gray blocks contain frozen pre-trained parameters.
} %  During inference, the input view maps to a masked panorama $X^{\text{known}}_t$ (top left).
    \label{fig:over-diagram}
    \vspace{-1.0em}
\Description{Overview diagram.}
\end{figure*}
\section{Method}
The problem of scene video diffusion from a single input image can be formulated as sampling from the joint distribution $p_{\theta}$ to generate a sequence of images along a specified trajectory while maintaining both spatial and temporal consistency,
\begin{equation}
p_{\theta}(\mathbf{x}_i | \mathbf{x}_0, {\mathbf{R}_i, \mathbf{T}_i, \mathbf{K}_i}),\quad (i = 1,\ldots,N).
\label{eq:problem-state}
\end{equation}
Here, $\mathbf{x}_{0} \in \mathbb{R}^{H \times W \times C}$ is the source input image or text prompt of the scene, where $H$, $W$, and $C$ represent height, width, and number of channels, respectively. The trajectory is defined by a sequence of camera poses, each including rotation $\mathbf{R}_i$ and translation $\mathbf{T}_i$, along with intrinsic parameters $\mathbf{K}_i$, and the intrinsic matrix $\mathbf{K}_i$ remains constant across all images in the same sequence.
Directly sampling from this joint conditional distribution can lead to ambiguities in long-term video generation. Therefore, we propose to decompose the distribution into a factorized keyframe distribution, followed by view frame interpolation between keyframes, as illustrated in Figure \ref{fig:teaser}. First, the input image $\mathbf{x}_0$ is mapped onto a $360^{\circ}$ panorama mask and outpainted by a diffusion model to generate a panorama image $\mathbf{X}_0$. This panorama is then decomposed into several perspective keyframes $\mathbf{x}_k, (k = {1, \ldots, N^*})$. A subset of these keyframes are selected and paired based on their relative positions in the panorama, to serve as source-target pairs for video diffusion-based novel view interpolation.
The key aspects of this formulation are the use of a panoramic scene representation as a geometric prior, and the proposed decomposition of the joint distribution into a more tractable keyframe-based process. This approach helps maintain long-term visual coherence in the generated video sequences.

In the following, we begin describing the use of a Diffusion Transformer (DiT) to generate a $360^{\circ}$ panorama image. Section \ref{sec:block2} covers the extraction of keyframes from the panorama image, which, along with raymaps based on trajectory poses described in section \ref{sec:block3}, serve as conditioning inputs for the video diffusion model in the second stage, as detailed in Section \ref{sec:block4}. The complete pipeline is illustrated in Figure~\ref{fig:over-diagram}.
%When the target view $X_j$ is significantly distant from the source view $X_i$, the overlap may be minimal, limiting its contribution to the sampling process and complicating the sampling distribution task. Our model, as illustrated, leverages relative view geometry as guidance to address these challenges. It is designed to manage view ambiguity when there is minimal overlap between the source and target views. The model seamlessly integrates the conditioning semantic features of the source image $X_i$ with the warped feature map of $X_j^{\text{known}}$, derived from the source image using depth and relative camera transformation.
\subsection{Panorama Diffusion} 
\label{sec:block1}
The panorama diffusion recovers progressively a $360^{\circ}$ panorama image $\mathbf{X}_0$ from a Gaussian noise-corrupted panorama mask $\mathbf{X}_t$. Through iterative reverse denoising, a coherent $360^{\circ}$ panorama can be obtained as the scene context.
The diffusion operates in the latent space of a pre-trained encoder $\Phi_1(\mathbf{X}_0)$ and decoder $\boldsymbol{g}(\cdot)$, along with a conditioned image embedding $\Phi_2(\mathbf{x}_0)$ via CLIP of the input image.
Particularly during inference, the perspective view image $\mathbf{x}_0$ is first mapped onto the panorama mask through the camera model mapping function $\Pi_{\text{Per}}^{\text{Pan}}$ for inference outpainting, which will be clarified below in reverse form. The outpainting diffusion then operates in the latent space of a VAE encoder \cite{kingma2013auto}, denoted as $\Phi_1(\mathbf{X}_t) = \mathbf{z} \in \mathbb{R}^{ H \times W \times C}$, where $C$ is the number of latent channels. The outpainting process can be formulated as:
\begin{align}
\mathbf{z}_{t-1}^{\text{known}} &= p_{\theta}(\mathbf{z}_{t-1}|\mathbf{z}_{t-2}), \\
\mathbf{z}_{t-1}^{\text{unknown}} &= q{}(\mathbf{z}_{t}|\mathbf{z}_{t-1}),  \\
\mathbf{z}_{t-1} &= \mathbf{m} \odot \mathbf{z}_{t-1}^{\text{known}} + (\mathbf{1} - \mathbf{m}) \odot \mathbf{z}_{t-1}^{\text{unknown}},
\label{eq:inpaint}
\end{align}
where the binary mask $\mathbf{m}$ represents known regions,and its complement $(\mathbf{1} - \mathbf{m})$ indicates unknown regions. The feature map $\mathbf{z}_{t-1}$ is derived by fusing the known regions through forward diffusion $p_{\theta}$ and the unknown regions through reverse denoising $q$.
Specifically, $\mathbf{z}_{t-1}$ is computed as below:
\begin{equation}
\mathbf{z}_{t-1} = \frac{1}{\sqrt{\alpha_t}} \left( \mathbf{z}_t - \frac{1 - \alpha_t}{\sqrt{1 - \bar{\alpha}_t}} \epsilon_\theta(\mathbf{z}_t, t) \right) + \sigma_t \mathbf{u}_t,
\label{eq:denoise}
\end{equation}
Where $\alpha_t$ is a gain to control the noise added in the forward process, $\bar{\alpha}_t = \prod_{k=1}^t \alpha_k$, and $\mathbf{u}_t \sim \mathcal{N}(0, \mathbf{I})$. The term $\epsilon\theta(\mathbf{z}_t, t)$ represents the predicted score function of noise to denoise $\mathbf{z}_t$ into $\mathbf{z}_{t-1}$.
To preserve the 360-cycle view consistency of the generated panorama, we use a cycle consistency loss from PanoDiffusion \cite{wu2023panodiffusion} during sampling, rotating the denoised panorama feature map by 90-degree intervals iteratively to form a loop view.

% Given a static input panorama \( I \) and an initial spherical latent code \( S^0: S^2 \rightarrow \mathbb{R}^{L \times c} \), we progressively remove noise employing a project-and-fuse procedure at each denoising step. Specifically, the spherical latent code at the \( t^{th} \) denoising step, \( S^t: S^2 \rightarrow \mathbb{R}^{L \times c} \), is projected into multiple perspective latent codes \( \mathcal{Z}^t = \{z_1^t, z_2^t, \dots, z_n^t\} \), where each \( z_k^t = \gamma(S^t, d_k) \in \mathbb{R}^{p_H \times p_W \times (L \times c)} \) represents the \( k^{th} \) perspective latent code projected in the equirectangular format

% Each perspective latent code is then denoised by one step using a pre-trained perspective denoiser, denoted as \( z_k^{t-1} = \Phi(z_k^t, i_k) \), where \( i_k = \gamma(I, d_k) \in \mathbb{R}^{p_H \times p_W \times c} \) is the perspective conditioning image projected from the panorama \( I \). 

% Subsequently, we optimize the spherical latent code \( S^t: S^2 \rightarrow \mathbb{R}^{L \times c} \) at step \( t-1 \) by fusing all the denoised perspective latent codes \( z_k^{t-1} \). Formally, the denoising procedure at step \( t \), denoted as \( S^{t-1} = \Psi(S^t, I) \)

\subsection{Panorama to Perspective Mapping}
\label{sec:block2}
The final denoised panorama feature from the diffusion model, $\mathbf{z}_0$, is decoded by $\boldsymbol{g}(\mathbf{z}_0)$ to generate the panorama image $\mathbf{X}_0$. Perspective view keyframes are then generated from this panorama image via equi-projection $\Pi_{\text{Pan}}^{\text{Per}}$,
\begin{equation}
\mathbf{x}_k = \Pi_{\text{Pan}}^{\text{Per}}(\mathbf{X}_0, \theta_k, \phi_k, \beta, \gamma, H, W), \quad k = {1, \ldots, N^{*}},
\label{eq:pano-pers}
\end{equation}
where $\mathbf{X}_0$ is the equirectangular panorama. $\theta_k^{}$ and $\phi_k^{}$ are the yaw and pitch angles for each perspective view $\mathbf{x}_k$ position on panorama view. $\beta$ and $\gamma$ are the horizontal and vertical field of view (FOV), and $H$ and $W$ are the height and width of the perspective images.
%consists of five steps: a) converting pixel coordinates to 3D direction vectors, b) applying a rotation matrix based on $\theta_k$ and $\phi_k$, c) computing latitude and longitude, d) mapping back to pixel coordinates in $\mathbf{X}_0$, and e) extracting the color values.
\begin{figure}[!thbp]
    % \vspace{-1.0em}
    \centering
    \includegraphics[width=\columnwidth]{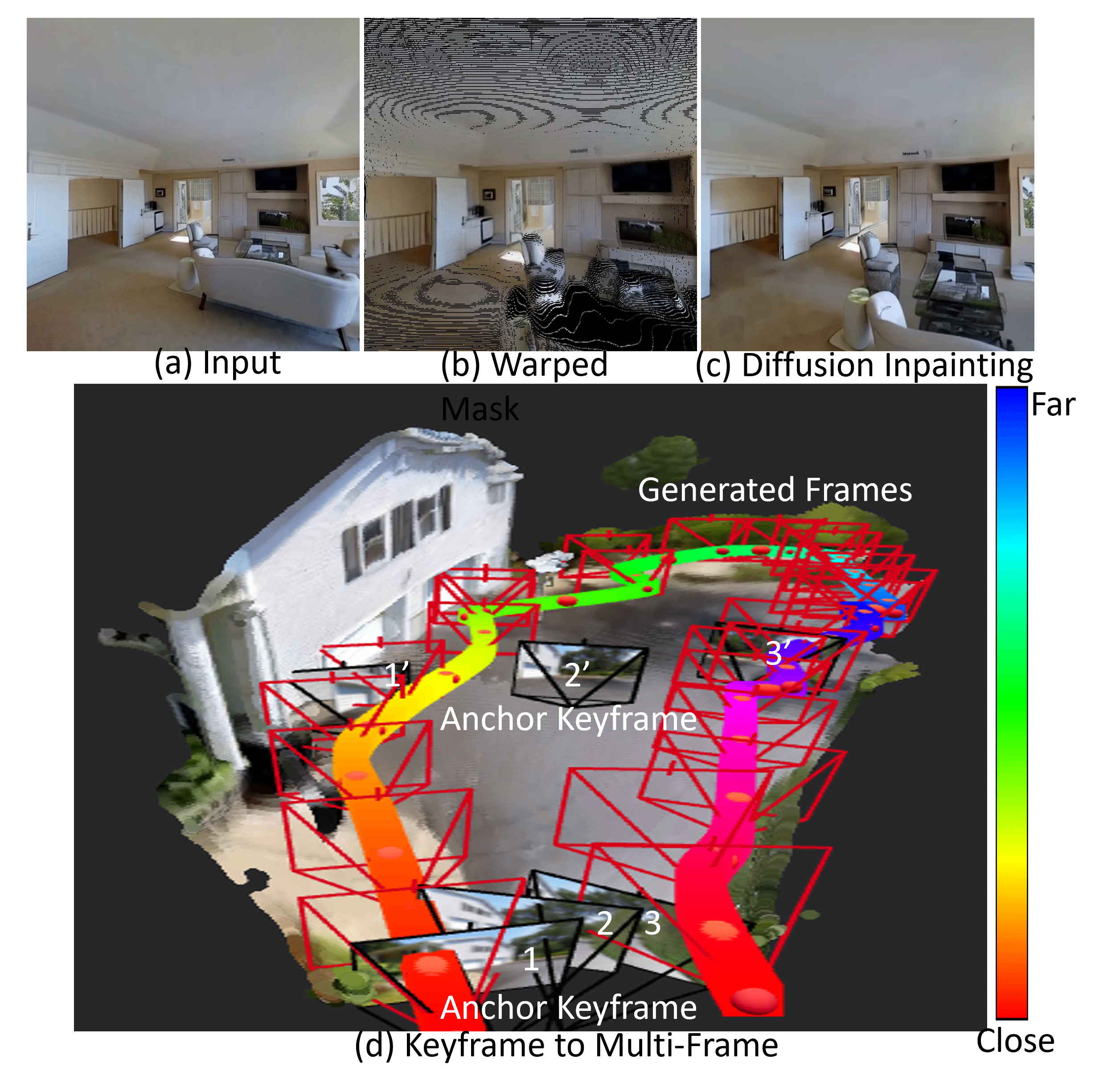}
    \vspace{-2.6em}
    \caption{(a) Input view. (b) Linearly warped view from source to target via motion transform, with inpainting mask. (c) The target view was synthesized using our video diffusion model. (d) Scene walk trajectory with red camera frustums indicating generated views and black frustums as anchor keyframes. Keyframes 1-3 are perspective views from panoramas; 1'-3' are corresponding warped views via walk-in motion. The color bar shows the relative distance from the camera.}
    \label{fig:keyframe-frame-inpainting}
    \vspace{-2.0em}
\Description{Keyframe extraction.}
\end{figure}

We generate perspective keyframes using two strategies: (1) selecting neighboring views on the panorama via panning rotation, and (2) simulating a walk-in motion from a chosen viewing direction. For (1), anchor frames are selected using a fixed angular interval (typically ${30-60}^{\circ}$) by rotating on the panorama, defining keyframes (e.g., 1, 2, 3). For (2), a forward walk-in motion is simulated by generating offset keyframes (1', 2', 3') along the same direction. As illustrated at the bottom of \ref{fig:keyframe-frame-inpainting}, this results in warped keyframes distributed farther and more uniformly across the scene compared to the original ones (1–3), which are mapped from the panorama using \ref{eq:pano-pers}. Each keyframe pair (e.g., 1–1') is used for interpolated frame generation. These additional keyframes provide broader spatial overlap, enhancing consistency for novel view synthesis. The target view inpainting is conditioned on the warped mask and guided by the view diffusion process in the second stage. The source view is rescaled according to the walking distance for walk-in motion. The scaling factor for resizing the cropped region is computed as:

\begin{align}
s &= f \cdot {\tan\left((1 - c) \cdot \beta/2\right)}^{-1},
\label{eq:scale-factor}
\end{align}

where $c = \hat{d} / d_{\text{Max}}$ is the normalized depth, $\hat{d}$ is the target distance, $d_{\text{Max}}$ is the maximum depth of the scene scene and $f$ is the focal length. This scale \( s \) adjusts the panorama crop to a perspective view of size $H~\times W~\times 3$, simulating a closer view. The cropped region is centered on the candidate view, whereas interpolated frames generate peripheral out-of-FOV areas. More details on the panorama-to-perspective mapping and analysis results are provided in the supplementary material. The warp is applied on the downsampled feature map $\mathbf{z}_i \in \mathbb{R}^{H' \times W' \times 3}$ extracted from the source image $\mathbf{x}_i$. The warping function $h(\mathbf{x}_i, d_i, \mathbf{q}_i, \mathbf{T}_j)$ maps the source pixels to the target view using:

\begin{align}
(m', n') \simeq \pi\left( \mathbf{T}_j \cdot \pi^{-1}((m,n), \mathbf{d}_i(m,n)) \right),
\label{eq:warp_brief}
\end{align}

where $\pi$ and $\pi^{-1}$ are the projection and back-projection operations defined by the intrinsics $\mathbf{K}$. Here, $(m,n)$ are the source pixel position, and $(m', n') \in \mathbf{M}_{\text{warp}}$ are the warped target position.

%As the depth used here is either measured by an RGB-D sensor, as in ScanNet \cite{dai2017scannet}, or predicted through a monocular depth model  \cite{bhat2023zoedepth}, it contains some uncertainties.
%For a perspective camera with focal length \(f\) and initial width \(W\) of the view, the horizontal FOV angle \(\theta\) is defined by \(\theta = 2 \arctan{\left(\frac{W}{2f}\right)}\). Moving to a closer depth-normalized distance results in a zoom factor \(z = \frac{1}{\text{depth\_norm}}\), where \(\text{depth\_norm} = \frac{d}{D}\). The cropped width \(W'\) and height \(H'\) are scaled by this zoom factor, giving \(W' = W \cdot \text{depth\_norm}\) and \(H' = H \cdot \text{depth\_norm}\). This scaling aligns with a new effective focal length \(f' = f \cdot \text{depth\_norm}\), maintaining the FOV angle \(\theta\) by preserving \(\tan{\left(\frac{\theta}{2}\right)} = \frac{W'}{2f'}\). Thus, the cropped region centers on the target with intrinsic parameters adjusted to reflect the new viewing distance \(d\). 

%\cite{seo2024genwarp}

\subsection{Pose Embedding Raymap}
\label{sec:block3}
For camera pose control, we use ground truth poses ${\mathbf{R}_i, \mathbf{T}_i, \mathbf{K}_i}, \; i = 1, \ldots, N$ provided in the dataset. Each view is represented by a raymap $\mathbf{r}_i \in \mathbb{R}^{H \times W \times 6}$, initialized using the Plücker 6D embedding \cite{zhang2024raydiffusion}. The raymaps are concatenated along the channel dimension and encoded by a shared encoder $\Phi(\cdot)'$, yielding feature maps $\mathbf{W}_{\mathbf{r}} \in \mathbb{R}^{H' \times W' \times (N \times 6)}$. These features serve as conditioning input to the video diffusion model via cross-attention.

%Otherwise, like out of distribution demonstration, the custom video trajectory generation begins with the input of a panorama image $X_0$ and its associated depth map $D_P$, which is either retrieved from the dataset or predicted using a monocular depth model. The panorama is then divided into four parts in horizontal direction and undistorted from spherical coordinates into perspective coordinates. Wall corner points $P_i \, (i=1, \dots, 8)$ are identified from the horizontal middle line of the undistorted panorama depth image, and these points are connected to form a top-down room structure layout. Two strategies are applied to generate the trajectories: one involves walking into and out of the room, and the other employs circular sectors in a room with wall-enclosed space. The resulting trajectory consists of a combination of turns and straight-line motions for navigation and testing. More information about custom trajectory pose generation is shared in the supplementary part.

\begin{algorithm}[!tbp] 
\caption{Sampling with Spatial Diffusion} 
\label{alg:spatial_diffusion} 
\begin{algorithmic}[1] 
\Require Keyframes $(x_{i}, x_{i+1})$, $\mathbf{d}_i$, $\mathbf{W}_{\mathbf{r}}^{}$, poses $(\mathbf{q}_{i}^{}, \mathbf{T}_{i}), (\mathbf{q}_{i+1}^{}, \mathbf{T}_{i+1})$, $f_\theta (\cdot)$, $\mathcal{D}(\cdot)$, time step $t$, inpainting mask $\mathbf{M}$ 
\State \text{Compute CLIP features} $\mathbf{c}_{i} = \text{CLIP}(x_{i})$, $\mathbf{c}_{i+1} = \text{CLIP}(x_{i+1})$; 
\State \text{Warped mask} $\mathbf{M}_{\text{warp}} = h(x_{i}, \mathbf{d}_i, \mathbf{q}_{i}^{}, \mathbf{T}_{i}^{})$; 
\State \text{Apply warped mask to keyframe:} $x_{i}^{\text{masked}} \gets x_{i} \odot (1 - \mathbf{M}_{\text{warp}})$; 
\State \text{$(\mathbf{c}_{i}, \mathbf{c}_{i+1})$ Condition diffusion with inpainting mask $x_{i}^{\text{masked}}$} ; 
\State \text{Set} $\mathbf{W}_{\mathbf{r}} \gets \{\mathbf{r}_i\}$ \text{, where} ${\mathbf{r}}_i$ \text{is Plücker embedding raymap}; 
\State \text{Initialize} $\mathbf{z}_T \sim \mathcal{N}(0, \mathbf{I})$; 
\For{$t = T \to 1$} 
    \State \text{Get interpolated pose} $\mathbf{q}_{j}^{}, \mathbf{T}_{j}^{}$ \text{for frame} $j$ \text{}; 
    \State \text{Compute position weights} ${\omega}_j^{i}$ \text{from} $d_\textsc{Pos}(\mathbf{T}_{j}^{}, \mathbf{T}_{i}^{})$; 
    \State \text{Compute orientation weights} ${\beta}_j^{i}$ \text{from} $d_{\theta}(\mathbf{q}_{j}^{} , \mathbf{q}_{i}^{})$; 
    \State \text{Normalize weights} ${\gamma}_j^{i} \gets \frac{({\omega}_j^{i} \cdot {\beta}_j^{i})}{\sum_k ({\omega}_j^{k} \cdot {\beta}_j^{k})}$; 
    \State \text{Predict} $\hat{\mathbf{v}}_{t}^j \gets f_\theta(\mathbf{z}_t; t, (\mathbf{c}_i, \mathbf{c}_{i+1}), \mathbf{W}_{\mathbf{r}}^{})$; 
    \State \text{Compute noise} $\boldsymbol{\epsilon}_\theta \gets (\mathbf{z}_t -  {\gamma}_j^{i}\cdot\sqrt{\alpha_t} \cdot \hat{\mathbf{v}}_{t}^j) / \sqrt{1 - \alpha_t}$; 
    \State \text{Update} $\mathbf{z}_{t-1} \gets \frac{1}{\sqrt{\alpha_t}} \left( \mathbf{z}_t - \frac{1 - \alpha_t}{\sqrt{1 - \bar{\alpha}_t}} \boldsymbol{\epsilon}_\theta \right) + \sigma_t \mathbf{u}_t$; 
\EndFor 
\State \Return $\mathcal{D}(\mathbf{z}_0)$ 
\end{algorithmic} 
\end{algorithm}
\vspace{-0.6em}
\subsection{Video Diffusion}
\label{sec:block4}
Given keyframe pairs $(x_i, x_{i+1})$, video frame interpolation under a given camera trajectory is synthesized via Algorithm~\ref{alg:spatial_diffusion}. The proposed sampling with spatial diffusion incorporates keyframe-based operations (rows 1-5) for CLIP feature extraction and warped mask-based conditioning, followed by latent video diffusion (rows 6-16) with temporally-aware and spatially-aware denoising processes. The method uses CLIP-encoded features $(\mathbf{c}_i, \mathbf{c}_{i+1})$ from both source and target keyframes as cross-attention conditions for frame interpolation, while raymap $\mathbf{W}_{\mathbf{r}}$ provides pose guidance for view-aligned synthesis. The process consists of an outer loop over reverse time steps and inner computation for spatially guided weighting across interpolated frames between keyframes, which are then composed into longer video sequences.

\noindent\textbf{Spatial Interpolation (Lines 8--11 in Algorithm~\ref{alg:spatial_diffusion}):}
To guide denoising with spatial correspondence, the algorithm computes per-frame weights based on camera geometric relations. At each time step $t$, it retrieves the pose $\mathbf{q}_i^{}, \mathbf{T}_{i}^{}$ of the current frame $i$ and compares it to anchor frame $j$. The positional proximity ${\omega}_i^{j}$ and orientation similarity ${\beta}_i^{j}$ are calculated via: 

% \vspace{-1.2em} 
\begin{align} d_\textsc{Pos}^{}(\mathbf{T}_{i}^{}, \mathbf{T}_{j}^{}) &= \exp\left(-|\mathbf{T}_{i}^{} - \mathbf{T}_{j}^{}|2 / \tau_\mathbf{T}\right), \\
\ d_{\theta}^{}(\mathbf{q}_{i}^{}, \mathbf{q}_{j}^{}) &= \exp\left(-2\cdot\arccos\left(\mathbf{q}_{i}^{} \cdot \mathbf{q}_{j}^{*}\right) / \tau_\mathbf{q}\right), \label{eq:distance} \end{align} 
% \vspace{-1.2em}

\noindent where $\tau_\mathbf{T}$ and $\tau_\mathbf{q}$ are temperature factors controlling spatial and angular influence, respectively, and $\mathbf{q}_{j}^{*}$ is the conjugate quaternion of $\mathbf{q}_{j}^{}$. The final fusion weight is obtained by normalizing the joint product $\gamma_i^{j} \gets \frac{(\omega_i^{j} \cdot \beta_i^{j})}{(\omega_i^{j} \cdot \beta_i^{j})}$ for spatial consistency guidance in the diffusion step.

\noindent\textbf{Diffusion Noise (Lines 6 and 12--14 in Algorithm~\ref{alg:spatial_diffusion}):}
The outer loop performs a denoising diffusion procedure from $t = T$ to $1$. The latent variable $\mathbf{z}_T$ is initialized as Gaussian noise $\mathcal{N}(0, \mathbf{I})$. At each step, the model predicts intermediate noise-free latent feature volume $\hat{\mathbf{v}}_t^i$ step by step using the denoising network $f\_theta(\cdot)$, conditioned on masked keyframe feature $\mathbf{c}_i$ and ray-based embedding $\mathbf{W}_{\mathbf{r}}$. The noise estimate $\boldsymbol{\epsilon}_\theta$ is derived by subtracting the weighted clean prediction ${\gamma}_i^{j} \cdot \sqrt{\alpha_t} \cdot \hat{\mathbf{v}}_{t}^i$ from the current noisy latent $\mathbf{z}_t$, and scaling by $\sqrt{1 - \alpha_t}$. The update rule follows the standard DDPM backward step:
% \vspace{-0.5em} 
\begin{equation} 
\mathbf{z}_{t-1} \gets \frac{1}{\sqrt{\alpha_t}} \left( \mathbf{z}_t - \frac{1 - \alpha_t}{\sqrt{1 - \bar{\alpha}_t}} \boldsymbol{\epsilon}_\theta \right) + \sigma_t \mathbf{u}_t, \end{equation} 
% \vspace{-0.8em}

\noindent where $\sigma_t$ is the noise variance and $\mathbf{u}_t \sim \mathcal{N}(0, \mathbf{I})$. The final denoised feature volume $\mathbf{z}_0$ is decoded into multiple output video frames via the decoder $\mathcal{D}(\cdot)$. The algorithm performs keyframe-anchored inpainting with diffusion generation along a user-defined trajectory. Each keyframe $x_t$ is fused with an inpainting mask $m_t$ via a pixel-wise Hadamard product, \emph{i.e.}, $x_t \otimes m_t$, guiding content restoration to occluded regions, similar to the inpainting strategy in Eq.~\ref{eq:inpaint}. This encourages the diffusion model to synthesize semantically and geometrically consistent content in the masked areas and generated frames. For additional details on the video diffusion with inpainting, please refer to the supplementary material.

\section{Experiments}
\subsection{Evaluation Datasets \& Metrics} %Kernel Inception Distance (KID) \cite{binkowski2018demystifying} ands
We evaluate panorama generation on the Matterport3D~\cite{Matterport3D} test set, and video diffusion on both RealEstate10K~\cite{zhou2018stereo} and Matterport3D~\cite{Matterport3D}. For Matterport3D video, we generate 500 test episodes (200–300 frames) along navigable trajectories, following Lookout\cite{ren2022look}, using an embodied agent with Habitat~\cite{savva2019habitat}.

We use multiple metrics to assess image and video quality. PSNR measures pixel-wise differences; SSIM~\cite{wang2004image}, Inception Score~\cite{xia2017inception}, and LPIPS~\cite{zhang2018unreasonable} assess pixel-level and perceptual similarity. FID~\cite{heusel2017gans} evaluates feature distribution similarity between real and generated images, while FVD~\cite{unterthiner2018towards} captures temporal coherence in video. FID reflects image quality; FVD captures frame consistency.

To evaluate geometric consistency, we use median threshold symmetric epipolar distance (mTSED)~\cite{yu2023long}. A match is valid if the epipolar error $\mathbf{T}_{\text{error}} < 2.5$ and the matched feature count $\mathbf{T}_{\text{match}} > 10$; we report the match percentage.

FVD is computed by stacking normalized ResNet50~\cite{he2016deep} feature embeddings, scaled back post-encoding to manage memory. See supplementary for FVD details and Matterport3D processing.
% We follow the rendering process described in MVDiffusion \cite{tang2023MVDiffusion} to use a rendering pipeline to generate video images along a customized trajectory, by leveraging the reconstructed scene mesh from panorama images, which can be used as pseudo-GT videos. 
%It firstly extracts the SIFT features \cite{lowe1999object} of a pair of images. Next, given the camera intrinsics and the relative camera motion, the minimal epipolar Euclidean distance between the pixel locations to the corresponding mapped epipolar lines in another image is calculated.
%indicates that all pairwise frames are matched.
% To ensure a comprehensive evaluation of view synthesis consistency, we set the synthesis of target view images to 24 frames from the source view frame. The first 12 frames are categorized as short-range, and the remaining half as long-range, with metric results provided for both subsets.
\subsection{Implementation Details}
Since RealEstate10K~\cite{zhou2018stereo} lacks ground-truth panoramas, we select input frames roughly centered in a room and sample anchor keyframes with sufficient parallax. These are augmented using GenWarp~\cite{seo2024genwarp} to enrich spatial coverage and improve video diffusion learning. Our fine-tuning set includes 5,000 walk-through clips (200–400 frames each), cropped into 20–40 frame segments for parallel training. Randomly sampled anchors guide second-stage interpolation.

For Matterport3D~\cite{Matterport3D}, each panorama is paired with a video starting at the panorama point, incorporating walk-in and rotation to increase perspective diversity. The dataset has ~11,000 episodes (400–600 frames each), following loop or “U”-shaped trajectories generated via Habitat simulator~\cite{savva2019habitat}.

Our panorama diffusion model is based on DiT~\cite{peebles2023scalable}, modified for panoramic resolution, CLIP conditioning, and outpainting-guided sampling. We use an XL DiT with 24 blocks of self- and cross-attention, followed by a linear layer. Patch size is 2, and diffusion features are $H' \times W' \times C' = (32 \times 64 \times 4)$, with input size $(256 \times 512 \times 3)$. During inference, we use a $90^{\circ}$ rotation step for consistency loss. For layout verification or keyframe warping, Depth Anything V2~\cite{depth_anything_v2} estimates depth for RealEstate10K. The walk-in ratio is set to 0.8 of the depth from the panorama center to the room boundary. Each panorama is split into 3–6 overlapping perspective views (one-third or no overlap). The model generates 48 frames per pass with 3–6 inpainted anchors, composing a 30-second scene tour at 12 FPS.

Our panorama diffusion model builds on the Diffusion Transformer (DiT) architecture \cite{peebles2023scalable}, adapted to support panoramic resolutions, CLIP conditioning, and sampling with outpainting score. We employ an XL model with 24 repeating DiT blocks, each containing interleaved self-attention and cross-attention layers, followed by a linear layer. Patch size is set to 2, and the diffusion feature map dimensions are configured as , with input images resized to . During inference sampling, the rotation interval for consistency cycle loss is set to . For room layout verification or keyframe warp mapping, we use Depth Anything V2 \cite{depth_anything_v2} to estimate monocular depth for RealEstate10K, as ground-truth depth is unavailable. The walk-in ratio for the forward motion is set to 0.8 times the scene depth from the center of the panorama to the end of the room. In the keyframe-based method, the panorama is horizontally divided into three to six overlapping perspective views, with one-third or no overlap between adjacent views. The video diffusion model produces 48 frames by decoding feature volume at once, with 3–6 anchor keyframes inpainted. Combining these clips yields a 30-second scene tour video at 12 FPS. %The interval between neighboring frame poses is limited to $5^{\circ}$ and $0.2m$. 

Training the panorama diffusion model requires approximately 7 GPU days on one H100 GPU. Stable Video Diffusion (SVD) is distilled with LoRA \cite{wang2024generative} and fine-tuned over 400k iterations, taking around 3 GPU days on an H100, with fine-tuning restricted to the conditioning encoder layers of raymap features, leaving the SVD model frozen.

\subsection{Baseline Comparisons}

\begin{figure}[!thbp]
    \vspace{-1.0em}
    \centering
    \includegraphics[width=\columnwidth, trim=2.0cm 0cm 0cm 0cm, clip]{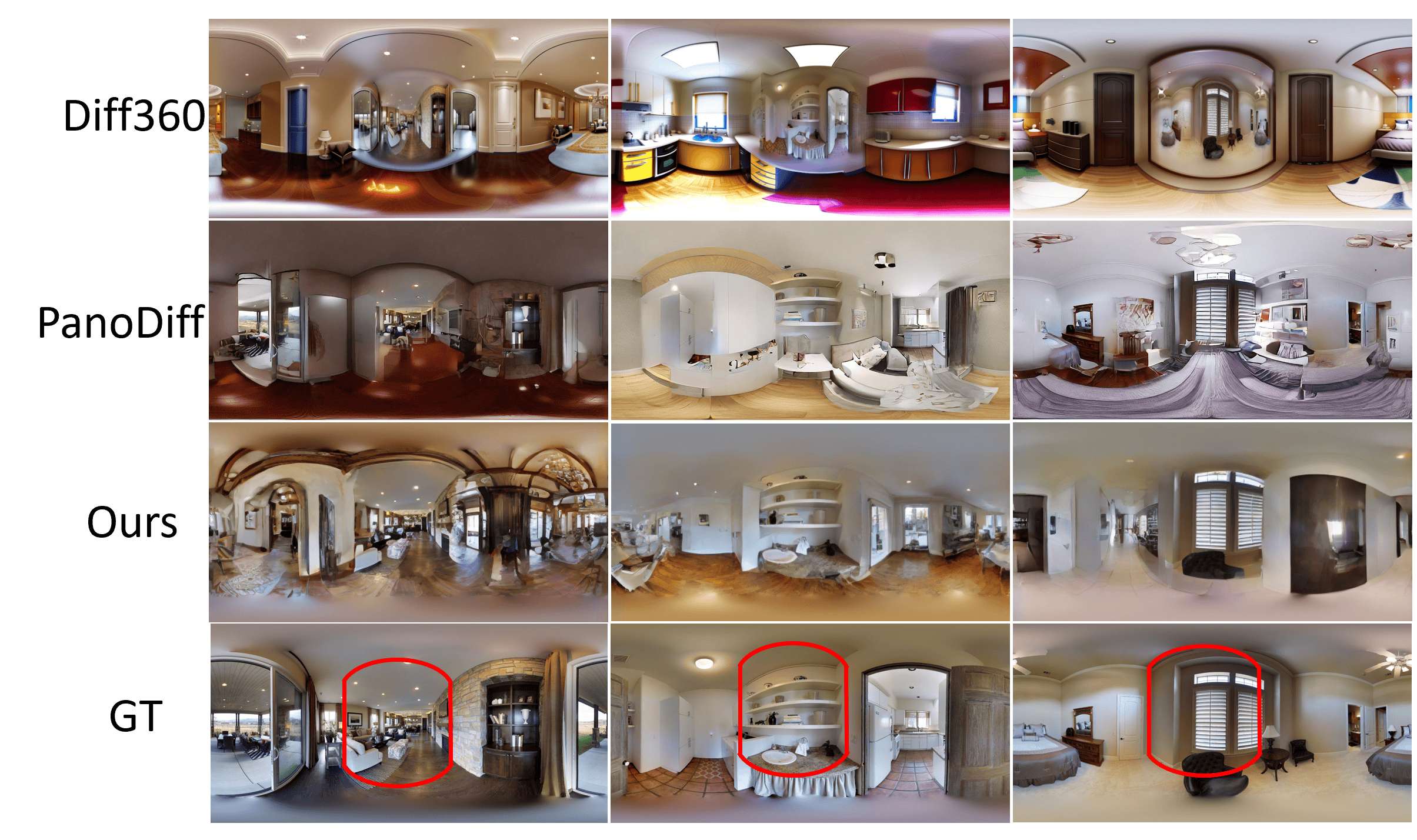}
    \vspace{-2.6em}
    \caption{Qualitative comparison of panorama generation.}
    \label{fig:baseline-pano-cmps}
\Description{A visual baseline comparison for panorama generation.}
    \vspace{-1.4em}
\end{figure}

The qualitative results in Figure \ref{fig:baseline-pano-cmps} show significant differences in panorama generation quality. Diffusion360 \cite{feng2023diffusion360} and PanoDiffusion \cite{wu2023ipoldm} fail to preserve contextual consistency and produce panoramas with lighting and scene elements that differ substantially from the ground truth (GT). The generated panoramas exhibit inconsistent illumination, mismatched architectural details, and poor preservation of the input view context shown in the red-boxed regions. In contrast, our method generates panoramas that maintain better scene coherence, consistent lighting conditions, and more accurate preservation of the spatial context and visual appearance of the input view.
The metrics in Table \ref{tab:panorama-comparison} confirm visual observations, with our method achieving superior performance across all measures: lowest perceptual distance (LPIPS: 0.49), highest reconstruction quality (PSNR: 14.04), best structural similarity (SSIM: 0.49), significantly lower distributional difference (FID: 52.51), and highest inception score (IS: 3.51). These results demonstrate our method's ability to generate more realistic, contextually consistent panoramas that better preserve the input view's context while maintaining overall scene consistency.
\begin{table}[!htbp]
\centering
\caption{Quantitative comparison of panorama generation methods. Our method is the best across all metrics.}
\vspace{-0.8em}
\begin{adjustbox}{width=\columnwidth} 
\begin{tabular}{lccccc}
\toprule
Model & LPIPS $\downarrow$ & PSNR $\uparrow$ & SSIM $\uparrow$ & FID $\downarrow$ & IS $\uparrow$ \\
\midrule
Diffusion360 \cite{feng2023diffusion360} & 0.57 & 10.98 & 0.37 & 180.43 & 1.72 \\
PanoDiffusion \cite{wu2023ipoldm} & 0.56 & 11.56 & 0.49 & 203.41 & 1.85 \\
\midrule
\textbf{Ours} & \textbf{0.49} & \textbf{14.04} & \textbf{0.49} & \textbf{52.51} & \textbf{3.51} \\
\bottomrule
\end{tabular}
\end{adjustbox}
\label{tab:panorama-comparison}
\end{table}

\begin{figure}[!thbp]
    \vspace{-1.0em}
    \centering
    % \includegraphics[width=\columnwidth]{figures/matterport1-cmps.png} \\
    % \vspace{-2.0em}
    % % Dashed line separator
    % % \vspace{0.5em}
    % \tikz{\draw[dashed, thick] (0,0) -- (\linewidth,0);}
    % % \vspace{0.5em}
    \includegraphics[width=\columnwidth, trim=0.0cm 0.4cm 0cm 0cm, clip]{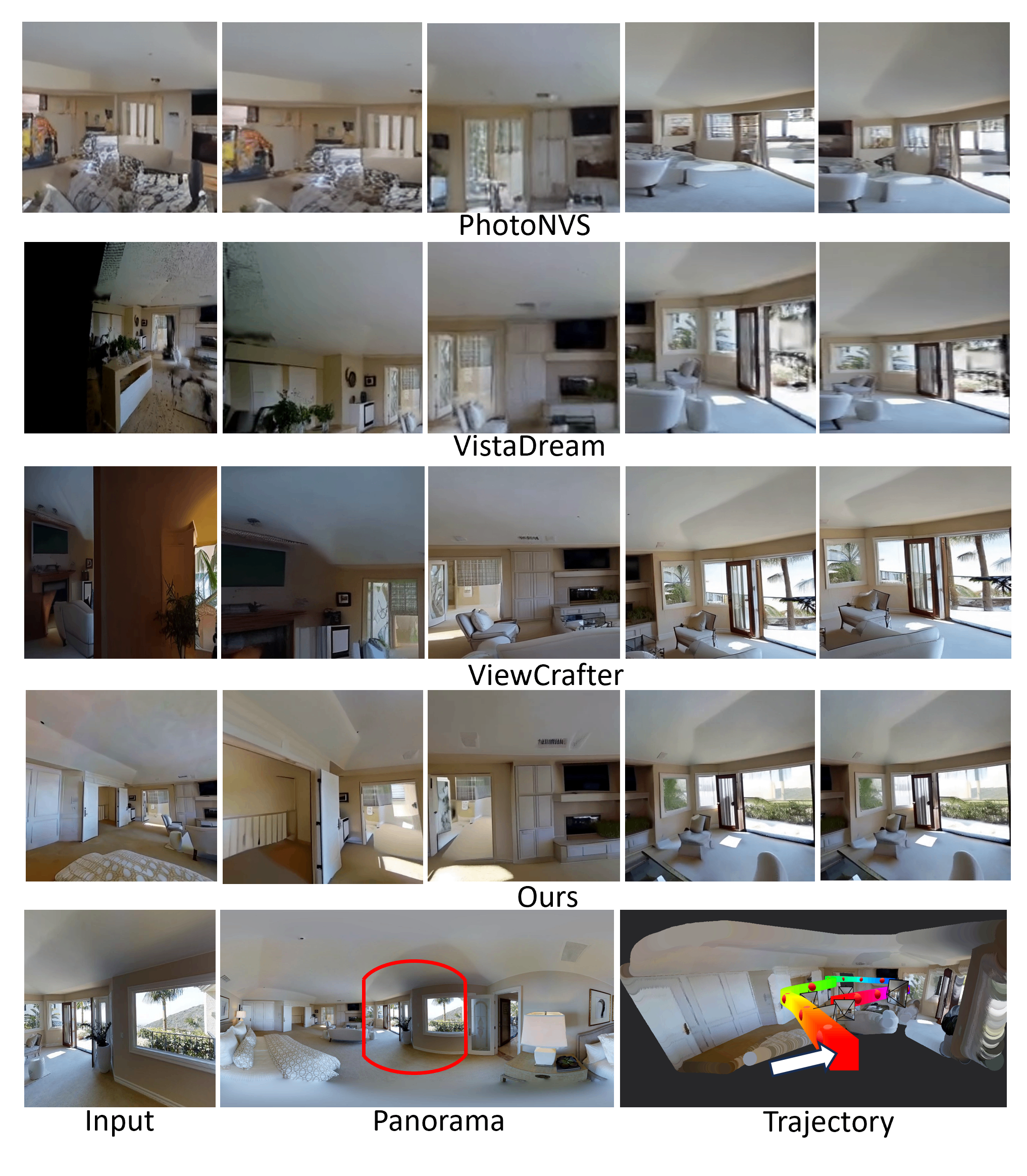}
    \vspace{-2.6em}
    \caption{Generated indoor frames along our predefined trajectory on Matterport3D~\cite{Matterport3D}, compared with baselines. The input image (bottom left) corresponds to the red box location in the panorama. Our reconstructed scene with the overlaid trajectory is shown (bottom right); the white arrow marks the start, and color encodes distance from the input.}
    \label{fig:baseline-cmps}
\Description{A visual baseline comparison for the final video frame generation on Matterport3D}
    \vspace{-1.0em}
\end{figure}

We selected recent baseline models for novel view synthesis from a single image, including ViewCrafter \cite{yu2024viewcrafter}, VistaDream \cite{wang2024vistadream}, and PhotoNVS \cite{yu2023long}, and evaluated them on the RealEstate10K \cite{zhou2018stereo} and Matterport3D \cite{Matterport3D} datasets. Additionally, we tested Wonderjourney \cite{yu2024wonderjourney}, which leverages large language models (LLM) and vision-language models (VLM) to progressively hallucinate video sequences. However, Wonderjourney struggles to maintain view consistency due to its reliance on rich text-based prompts, so its results are not included in the paper.

\begin{figure}[!thbp]
    \vspace{-1.0em}
    \centering
    \includegraphics[width=\columnwidth, trim=0.0cm 0.4cm 0cm 0cm, clip]{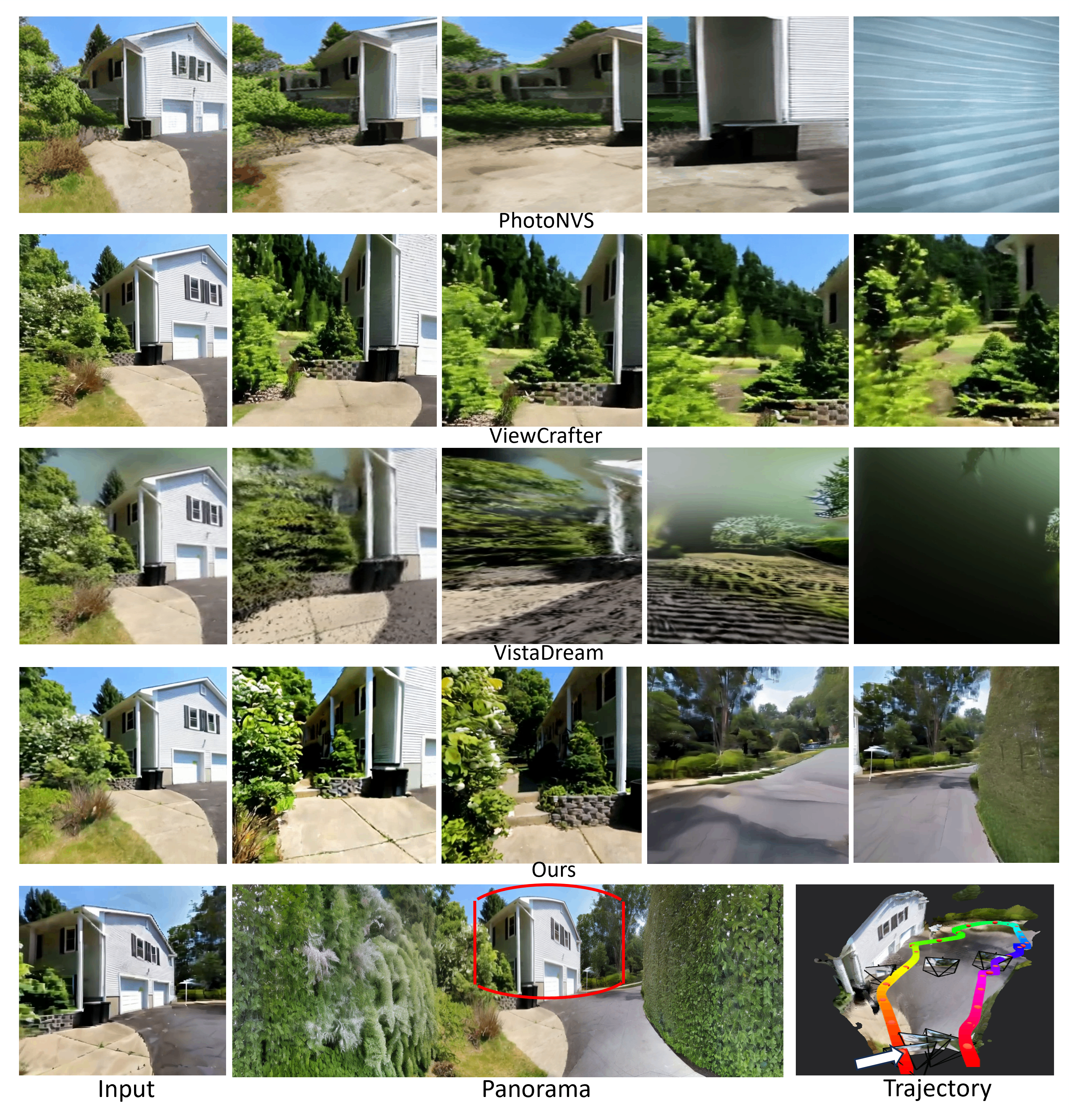}
    \vspace{-2.2em}
    \caption{Comparison of outdoor frames with baselines on RealEstate10K~\cite{zhou2018stereo} along our predefined walk trajectory. The input image (bottom left) matches the red-marked location in the panorama. The reconstructed scene with overlaid trajectory (bottom right) shows the starting point (white arrow) and color-coded distance from the input.}
    \label{fig:baseline-rs-cmps}
    \vspace{-0.8em}
\Description{A visual baseline comparison for the final video frame generation on RealEstate10K.}
\end{figure}

% We present the qualitative evaluation results for RealEstate10K \cite{zhou2018stereo} (Figure \ref{fig:baseline-cmps}) and quantitative results (Table \ref{tab:cmps-matterport}) on both the Matterport3D \cite{Matterport3D} and RealEstate10K \cite{zhou2018stereo} datasets. As illustrated in Figure \ref{fig:baseline-cmps}, our diffusion model generates view-consistent frames closely aligned with ground truth (GT) even beyond 200 frames, whereas all baseline models struggle to maintain plausible views over long sequences. For example, ViewCrafter \cite{yu2024viewcrafter} fails to predict lighting conditions after 164 frames correctly, \emph{e.g.}, as seen in the right part of the second row, and severe geometry distortion, such as shifting sofa positions, becomes evident in the rightmost column. Our model consistently outperforms all baselines across metrics on both datasets (as shown in Table \ref{tab:cmps-matterport}). However, the performance for all models declines on Matterport3D compared to RealEstate10K, likely due to the noisy GT images with incomplete reconstructions and came pose of in-place rotations that disrupt the epipolar condition. Lastly, Figure \ref{fig:long-range-plot} demonstrates the stability of our generated frames in terms of FVD (left y-axis in solid lines) and mTSED (right y-axis in dashed lines) as the frame count increases, in contrast to VistaDream (VD), ViewCrafter (VC), and PhotoNVS (PhoNVS), where frame inconsistency scales dramatically over time. Additional qualitative results for both datasets are provided in the supplementary part.

We present qualitative results on an outdoor scene from the RealEstate10K dataset (Figure~\ref{fig:baseline-rs-cmps}) and two indoor scenes from Matterport3D (Figure~\ref{fig:baseline-cmps}), along with quantitative results in Table~\ref{tab:cmps-matterport} on both Matterport3D and RealEstate10K. As shown in Figures~\ref{fig:baseline-cmps} and~\ref{fig:baseline-rs-cmps}, our diffusion model generates view-consistent frames aligned with the loop walk trajectory in indoor and outdoor settings. In contrast, baseline models fail to maintain view or scene consistency over long sequences. Note that in Figures~\ref{fig:baseline-cmps} and~\ref{fig:baseline-rs-cmps}, the first generated frame in each row does not directly correspond to the input of each method, as the starting point of the walk in a scene may differ. Our model maintains strict scene and viewpoint consistency throughout the entire loop. At the same time, \textit{ViewCrafter} exhibits flipped scenes when returning from the reverse direction, evident in the chandelier and the table with chairs. \textit{VistaDream} often misses regions when the view extends far from the input, and although \textit{PhotoNVS} preserves view consistency, it struggles with alignment over long-range motions. We overcome these issues by conditioning the panoramic scene and anchor keyframes, enabling consistent scene structure and coherent novel views.

\begin{figure}[!tbp]
        \centering
        \vspace{0.0em}
       \includegraphics[width=\linewidth]{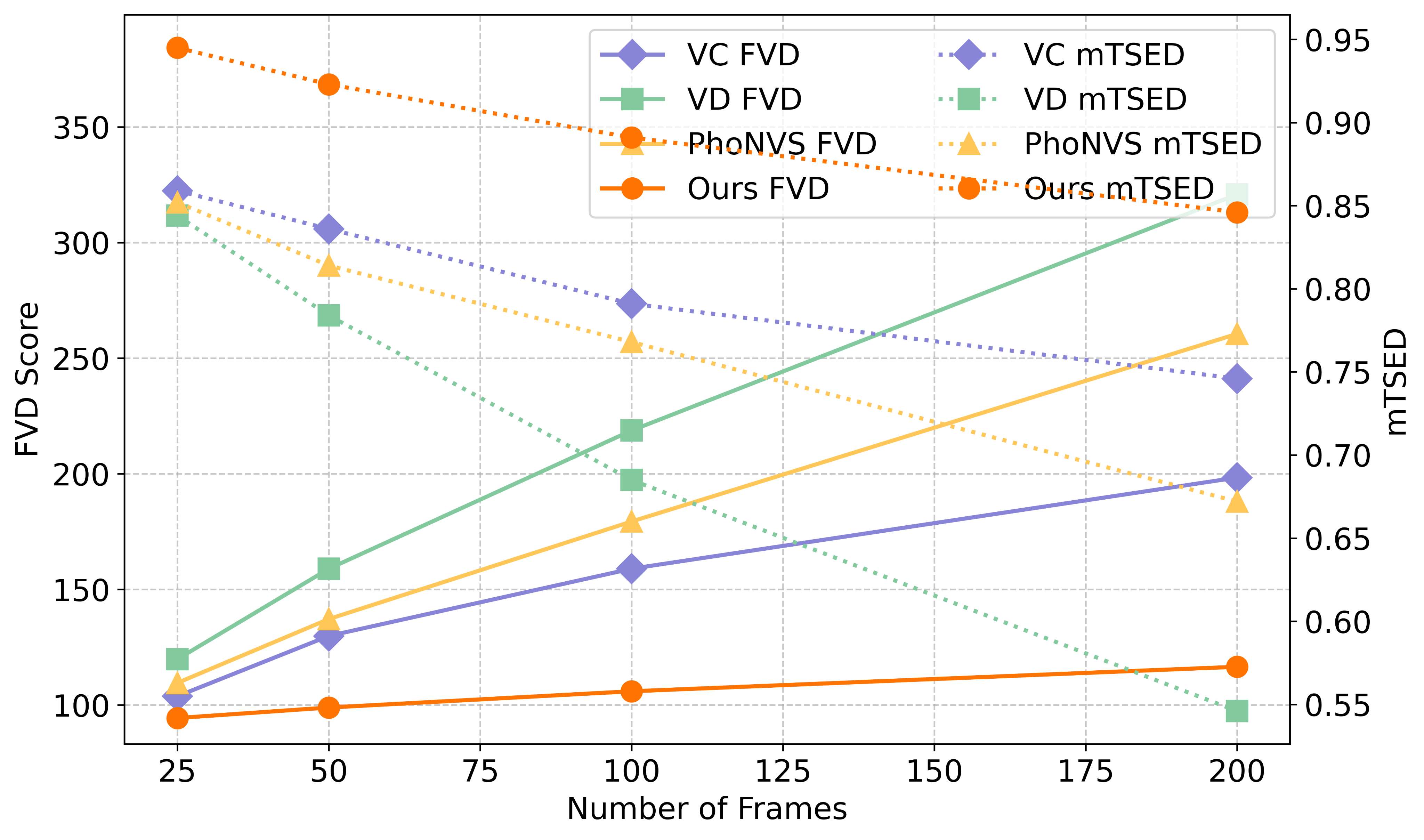}
    \vspace{-2.4em}
    \caption{Comparison of FVD and mTSED frame pair match percentage scores across different numbers of generated frames for our model and baseline methods on RealEstate10K.}
    \label{fig:long-range-plot}
    \vspace{-1.0em}
\Description{FVD and mTSED plot.}
\end{figure}

 %For example, ViewCrafter \cite{yu2024viewcrafter} fails to predict lighting conditions after 164 frames correctly, \emph{e.g.}, as seen in the right part of the second row, and severe geometry distortion, such as shifting sofa positions, becomes evident in the rightmost column. 
Our model consistently outperforms all baselines across metrics on both datasets (as shown in Table \ref{tab:cmps-matterport}). Figure \ref{fig:long-range-plot} demonstrates the stability of our generated frames in terms of FVD (left y-axis in solid lines) and mTSED (right y-axis in dashed lines) as the frame count increases, in contrast to VistaDream (VD), ViewCrafter (VC), and PhotoNVS (PhoNVS), where frame inconsistency scales dramatically over time. Additional qualitative results for both datasets are provided in the supplementary material.

%Wonderjourney
%using video diffusion and 3D Gaussian Splatting (GS) iteratively to generate consistent videos, \cite{yu2024wonderjourney}, leveraging large language models (LLM) to hallucinate the text prompt progressively to guide video generation, 
%generating video from LLM description and 3D GS to generate video in two stages. We also implemented text-based SceneScape \cite{fridman2024scenescape} and PhotoNVS \cite{yu2023long} using VQ-VAE, but we found the generated quality either too saturated or took too long to generate a video via a frame-to-frame manner; thus, we chose not to present these model results in the paper. %We also use Persistent Nature \cite{chai2023persistent} for out-of-distribution landscape video generation comparisons. 
% For baseline evaluation, and Table \ref{tab:cmps-realestate} provide the quantitative comparison results on ScanNet \cite{dai2017scannet} and RealEstate10K \cite{zhou2018stereo} respectively. Our model performance dominates across all the metrics consistently in both short and long-range view synthesis. PhotoNVS is the second-best model on  RealEstate10K, and the GAN-based SE3DS is the second-best on ScanNet, particularly in the long range. The pure inpainting-based method SD-Inpainting does not achieve satisfying results, which requires a precise depth map to warp the source image, and thus cannot extrapolate well farther from the input view. The Look Out works well in near-view generation; however, its fixed autoregressive window limits scalability, resulting in diminished performance in long-range generation.

In terms of efficiency as measured by the average video generation time for 100 frames, PhotoNVS takes 13,800s $\gg$ ViewCrafter at 1187.4s $>$ Vistadream at 683.8s  $>$ Ours at 654.9s (Panorama generation $\approx$ 34s + video diffusion $\approx$ 620.9s).  
 %less than half an hour.
%Inference time comparison.............
\begin{table}[!thbp]
% \vspace{-0.4em}
\centering
    \caption{Comparison results on test scenes of Matterport3D \cite{Matterport3D} (upper part) and RealEstate10K \cite{zhou2018stereo} (lower part).
     \colorbox{orange!50}{\textbf{\underline{Orange}}} indicates the best and \colorbox{pink!80}{\textbf{pink}} represents the second best.}
     \vspace{-0.8em}
    \label{tab:cmps-matterport}
\begin{adjustbox}{width=\columnwidth}    
%\begin{footnotesize}
% %\resizebox{\linewidth}{!}{
     \begin{tabular}{l c c c c  c c c }
%     \centering
        \toprule %\multirow{1}{*}{}
        
       % & \multicolumn{4}{c}{Short Range} & \multicolumn{4}{c}{Long Range}\\
       % \cmidrule(lr){2-5} \cmidrule(lr){6-9} %\cmidrule{10-10} 
       %   %& KID $\downarrow$
        Model & LPIPS $\downarrow$ & PSNR $\uparrow$ & SSIM $\uparrow$ & FID $\downarrow$  & FVD $\downarrow$ & mTSED $\uparrow$ \\
        \hline            
        % PhotoNVS &0 &0.0 & 0.0 & 0.0 & 0.0    \\%& 15\%
        % \hline            
        % MVDiffusion \cite{tang2023MVDiffusion} & 0.593 & 9.852 & 0.391 & 298.358 & 386.401  & 0.483 \\
        % \hline            
        % MultiDiff \cite{muller2024multidiff} &00 & 00 & 0.0 & 0.0 & 0.0    \\
        \hline            
        ViewCrafter \cite{yu2024viewcrafter} & \colorbox{pink!80}{\textbf{0.555}} & 10.899 & 0.477 & \colorbox{pink!80}{\textbf{252.684}} & 346.728 & 0.483  \\%& 1.918 \\
        \hline
        VistaDream \cite{wang2024vistadream} &0.574 & 10.109 & 0.358 & 267.253  & \colorbox{pink!80}{\textbf{320.813}} & \colorbox{pink!80}{\textbf{0.549}}
           \\ %& 2.668\%  \\
       %& 0.168 &  0.126 &  0.090 &  0.154 & $\mathbf{0.036}$\\
       \hline
       PhotoNVS \cite{yu2023long}  &0.560 & \colorbox{pink!80}{\textbf{12.211}} & \colorbox{pink!80}{\textbf{0.483}} & 281.069  & 339.680 &  0.508 \\ %6.538 

       \hline
        Ours &  \colorbox{orange!50}{\textbf{\underline{0.524}}}  &   \colorbox{orange!50}{\textbf{\underline{14.416}}} &  \colorbox{orange!50}{\textbf{\underline{0.510}}} &  \colorbox{orange!50}{\textbf{\underline{199.870}}} &  \colorbox{orange!50}{\textbf{\underline{210.352}}} & \colorbox{orange!50}{\textbf{\underline{0.726}}} \\ % 1.842& ----\\
        \hline
        \hline            
        ViewCrafter \cite{yu2024viewcrafter} &0.456 & 11.373 & 0.462 & \colorbox{pink!80}{\textbf{150.217}} & \colorbox{pink!80}{\textbf{198.269}}& \colorbox{pink!80}{\textbf{0.740}} \\%& 15\%  2.715\\
        \hline
        VistaDream \cite{wang2024vistadream} &0.578 & 9.929 & 0.386 & 234.762  & 326.372 & 0.542
           \\ %& 25\%  \\
       %& 0.168 &  0.126 &  0.090 &  0.154 & $\mathbf{0.036}$\\
       \hline
       PhotoNVS \cite{yu2023long}  & \colorbox{pink!80}{\textbf{0.428}} & \colorbox{pink!80}{\textbf{12.976}} & \colorbox{pink!80}{\textbf{0.485}} & 229.188 & 265.835 & 0.637 \\ %3.776

       \hline
        Ours &  \colorbox{orange!50}{\textbf{\underline{0.218}}} &  \colorbox{orange!50}{\textbf{\underline{15.715}}} &  \colorbox{orange!50}{\textbf{\underline{0.549}}} &   \colorbox{orange!50}{\textbf{\underline{78.858}}}  & \colorbox{orange!50}{\textbf{\underline{113.561}}} & \colorbox{orange!50}{\textbf{\underline{0.835}}}\\ %& 21.164 ----\\
        \bottomrule
    \end{tabular}
%}
%\end{footnotesize}
\end{adjustbox}
\vspace{-0.6em}
\end{table}

\subsection{Ablation Study}
\label{sec:ablation}
\begin{figure}[!thbp]
    \vspace{-1.0em}
    \centering    \includegraphics[width=\columnwidth]{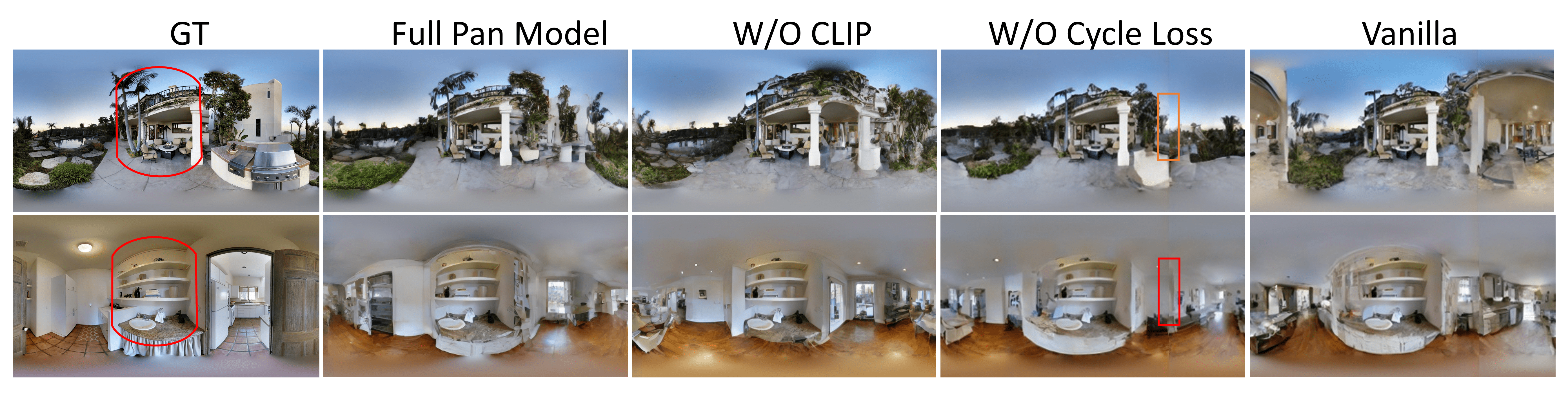} 
    \vspace{-2.4em}
    \caption{Ablation of panorama inference sampling. Red in GT marks the input mask region; the red box highlights view misalignment.}
    \label{fig:ablation-pano}
    \vspace{-0.6em}
\Description{Panorama ablation.}
\end{figure}

We first present visual comparison results in Figure \ref{fig:ablation-pano} for our panorama outpainting diffusion model, including variants without CLIP conditioning, without cycle consistency loss during sampling, and the baseline Diffusion Transformer model. It is evident that incorporating CLIP embedding preserves scene details, while cycle consistency loss is essential for maintaining coherence across the entire panorama, as highlighted by the red gaps in the second-to-last column of Figure \ref{fig:ablation-pano} without cycle loss. The same perspective view input, mapped onto the panorama mask, is used for all ablation tests on each scene. The corresponding metric results in Table \ref{tab:ablation-pano} show a sharp performance drop when either CLIP or cycle consistency loss is removed.

\begin{table}[!thb]
\vspace{-0.6em}
\centering
    \caption{Ablation study results for panorama diffusion.
     \colorbox{orange!50}{\textbf{\underline{Orange}}} indicates the best.}
     \vspace{-0.8em}
    \label{tab:ablation-pano}
\begin{adjustbox}{width=\columnwidth}    
%\begin{footnotesize}
% %\resizebox{\linewidth}{!}{
     \begin{tabular}{l c c c c  c   }
%     \centering
        \toprule %\multirow{1}{*}{}
        
       % & \multicolumn{4}{c}{Short Range} & \multicolumn{4}{c}{Long Range}\\
       % \cmidrule(lr){2-5} \cmidrule(lr){6-9} %\cmidrule{10-10} 
       %   %& KID $\downarrow$
        Model & LPIPS $\downarrow$ & PSNR $\uparrow$ & SSIM $\uparrow$ & FID $\downarrow$ & IS $\uparrow$ \\ %& A/B $\uparrow$ \\
        \hline 
        \hline
        Vanilla & 0.83 & 8.57 & 0.24 & 64.18 & 1.63\\%& 15\%
        \hline            
        Pano w/o Cycle Loss &0.61 & 12.62 & 0.39 & 55.34   & 2.88    \\%& 15\% \\
        \hline
        Pano w/o CLIP Condition & 0.53& 13.44 & 0.46 &  54.69 &  3.12
           \\ %& 25\%  \\
       %& 0.168 &  0.126 &  0.090 &  0.154 & $\mathbf{0.036}$\\
       \hline
       Full Pano Diffusion & \cellcolor{orange!50} 0.49 & \cellcolor{orange!50} 14.04& \cellcolor{orange!50}0.49 & \cellcolor{orange!50} 52.51 & \cellcolor{orange!50}3.51  \\ %& ----\\
        \bottomrule
    \end{tabular}
%}
%\end{footnotesize}
\end{adjustbox}
\vspace{-0.6em}
\end{table}
%Panorama diffusion ablation test
%%%%%%%%% albeit with
% 1. Single view image mapped to panorama view for outpainting
% 2. Single view image and depth mapped to a panorama view for outpainting
% 3. Text CLIP embedding condition + panorama outpainting color\&depth
% 4. Panorama outpainting color \& depth + text CLIP embedding + panorama rotate 90-270 sampling
% fixed 100 sampling steps

%1. A baseline test for the video generation metric results, 
% on scannet(or scannet++), RealEstate10K,
% One out-of-distribution visual comparison results on the landscape dataset or the Matterport outdoor.

\begin{figure}[!th]
  \vspace{-0.2em}
    \centering
    \includegraphics[width=\linewidth, trim=5.4cm 2cm 4cm 3cm, clip]{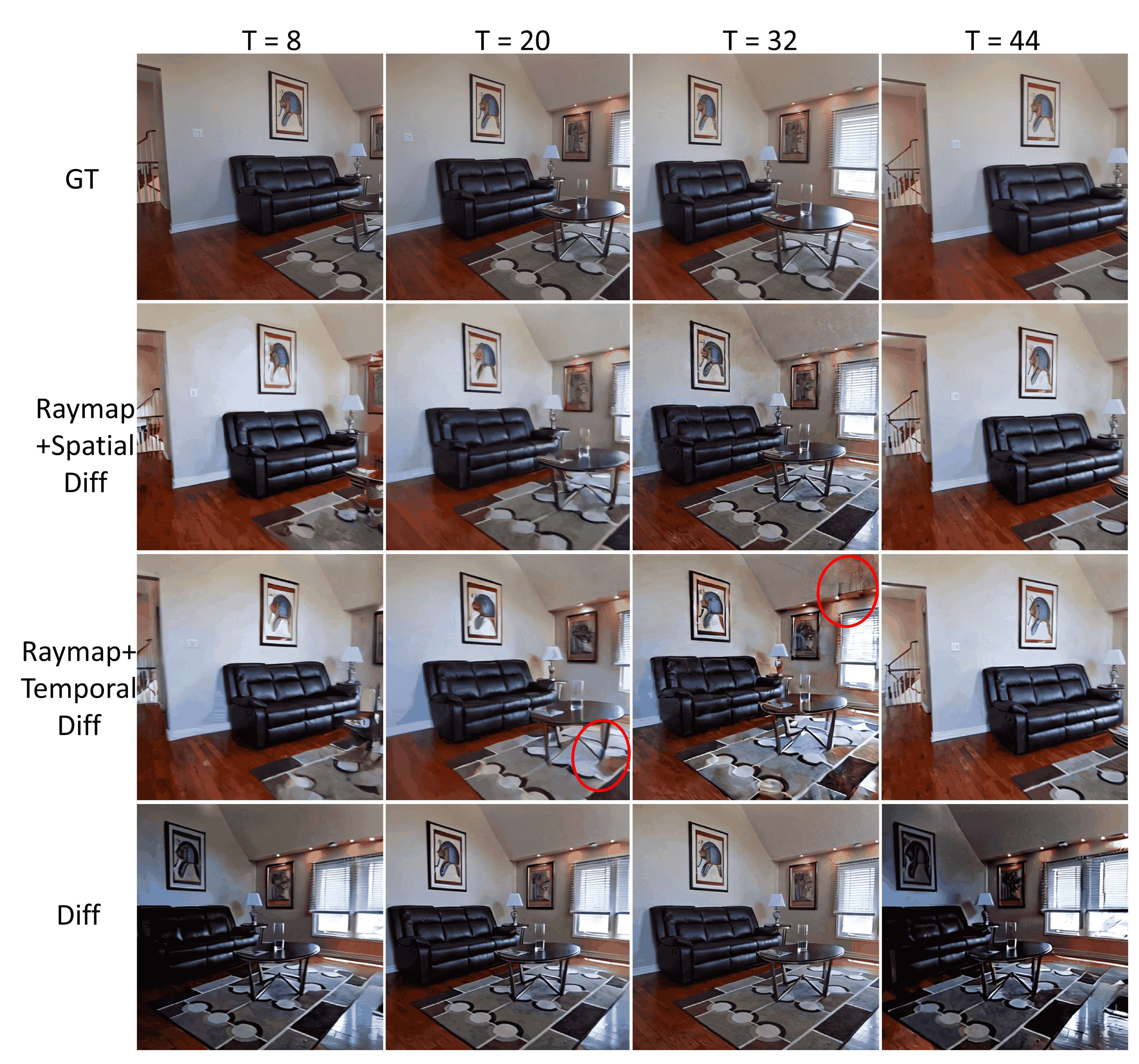}
    \vspace{-1.4em}
    \caption{Spatial diffusion ablation study for novel video view interpolation. The red circle indicates artifacts in raymap-based control with temporal diffusion (third row).  
}
    \label{fig:ablation-bidiff}
    \vspace{-0.4em}
\Description{Video spatial diffusion ablation.}
\end{figure}

% b Visual results comparison
% 2. Ablation study of video diffusion, sampling strategy, and hyperparameters. metric table and visual figure comparisons
% 3. Video diffusion input source comparison, visual comparisons, TSED consistency plot, metric results
% 4. Raymap design comparisons, metric results, and visual figure results (Appendix)
Finally, we conduct an ablation experiment for our video diffusion model. The interpolation model is based on a 3D UNet with layers initialized from SVD-Interpolation \cite{wang2024generative}. The vanilla 3D diffusion for video frame interpolation uses a linear temporal weight (relating each generated view index to the source and target view indices across the entire frame sequence). As shown in the last row of Figure \ref{fig:ablation-bidiff}, this temporal diffusion model does not account for camera motion, resulting in a stationary viewpoint across frames. The concatenated raymaps' conditioning can guide interpolated views along the camera poses, but appearance artifacts are visible, as highlighted by the red circle in the third row. Our proposed spatial weight, which depends on camera poses, improves spatial alignment between the generated view at a given pose and the source or target views. The second row illustrates the more stable view synthesis enabled by spatial weighting compared to the third row. Table \ref{tab:interp-video} shows that our final video diffusion model design outperforms all other designs across metrics, except FID, where temporal diffusion + raymap control achieves a slightly better score. The plot of video consistency scores as a function of spatial temperature hyperparameters $\tau_\mathbf{T}$ and $\tau_\mathbf{q}$ is included in the appendix.

\begin{figure}[!tbp]
    \centering
    \vspace{-0.4em}
    \begin{subfigure}[b]{0.495\columnwidth}
        \centering
        \includegraphics[width=\columnwidth]{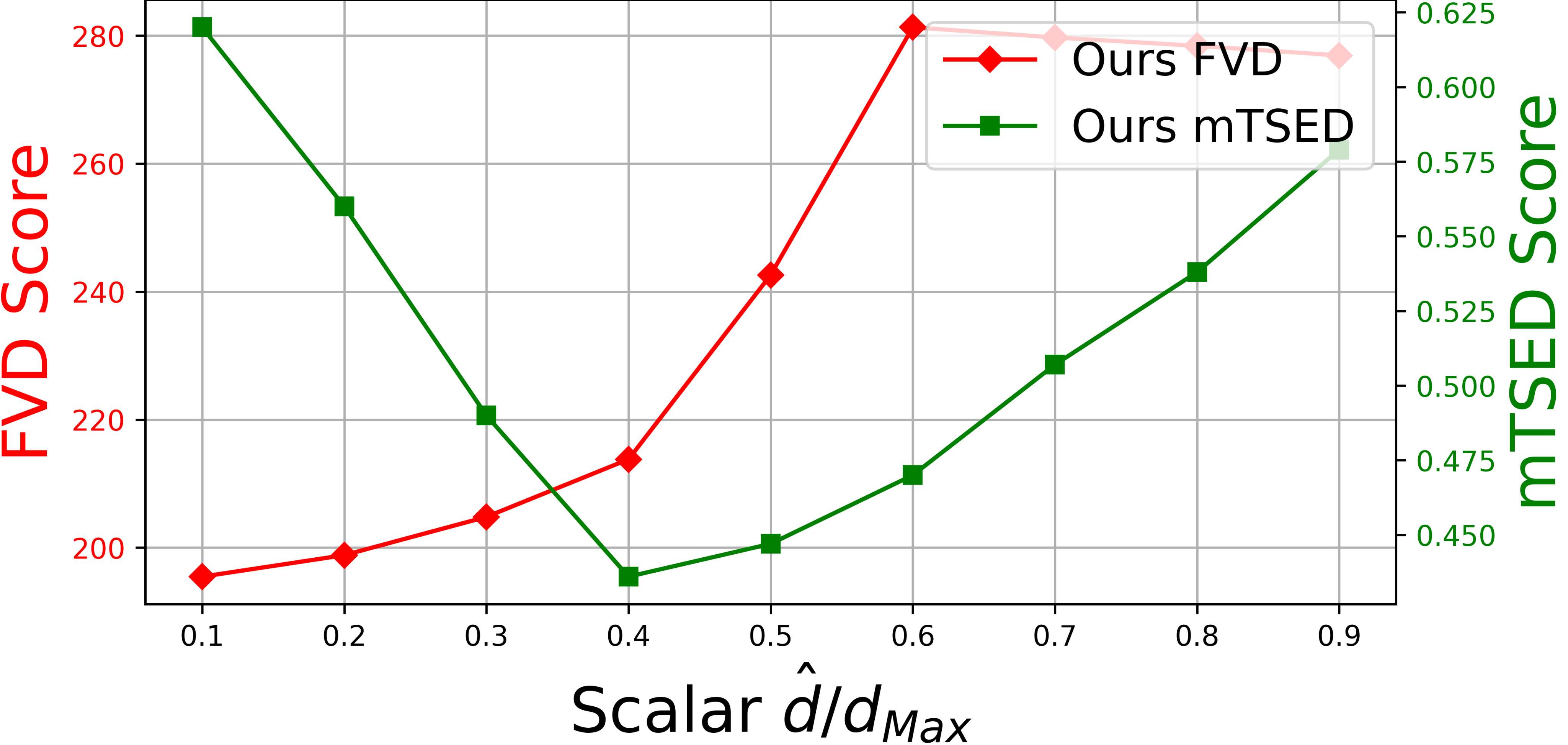}
        % \caption{FVD and mTSED plot of scale $\hat{d}/d_{Max}$}
        \label{fig:fvd-mtsed}
        \vspace{-1.6em}
    \end{subfigure}
    \hfill
    \begin{subfigure}[b]{0.495\columnwidth}
        \centering        \includegraphics[width=\columnwidth]{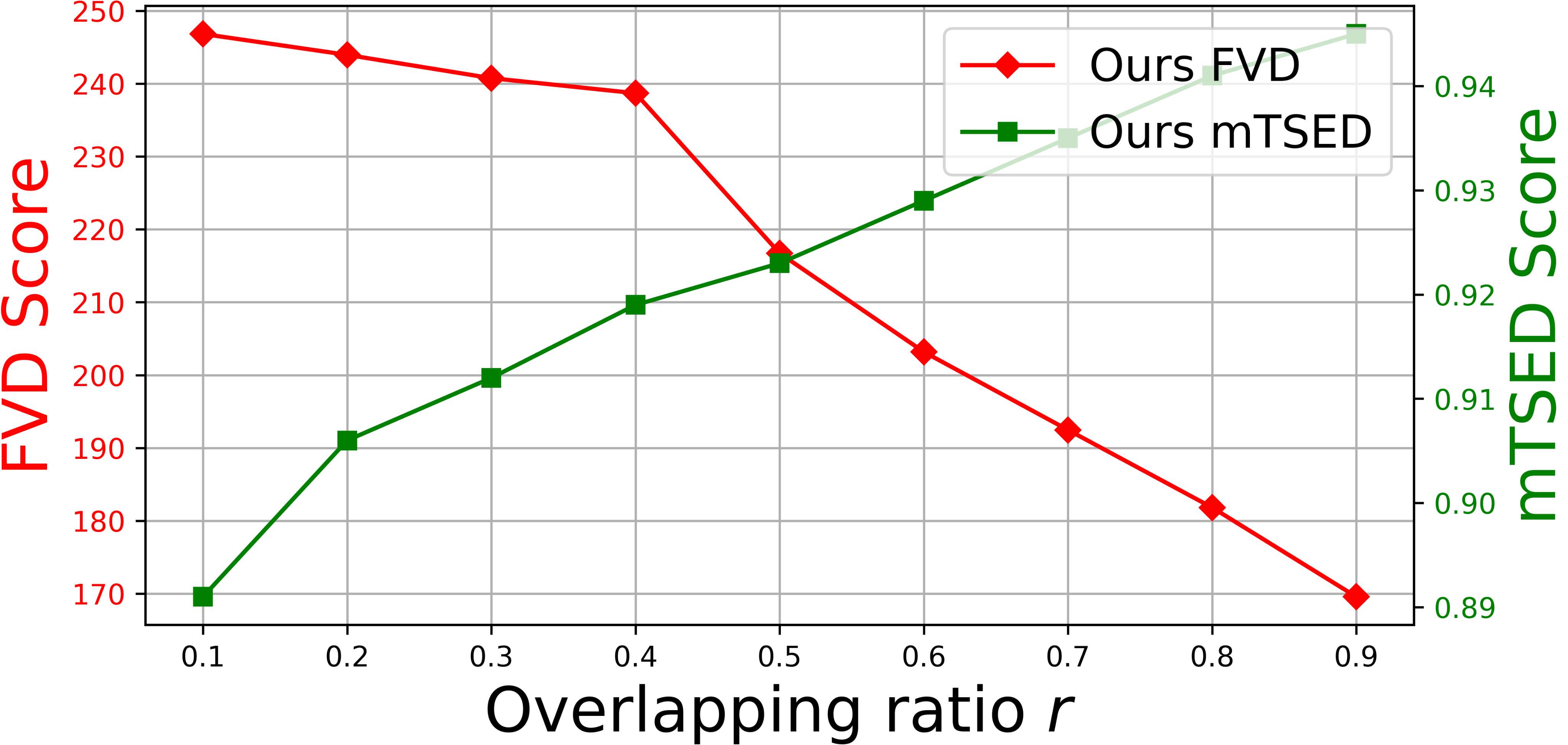}
        % \caption{Video consistency metric w.r.p overlapping ratio $r$}
        \label{fig:overlapping-ratio}
        \vspace{-1.6em}
    \end{subfigure}
    % \vspace{-1.0em}
    \caption{Video consistency evaluation: a) FVD and mTSED vs. scale ratio $\hat{d}/d_{Max}$ under varying source-target overlaps; b) impact of target-to-source scale $r$ on consistency (Matterport3D).}
    \label{fig:keyframe-fact-comparison}
    \vspace{-0.6em}
\Description{Plot of consistency evaluation.}
\end{figure}

We analyze generated video consistency for two perspective keyframe extraction methods from panorama views (Figure \ref{fig:keyframe-fact-comparison}). First, we plot FVD (orange) and mTSED (green) against the scaling factor $z=(\hat{d}/d_{Max})$ from Equation \ref{eq:scale-factor}. Video consistency significantly degrades at a larger resizing ratio due to limited known target view regions. While FVD decreases as the ratio increases, mTSED slightly increases after 0.4 due to more parallel epipolar cues, making it less reliable for measurement in this scenario. Second, we plot the overlapping ratio between neighboring source and target keyframes, showing improved consistency with higher overlap between perspective views extracted from the panorama. 

% \begin{table*}[!th]
% \vspace{-0.6em}
% \centering
% \begin{adjustbox}{width=\linewidth}    
% % %\resizebox{\linewidth}{!}{
%      \begin{tabular}{c c c c}
%      \centering
% % \begin{small}
%        \tiny{Full Model} & \tiny{W/O CLIP} & \tiny{W/O Cycle} & \tiny{Vanilla} 
% % \end{small}
%     \end{tabular}
% %}
% \end{adjustbox}
% \vspace{-3.4em}
% \end{table*} 
\begin{table}[!th]
\vspace{-0.4em}
\centering
    \caption{Ablation study results for view interpolation via video diffusion.
     \colorbox{orange!50}{\textbf{Orange}} indicates the best.} % and \colorbox{pink!80}{\textbf{pink}} represents the second best.
     \vspace{-0.8em}
    \label{tab:interp-video}
\begin{adjustbox}{width=\columnwidth}    
%\begin{footnotesize}
% %\resizebox{\linewidth}{!}{
     \begin{tabular}{l c c c c c c }
%     \centering
        \toprule %\multirow{1}{*}{}
        
       % & \multicolumn{4}{c}{Short Range} & \multicolumn{4}{c}{Long Range}\\
       % \cmidrule(lr){2-5} \cmidrule(lr){6-9} %\cmidrule{10-10} 
       %   %& KID $\downarrow$
        Model & LPIPS $\downarrow$ & PSNR $\uparrow$ & SSIM $\uparrow$ & FID $\downarrow$ & FVD $\downarrow$ & mTSED $\uparrow$\\ %& A/B $\uparrow$ \\
        \hline 
        \hline
        Temporal Diff & 0.392 & 12.502 & 0.458 & 123.694 & 124.53 &  0.75 \\%& 15\% 3.332 1.789
        \hline
        Temporal Diff + Raymap & 0.226& 15.427& 0.540 & \cellcolor{orange!50} 76.902 & 81.16 & 0.83  \\ %& 25\%  3.285\\
       %& 0.168 &  0.126 &  0.090 &  0.154 & $\mathbf{0.036}$\\
       \hline
      Spatial DIff + Raymap & \cellcolor{orange!50} 0.218 & \cellcolor{orange!50} 15.715 & \cellcolor{orange!50} 0.549 &  78.858  & \cellcolor{orange!50} 78.09 &\cellcolor{orange!50} 0.88\\ %& (1.164 ----\\
        \bottomrule
    \end{tabular}
%}
%\end{footnotesize}
\end{adjustbox}
\vspace{-0.8em}
\end{table}

\section{Conclusion}
%We propose a two-stage video generation framework, given a single image, to generate consistent video with camera motion control. 
We present a novel framework for long-term, consistent novel view synthesis from a single image by decomposing the task into 360° scene extrapolation and trajectory-aware view interpolation. Conditioned on panoramic keyframes, our method supports coherent video generation along arbitrary paths. Despite promising results, limitations remain. Training and inference speed could benefit from flow-matching techniques. Integrating autonomous trajectory planning would enable topology-aware navigation. While we currently model static scenes, dynamic foreground entities could be incorporated as overlaid layers. Video generation efficiency can also be improved using recent advancements. Moreover, we observed failures under extreme lighting, motion blur, occlusion, cluttered or cases with dynamic objects. 
%which presents an important avenue for future research to achieve even more realistic scene synthesis.
%Experimental results demonstrate that our method consistently outperforms existing approaches in generating long-term coherent and flexible views across complex scenes.

\begin{acks}
The first three authors acknowledge the financial support from the University of Melbourne through the Melbourne Research Scholarship. This research was supported by the University of Melbourne's Research Computing Services and the Petascale Campus initiative. 
\end{acks}
\balance
\bibliographystyle{ACM-Reference-Format}
\bibliography{main}

%%% -*-BibTeX-*-
%%% Do NOT edit. File created by BibTeX with style
%%% ACM-Reference-Format-Journals [18-Jan-2012].

\begin{thebibliography}{72}

%%% ====================================================================
%%% NOTE TO THE USER: you can override these defaults by providing
%%% customized versions of any of these macros before the \bibliography
%%% command.  Each of them MUST provide its own final punctuation,
%%% except for \shownote{}, \showDOI{}, and \showURL{}.  The latter two
%%% do not use final punctuation, in order to avoid confusing it with
%%% the Web address.
%%%
%%% To suppress output of a particular field, define its macro to expand
%%% to an empty string, or better, \unskip, like this:
%%%
%%% \newcommand{\showDOI}[1]{\unskip}   % LaTeX syntax
%%%
%%% \def \showDOI #1{\unskip}           % plain TeX syntax
%%%
%%% ====================================================================

\ifx \showCODEN    \undefined \def \showCODEN     #1{\unskip}     \fi
\ifx \showDOI      \undefined \def \showDOI       #1{#1}\fi
\ifx \showISBNx    \undefined \def \showISBNx     #1{\unskip}     \fi
\ifx \showISBNxiii \undefined \def \showISBNxiii  #1{\unskip}     \fi
\ifx \showISSN     \undefined \def \showISSN      #1{\unskip}     \fi
\ifx \showLCCN     \undefined \def \showLCCN      #1{\unskip}     \fi
\ifx \shownote     \undefined \def \shownote      #1{#1}          \fi
\ifx \showarticletitle \undefined \def \showarticletitle #1{#1}   \fi
\ifx \showURL      \undefined \def \showURL       {\relax}        \fi
% The following commands are used for tagged output and should be
% invisible to TeX
\providecommand\bibfield[2]{#2}
\providecommand\bibinfo[2]{#2}
\providecommand\natexlab[1]{#1}
\providecommand\showeprint[2][]{arXiv:#2}

\bibitem[Bahmani et~al\mbox{.}(2024)]%
        {bahmani2024vd3d}
\bibfield{author}{\bibinfo{person}{Sherwin Bahmani}, \bibinfo{person}{Ivan Skorokhodov}, \bibinfo{person}{Aliaksandr Siarohin}, \bibinfo{person}{Willi Menapace}, \bibinfo{person}{Guocheng Qian}, \bibinfo{person}{Michael Vasilkovsky}, \bibinfo{person}{Hsin-Ying Lee}, \bibinfo{person}{Chaoyang Wang}, \bibinfo{person}{Jiaxu Zou}, \bibinfo{person}{Andrea Tagliasacchi}, {et~al\mbox{.}}} \bibinfo{year}{2024}\natexlab{}.
\newblock \showarticletitle{VD{3D}: Taming Large Video Diffusion Transformers for {3D} Camera Control}.
\newblock \bibinfo{journal}{\emph{arXiv preprint arXiv:2407.12781}} (\bibinfo{year}{2024}).
\newblock


\bibitem[Bao et~al\mbox{.}(2019)]%
        {bao2019depth}
\bibfield{author}{\bibinfo{person}{Wenbo Bao}, \bibinfo{person}{Wei-Sheng Lai}, \bibinfo{person}{Chao Ma}, \bibinfo{person}{Xiaoyun Zhang}, \bibinfo{person}{Zhiyong Gao}, {and} \bibinfo{person}{Ming-Hsuan Yang}.} \bibinfo{year}{2019}\natexlab{}.
\newblock \showarticletitle{Depth-aware video frame interpolation}. In \bibinfo{booktitle}{\emph{Proceedings of the IEEE/CVF conference on computer vision and pattern recognition}}. \bibinfo{pages}{3703--3712}.
\newblock


\bibitem[Bar-Tal et~al\mbox{.}(2024)]%
        {bar2024lumiere}
\bibfield{author}{\bibinfo{person}{Omer Bar-Tal}, \bibinfo{person}{Hila Chefer}, \bibinfo{person}{Omer Tov}, \bibinfo{person}{Charles Herrmann}, \bibinfo{person}{Roni Paiss}, \bibinfo{person}{Shiran Zada}, \bibinfo{person}{Ariel Ephrat}, \bibinfo{person}{Junhwa Hur}, \bibinfo{person}{Yuanzhen Li}, \bibinfo{person}{Tomer Michaeli}, {et~al\mbox{.}}} \bibinfo{year}{2024}\natexlab{}.
\newblock \showarticletitle{Lumiere: A space-time diffusion model for video generation}.
\newblock \bibinfo{journal}{\emph{arXiv preprint arXiv:2401.12945}} (\bibinfo{year}{2024}).
\newblock


\bibitem[Chai et~al\mbox{.}(2023)]%
        {chai2023persistent}
\bibfield{author}{\bibinfo{person}{Lucy Chai}, \bibinfo{person}{Richard Tucker}, \bibinfo{person}{Zhengqi Li}, \bibinfo{person}{Phillip Isola}, {and} \bibinfo{person}{Noah Snavely}.} \bibinfo{year}{2023}\natexlab{}.
\newblock \showarticletitle{Persistent nature: A generative model of unbounded {3D} worlds}. In \bibinfo{booktitle}{\emph{Proceedings of the IEEE/CVF conference on computer vision and pattern recognition}}. \bibinfo{pages}{20863--20874}.
\newblock


\bibitem[Chang et~al\mbox{.}(2017)]%
        {Matterport3D}
\bibfield{author}{\bibinfo{person}{Angel Chang}, \bibinfo{person}{Angela Dai}, \bibinfo{person}{Thomas Funkhouser}, \bibinfo{person}{Maciej Halber}, \bibinfo{person}{Matthias Niessner}, \bibinfo{person}{Manolis Savva}, \bibinfo{person}{Shuran Song}, \bibinfo{person}{Andy Zeng}, {and} \bibinfo{person}{Yinda Zhang}.} \bibinfo{year}{2017}\natexlab{}.
\newblock \showarticletitle{Matterport{3D}: Learning from RGB-D Data in Indoor Environments}.
\newblock \bibinfo{journal}{\emph{International Conference on {3D} Vision ({3D}V)}} (\bibinfo{year}{2017}).
\newblock


\bibitem[Chen et~al\mbox{.}(2024b)]%
        {chen2024videocrafter2}
\bibfield{author}{\bibinfo{person}{Haoxin Chen}, \bibinfo{person}{Yong Zhang}, \bibinfo{person}{Xiaodong Cun}, \bibinfo{person}{Menghan Xia}, \bibinfo{person}{Xintao Wang}, \bibinfo{person}{Chao Weng}, {and} \bibinfo{person}{Ying Shan}.} \bibinfo{year}{2024}\natexlab{b}.
\newblock \bibinfo{title}{VideoCrafter2: Overcoming Data Limitations for High-Quality Video Diffusion Models}.
\newblock
\newblock
\showeprint[arxiv]{2401.09047}~[cs.CV]


\bibitem[Chen et~al\mbox{.}(2024a)]%
        {chen2024v3d}
\bibfield{author}{\bibinfo{person}{Zilong Chen}, \bibinfo{person}{Yikai Wang}, \bibinfo{person}{Feng Wang}, \bibinfo{person}{Zhengyi Wang}, {and} \bibinfo{person}{Huaping Liu}.} \bibinfo{year}{2024}\natexlab{a}.
\newblock \showarticletitle{V3d: Video diffusion models are effective 3d generators}.
\newblock \bibinfo{journal}{\emph{arXiv preprint arXiv:2403.06738}} (\bibinfo{year}{2024}).
\newblock


\bibitem[Danier et~al\mbox{.}(2023)]%
        {danier2023ldmvfi}
\bibfield{author}{\bibinfo{person}{Duolikun Danier}, \bibinfo{person}{Fan Zhang}, {and} \bibinfo{person}{David Bull}.} \bibinfo{year}{2023}\natexlab{}.
\newblock \showarticletitle{LDMVFI: Video Frame Interpolation with Latent Diffusion Models}.
\newblock \bibinfo{journal}{\emph{arXiv preprint arXiv:2303.09508}} (\bibinfo{year}{2023}).
\newblock


\bibitem[Esser et~al\mbox{.}(2023)]%
        {esser2023structure}
\bibfield{author}{\bibinfo{person}{Patrick Esser}, \bibinfo{person}{Johnathan Chiu}, \bibinfo{person}{Parmida Atighehchian}, \bibinfo{person}{Jonathan Granskog}, {and} \bibinfo{person}{Anastasis Germanidis}.} \bibinfo{year}{2023}\natexlab{}.
\newblock \showarticletitle{Structure and content-guided video synthesis with diffusion models}. In \bibinfo{booktitle}{\emph{Proceedings of the IEEE/CVF International Conference on Computer Vision}}. \bibinfo{pages}{7346--7356}.
\newblock


\bibitem[Feng et~al\mbox{.}(2023)]%
        {feng2023diffusion360}
\bibfield{author}{\bibinfo{person}{Mengyang Feng}, \bibinfo{person}{Jinlin Liu}, \bibinfo{person}{Miaomiao Cui}, {and} \bibinfo{person}{Xuansong Xie}.} \bibinfo{year}{2023}\natexlab{}.
\newblock \showarticletitle{Diffusion360: Seamless 360 degree panoramic image generation based on diffusion models}.
\newblock \bibinfo{journal}{\emph{arXiv preprint arXiv:2311.13141}} (\bibinfo{year}{2023}).
\newblock


\bibitem[Fridman et~al\mbox{.}(2024)]%
        {fridman2024scenescape}
\bibfield{author}{\bibinfo{person}{Rafail Fridman}, \bibinfo{person}{Amit Abecasis}, \bibinfo{person}{Yoni Kasten}, {and} \bibinfo{person}{Tali Dekel}.} \bibinfo{year}{2024}\natexlab{}.
\newblock \showarticletitle{Scenescape: Text-driven consistent scene generation}.
\newblock \bibinfo{journal}{\emph{Advances in Neural Information Processing Systems}}  \bibinfo{volume}{36} (\bibinfo{year}{2024}).
\newblock


\bibitem[Gao et~al\mbox{.}(2024)]%
        {gao2024cat3d}
\bibfield{author}{\bibinfo{person}{Ruiqi Gao}, \bibinfo{person}{Aleksander Holynski}, \bibinfo{person}{Philipp Henzler}, \bibinfo{person}{Arthur Brussee}, \bibinfo{person}{Ricardo Martin-Brualla}, \bibinfo{person}{Pratul Srinivasan}, \bibinfo{person}{Jonathan~T Barron}, {and} \bibinfo{person}{Ben Poole}.} \bibinfo{year}{2024}\natexlab{}.
\newblock \showarticletitle{CAT{3D}: Create Anything in {3D} with Multi-View Diffusion Models}.
\newblock \bibinfo{journal}{\emph{arXiv preprint arXiv:2405.10314}} (\bibinfo{year}{2024}).
\newblock


\bibitem[He et~al\mbox{.}(2025)]%
        {he2024cameractrl}
\bibfield{author}{\bibinfo{person}{Hao He}, \bibinfo{person}{Yinghao Xu}, \bibinfo{person}{Yuwei Guo}, \bibinfo{person}{Gordon Wetzstein}, \bibinfo{person}{Bo Dai}, \bibinfo{person}{Hongsheng Li}, {and} \bibinfo{person}{Ceyuan Yang}.} \bibinfo{year}{2025}\natexlab{}.
\newblock \showarticletitle{Cameractrl: Enabling camera control for text-to-video generation}.
\newblock  (\bibinfo{year}{2025}).
\newblock


\bibitem[He et~al\mbox{.}(2016)]%
        {he2016deep}
\bibfield{author}{\bibinfo{person}{Kaiming He}, \bibinfo{person}{Xiangyu Zhang}, \bibinfo{person}{Shaoqing Ren}, {and} \bibinfo{person}{Jian Sun}.} \bibinfo{year}{2016}\natexlab{}.
\newblock \showarticletitle{Deep residual learning for image recognition}. In \bibinfo{booktitle}{\emph{Proceedings of the IEEE conference on computer vision and pattern recognition}}. \bibinfo{pages}{770--778}.
\newblock


\bibitem[Heusel et~al\mbox{.}(2017)]%
        {heusel2017gans}
\bibfield{author}{\bibinfo{person}{Martin Heusel}, \bibinfo{person}{Hubert Ramsauer}, \bibinfo{person}{Thomas Unterthiner}, \bibinfo{person}{Bernhard Nessler}, {and} \bibinfo{person}{Sepp Hochreiter}.} \bibinfo{year}{2017}\natexlab{}.
\newblock \showarticletitle{Gans trained by a two time-scale update rule converge to a local nash equilibrium}.
\newblock \bibinfo{journal}{\emph{Advances in neural information processing systems}}  \bibinfo{volume}{30} (\bibinfo{year}{2017}).
\newblock


\bibitem[Ho et~al\mbox{.}(2022a)]%
        {ho2022imagen}
\bibfield{author}{\bibinfo{person}{Jonathan Ho}, \bibinfo{person}{William Chan}, \bibinfo{person}{Chitwan Saharia}, \bibinfo{person}{Jay Whang}, \bibinfo{person}{Ruiqi Gao}, \bibinfo{person}{Alexey Gritsenko}, \bibinfo{person}{Diederik~P Kingma}, \bibinfo{person}{Ben Poole}, \bibinfo{person}{Mohammad Norouzi}, \bibinfo{person}{David~J Fleet}, {et~al\mbox{.}}} \bibinfo{year}{2022}\natexlab{a}.
\newblock \showarticletitle{Imagen video: High definition video generation with diffusion models}.
\newblock \bibinfo{journal}{\emph{arXiv preprint arXiv:2210.02303}} (\bibinfo{year}{2022}).
\newblock


\bibitem[Ho et~al\mbox{.}(2022b)]%
        {ho2022video}
\bibfield{author}{\bibinfo{person}{Jonathan Ho}, \bibinfo{person}{Tim Salimans}, \bibinfo{person}{Alexey Gritsenko}, \bibinfo{person}{William Chan}, \bibinfo{person}{Mohammad Norouzi}, {and} \bibinfo{person}{David~J Fleet}.} \bibinfo{year}{2022}\natexlab{b}.
\newblock \showarticletitle{Video diffusion models}.
\newblock \bibinfo{journal}{\emph{Advances in Neural Information Processing Systems}}  \bibinfo{volume}{35} (\bibinfo{year}{2022}), \bibinfo{pages}{8633--8646}.
\newblock


\bibitem[Huang et~al\mbox{.}(2022)]%
        {huang2022real}
\bibfield{author}{\bibinfo{person}{Zhewei Huang}, \bibinfo{person}{Tianyuan Zhang}, \bibinfo{person}{Wen Heng}, \bibinfo{person}{Boxin Shi}, {and} \bibinfo{person}{Shuchang Zhou}.} \bibinfo{year}{2022}\natexlab{}.
\newblock \showarticletitle{Real-time intermediate flow estimation for video frame interpolation}. In \bibinfo{booktitle}{\emph{European Conference on Computer Vision}}. Springer, \bibinfo{pages}{624--642}.
\newblock


\bibitem[Ju et~al\mbox{.}(2024)]%
        {ju2024diffindscene}
\bibfield{author}{\bibinfo{person}{Xiaoliang Ju}, \bibinfo{person}{Zhaoyang Huang}, \bibinfo{person}{Yijin Li}, \bibinfo{person}{Guofeng Zhang}, \bibinfo{person}{Yu Qiao}, {and} \bibinfo{person}{Hongsheng Li}.} \bibinfo{year}{2024}\natexlab{}.
\newblock \showarticletitle{DiffInDScene: Diffusion-based High-Quality {3D} Indoor Scene Generation}. In \bibinfo{booktitle}{\emph{Proceedings of the IEEE/CVF Conference on Computer Vision and Pattern Recognition}}. \bibinfo{pages}{4526--4535}.
\newblock


\bibitem[Kang et~al\mbox{.}(2025)]%
        {kang2025multi}
\bibfield{author}{\bibinfo{person}{Xueyang Kang}, \bibinfo{person}{Zhengkang Xiang}, \bibinfo{person}{Zezheng Zhang}, {and} \bibinfo{person}{Kourosh Khoshelham}.} \bibinfo{year}{2025}\natexlab{}.
\newblock \showarticletitle{Multi-view Geometry-Aware Diffusion Transformer for Indoor Novel View Synthesis}. In \bibinfo{booktitle}{\emph{ICLR 2025 Workshop on Deep Generative Model in Machine Learning: Theory, Principle and Efficacy}}.
\newblock


\bibitem[Khachatryan et~al\mbox{.}(2023)]%
        {khachatryan2023text2video}
\bibfield{author}{\bibinfo{person}{Levon Khachatryan}, \bibinfo{person}{Andranik Movsisyan}, \bibinfo{person}{Vahram Tadevosyan}, \bibinfo{person}{Roberto Henschel}, \bibinfo{person}{Zhangyang Wang}, \bibinfo{person}{Shant Navasardyan}, {and} \bibinfo{person}{Humphrey Shi}.} \bibinfo{year}{2023}\natexlab{}.
\newblock \showarticletitle{Text2video-zero: Text-to-image diffusion models are zero-shot video generators}. In \bibinfo{booktitle}{\emph{Proceedings of the IEEE/CVF International Conference on Computer Vision}}. \bibinfo{pages}{15954--15964}.
\newblock


\bibitem[Kingma and Welling(2013)]%
        {kingma2013auto}
\bibfield{author}{\bibinfo{person}{Diederik~P Kingma} {and} \bibinfo{person}{Max Welling}.} \bibinfo{year}{2013}\natexlab{}.
\newblock \showarticletitle{Auto-encoding variational bayes}.
\newblock \bibinfo{journal}{\emph{arXiv preprint arXiv:1312.6114}} (\bibinfo{year}{2013}).
\newblock


\bibitem[Kuang et~al\mbox{.}(2024)]%
        {kuang2024cvd}
\bibfield{author}{\bibinfo{person}{Zhengfei Kuang}, \bibinfo{person}{Shengqu Cai}, \bibinfo{person}{Hao He}, \bibinfo{person}{Yinghao Xu}, \bibinfo{person}{Hongsheng Li}, \bibinfo{person}{Leonidas Guibas}, {and} \bibinfo{person}{Gordon. Wetzstein}.} \bibinfo{year}{2024}\natexlab{}.
\newblock \showarticletitle{Collaborative Video Diffusion: Consistent Multi-video Generation with Camera Control}. In \bibinfo{booktitle}{\emph{arXiv}}.
\newblock


\bibitem[Kwak et~al\mbox{.}(2024)]%
        {kwak2024vivid}
\bibfield{author}{\bibinfo{person}{Jeong-gi Kwak}, \bibinfo{person}{Erqun Dong}, \bibinfo{person}{Yuhe Jin}, \bibinfo{person}{Hanseok Ko}, \bibinfo{person}{Shweta Mahajan}, {and} \bibinfo{person}{Kwang~Moo Yi}.} \bibinfo{year}{2024}\natexlab{}.
\newblock \showarticletitle{Vivid-1-to-3: Novel view synthesis with video diffusion models}. In \bibinfo{booktitle}{\emph{Proceedings of the IEEE/CVF Conference on Computer Vision and Pattern Recognition}}. \bibinfo{pages}{6775--6785}.
\newblock


\bibitem[Li et~al\mbox{.}(2024)]%
        {minfenli2024GenRC}
\bibfield{author}{\bibinfo{person}{Ming-Feng Li}, \bibinfo{person}{Yueh-Feng Ku}, \bibinfo{person}{Hong-Xuan Yen}, \bibinfo{person}{Yu-Lun~Liu Chi~Liu}, \bibinfo{person}{Albert Y.~C. Chen}, \bibinfo{person}{Cheng-Hao Kuo}, {and} \bibinfo{person}{Min Sun}.} \bibinfo{year}{2024}\natexlab{}.
\newblock \showarticletitle{GenRC: {3D} Indoor Scene Generation from Sparse Image Collections}. In \bibinfo{booktitle}{\emph{ECCV}}.
\newblock


\bibitem[Lin et~al\mbox{.}(2023)]%
        {lin2023magic3d}
\bibfield{author}{\bibinfo{person}{Chen-Hsuan Lin}, \bibinfo{person}{Jun Gao}, \bibinfo{person}{Luming Tang}, \bibinfo{person}{Towaki Takikawa}, \bibinfo{person}{Xiaohui Zeng}, \bibinfo{person}{Xun Huang}, \bibinfo{person}{Karsten Kreis}, \bibinfo{person}{Sanja Fidler}, \bibinfo{person}{Ming-Yu Liu}, {and} \bibinfo{person}{Tsung-Yi Lin}.} \bibinfo{year}{2023}\natexlab{}.
\newblock \showarticletitle{Magic{3D}: High-resolution text-to-3d content creation}. In \bibinfo{booktitle}{\emph{Proceedings of the IEEE/CVF Conference on Computer Vision and Pattern Recognition}}. \bibinfo{pages}{300--309}.
\newblock


\bibitem[Liu et~al\mbox{.}(2024)]%
        {liu2024panofree}
\bibfield{author}{\bibinfo{person}{Aoming Liu}, \bibinfo{person}{Zhong Li}, \bibinfo{person}{Zhang Chen}, \bibinfo{person}{Nannan Li}, \bibinfo{person}{Yi Xu}, {and} \bibinfo{person}{Bryan~A Plummer}.} \bibinfo{year}{2024}\natexlab{}.
\newblock \showarticletitle{PanoFree: Tuning-Free Holistic Multi-view Image Generation with Cross-view Self-Guidance}.
\newblock \bibinfo{journal}{\emph{arXiv preprint arXiv:2408.02157}} (\bibinfo{year}{2024}).
\newblock


\bibitem[Liu et~al\mbox{.}(2019)]%
        {liu2019deep}
\bibfield{author}{\bibinfo{person}{Yu-Lun Liu}, \bibinfo{person}{Yi-Tung Liao}, \bibinfo{person}{Yen-Yu Lin}, {and} \bibinfo{person}{Yung-Yu Chuang}.} \bibinfo{year}{2019}\natexlab{}.
\newblock \showarticletitle{Deep video frame interpolation using cyclic frame generation}. In \bibinfo{booktitle}{\emph{Proceedings of the AAAI Conference on Artificial Intelligence}}, Vol.~\bibinfo{volume}{33}. \bibinfo{pages}{8794--8802}.
\newblock


\bibitem[Liu et~al\mbox{.}(2023)]%
        {Liu2023MeshDiffusion}
\bibfield{author}{\bibinfo{person}{Zhen Liu}, \bibinfo{person}{Yao Feng}, \bibinfo{person}{Michael~J. Black}, \bibinfo{person}{Derek Nowrouzezahrai}, \bibinfo{person}{Liam Paull}, {and} \bibinfo{person}{Weiyang Liu}.} \bibinfo{year}{2023}\natexlab{}.
\newblock \showarticletitle{MeshDiffusion: Score-based Generative {3D} Mesh Modeling}. In \bibinfo{booktitle}{\emph{International Conference on Learning Representations}}.
\newblock
\urldef\tempurl%
\url{https://openreview.net/forum?id=0cpM2ApF9p6}
\showURL{%
\tempurl}


\bibitem[Melas-Kyriazi et~al\mbox{.}(2024)]%
        {melas20243d}
\bibfield{author}{\bibinfo{person}{Luke Melas-Kyriazi}, \bibinfo{person}{Iro Laina}, \bibinfo{person}{Christian Rupprecht}, \bibinfo{person}{Natalia Neverova}, \bibinfo{person}{Andrea Vedaldi}, \bibinfo{person}{Oran Gafni}, {and} \bibinfo{person}{Filippos Kokkinos}.} \bibinfo{year}{2024}\natexlab{}.
\newblock \showarticletitle{Im-3d: Iterative multiview diffusion and reconstruction for high-quality 3d generation}.
\newblock \bibinfo{journal}{\emph{arXiv preprint arXiv:2402.08682}} (\bibinfo{year}{2024}).
\newblock


\bibitem[Meng et~al\mbox{.}(2024)]%
        {meng2024lt3sd}
\bibfield{author}{\bibinfo{person}{Quan Meng}, \bibinfo{person}{Lei Li}, \bibinfo{person}{Matthias Nie{\ss}ner}, {and} \bibinfo{person}{Angela Dai}.} \bibinfo{year}{2024}\natexlab{}.
\newblock \showarticletitle{LT3SD: Latent Trees for {3D} Scene Diffusion}.
\newblock \bibinfo{journal}{\emph{arXiv preprint arXiv:2409.08215}} (\bibinfo{year}{2024}).
\newblock


\bibitem[M{\"u}ller et~al\mbox{.}(2024)]%
        {muller2024multidiff}
\bibfield{author}{\bibinfo{person}{Norman M{\"u}ller}, \bibinfo{person}{Katja Schwarz}, \bibinfo{person}{Barbara R{\"o}ssle}, \bibinfo{person}{Lorenzo Porzi}, \bibinfo{person}{Samuel~Rota Bul{\`o}}, \bibinfo{person}{Matthias Nie{\ss}ner}, {and} \bibinfo{person}{Peter Kontschieder}.} \bibinfo{year}{2024}\natexlab{}.
\newblock \showarticletitle{MultiDiff: Consistent Novel View Synthesis from a Single Image}. In \bibinfo{booktitle}{\emph{Proceedings of the IEEE/CVF Conference on Computer Vision and Pattern Recognition}}. \bibinfo{pages}{10258--10268}.
\newblock


\bibitem[Niklaus et~al\mbox{.}(2017)]%
        {niklaus2017video}
\bibfield{author}{\bibinfo{person}{Simon Niklaus}, \bibinfo{person}{Long Mai}, {and} \bibinfo{person}{Feng Liu}.} \bibinfo{year}{2017}\natexlab{}.
\newblock \showarticletitle{Video frame interpolation via adaptive separable convolution}. In \bibinfo{booktitle}{\emph{Proceedings of the IEEE international conference on computer vision}}. \bibinfo{pages}{261--270}.
\newblock


\bibitem[Peebles and Xie(2023)]%
        {peebles2023scalable}
\bibfield{author}{\bibinfo{person}{William Peebles} {and} \bibinfo{person}{Saining Xie}.} \bibinfo{year}{2023}\natexlab{}.
\newblock \showarticletitle{Scalable diffusion models with transformers}. In \bibinfo{booktitle}{\emph{Proceedings of the IEEE/CVF International Conference on Computer Vision}}. \bibinfo{pages}{4195--4205}.
\newblock


\bibitem[Poole et~al\mbox{.}(2022)]%
        {poole2022dreamfusion}
\bibfield{author}{\bibinfo{person}{Ben Poole}, \bibinfo{person}{Ajay Jain}, \bibinfo{person}{Jonathan~T Barron}, {and} \bibinfo{person}{Ben Mildenhall}.} \bibinfo{year}{2022}\natexlab{}.
\newblock \showarticletitle{Dreamfusion: Text-to-{3D} using 2d diffusion}.
\newblock \bibinfo{journal}{\emph{arXiv preprint arXiv:2209.14988}} (\bibinfo{year}{2022}).
\newblock


\bibitem[Ren and Wang(2022)]%
        {ren2022look}
\bibfield{author}{\bibinfo{person}{Xuanchi Ren} {and} \bibinfo{person}{Xiaolong Wang}.} \bibinfo{year}{2022}\natexlab{}.
\newblock \showarticletitle{Look outside the room: Synthesizing a consistent long-term {3D} scene video from a single image}. In \bibinfo{booktitle}{\emph{Proceedings of the IEEE/CVF Conference on Computer Vision and Pattern Recognition}}. \bibinfo{pages}{3563--3573}.
\newblock


\bibitem[Savva et~al\mbox{.}(2019)]%
        {savva2019habitat}
\bibfield{author}{\bibinfo{person}{Manolis Savva}, \bibinfo{person}{Abhishek Kadian}, \bibinfo{person}{Oleksandr Maksymets}, \bibinfo{person}{Yili Zhao}, \bibinfo{person}{Erik Wijmans}, \bibinfo{person}{Bhavana Jain}, \bibinfo{person}{Julian Straub}, \bibinfo{person}{Jia Liu}, \bibinfo{person}{Vladlen Koltun}, \bibinfo{person}{Jitendra Malik}, {et~al\mbox{.}}} \bibinfo{year}{2019}\natexlab{}.
\newblock \showarticletitle{Habitat: A platform for embodied ai research}. In \bibinfo{booktitle}{\emph{Proceedings of the IEEE/CVF international conference on computer vision}}. \bibinfo{pages}{9339--9347}.
\newblock


\bibitem[Seo et~al\mbox{.}(2023)]%
        {seo2023ditto}
\bibfield{author}{\bibinfo{person}{Hoigi Seo}, \bibinfo{person}{Hayeon Kim}, \bibinfo{person}{Gwanghyun Kim}, {and} \bibinfo{person}{Se~Young Chun}.} \bibinfo{year}{2023}\natexlab{}.
\newblock \showarticletitle{Ditto-nerf: Diffusion-based iterative text to omni-directional {3D} model}.
\newblock \bibinfo{journal}{\emph{arXiv preprint arXiv:2304.02827}} (\bibinfo{year}{2023}).
\newblock


\bibitem[Seo et~al\mbox{.}(2024)]%
        {seo2024genwarp}
\bibfield{author}{\bibinfo{person}{Junyoung Seo}, \bibinfo{person}{Kazumi Fukuda}, \bibinfo{person}{Takashi Shibuya}, \bibinfo{person}{Takuya Narihira}, \bibinfo{person}{Naoki Murata}, \bibinfo{person}{Shoukang Hu}, \bibinfo{person}{Chieh-Hsin Lai}, \bibinfo{person}{Seungryong Kim}, {and} \bibinfo{person}{Yuki Mitsufuji}.} \bibinfo{year}{2024}\natexlab{}.
\newblock \showarticletitle{GenWarp: Single Image to Novel Views with Semantic-Preserving Generative Warping}.
\newblock \bibinfo{journal}{\emph{Advances in Neural Information Processing Systems}}  \bibinfo{volume}{37} (\bibinfo{year}{2024}).
\newblock


\bibitem[Shi et~al\mbox{.}(2023)]%
        {shi2023zero123++}
\bibfield{author}{\bibinfo{person}{Ruoxi Shi}, \bibinfo{person}{Hansheng Chen}, \bibinfo{person}{Zhuoyang Zhang}, \bibinfo{person}{Minghua Liu}, \bibinfo{person}{Chao Xu}, \bibinfo{person}{Xinyue Wei}, \bibinfo{person}{Linghao Chen}, \bibinfo{person}{Chong Zeng}, {and} \bibinfo{person}{Hao Su}.} \bibinfo{year}{2023}\natexlab{}.
\newblock \showarticletitle{Zero123++: a single image to consistent multi-view diffusion base model}.
\newblock \bibinfo{journal}{\emph{arXiv preprint arXiv:2310.15110}} (\bibinfo{year}{2023}).
\newblock


\bibitem[Shi et~al\mbox{.}(2022)]%
        {shi2022video}
\bibfield{author}{\bibinfo{person}{Zhihao Shi}, \bibinfo{person}{Xiangyu Xu}, \bibinfo{person}{Xiaohong Liu}, \bibinfo{person}{Jun Chen}, {and} \bibinfo{person}{Ming-Hsuan Yang}.} \bibinfo{year}{2022}\natexlab{}.
\newblock \showarticletitle{Video frame interpolation transformer}. In \bibinfo{booktitle}{\emph{Proceedings of the IEEE/CVF Conference on Computer Vision and Pattern Recognition}}. \bibinfo{pages}{17482--17491}.
\newblock


\bibitem[Singer et~al\mbox{.}(2022)]%
        {singer2022make}
\bibfield{author}{\bibinfo{person}{Uriel Singer}, \bibinfo{person}{Adam Polyak}, \bibinfo{person}{Thomas Hayes}, \bibinfo{person}{Xi Yin}, \bibinfo{person}{Jie An}, \bibinfo{person}{Songyang Zhang}, \bibinfo{person}{Qiyuan Hu}, \bibinfo{person}{Harry Yang}, \bibinfo{person}{Oron Ashual}, \bibinfo{person}{Oran Gafni}, {et~al\mbox{.}}} \bibinfo{year}{2022}\natexlab{}.
\newblock \showarticletitle{Make-a-video: Text-to-video generation without text-video data}.
\newblock \bibinfo{journal}{\emph{arXiv preprint arXiv:2209.14792}} (\bibinfo{year}{2022}).
\newblock


\bibitem[Stan et~al\mbox{.}(2023)]%
        {stan2023ldm3d}
\bibfield{author}{\bibinfo{person}{Gabriela Ben~Melech Stan}, \bibinfo{person}{Diana Wofk}, \bibinfo{person}{Scottie Fox}, \bibinfo{person}{Alex Redden}, \bibinfo{person}{Will Saxton}, \bibinfo{person}{Jean Yu}, \bibinfo{person}{Estelle Aflalo}, \bibinfo{person}{Shao-Yen Tseng}, \bibinfo{person}{Fabio Nonato}, \bibinfo{person}{Matthias Muller}, {et~al\mbox{.}}} \bibinfo{year}{2023}\natexlab{}.
\newblock \showarticletitle{Ldm{3D}: Latent diffusion model for {3D}}.
\newblock \bibinfo{journal}{\emph{arXiv preprint arXiv:2305.10853}} (\bibinfo{year}{2023}).
\newblock


\bibitem[Tang et~al\mbox{.}(2024)]%
        {tang2024diffuscene}
\bibfield{author}{\bibinfo{person}{Jiapeng Tang}, \bibinfo{person}{Yinyu Nie}, \bibinfo{person}{Lev Markhasin}, \bibinfo{person}{Angela Dai}, \bibinfo{person}{Justus Thies}, {and} \bibinfo{person}{Matthias Nie{\ss}ner}.} \bibinfo{year}{2024}\natexlab{}.
\newblock \showarticletitle{Diffuscene: Denoising diffusion models for generative indoor scene synthesis}. In \bibinfo{booktitle}{\emph{Proceedings of the IEEE/CVF Conference on Computer Vision and Pattern Recognition}}.
\newblock


\bibitem[Tseng et~al\mbox{.}(2023)]%
        {tseng2023consistent}
\bibfield{author}{\bibinfo{person}{Hung-Yu Tseng}, \bibinfo{person}{Qinbo Li}, \bibinfo{person}{Changil Kim}, \bibinfo{person}{Suhib Alsisan}, \bibinfo{person}{Jia-Bin Huang}, {and} \bibinfo{person}{Johannes Kopf}.} \bibinfo{year}{2023}\natexlab{}.
\newblock \showarticletitle{Consistent view synthesis with pose-guided diffusion models}. In \bibinfo{booktitle}{\emph{Proceedings of the IEEE/CVF Conference on Computer Vision and Pattern Recognition}}. \bibinfo{pages}{16773--16783}.
\newblock


\bibitem[Unterthiner et~al\mbox{.}(2018)]%
        {unterthiner2018towards}
\bibfield{author}{\bibinfo{person}{Thomas Unterthiner}, \bibinfo{person}{Sjoerd Van~Steenkiste}, \bibinfo{person}{Karol Kurach}, \bibinfo{person}{Raphael Marinier}, \bibinfo{person}{Marcin Michalski}, {and} \bibinfo{person}{Sylvain Gelly}.} \bibinfo{year}{2018}\natexlab{}.
\newblock \showarticletitle{Towards accurate generative models of video: A new metric \& challenges}.
\newblock \bibinfo{journal}{\emph{arXiv preprint arXiv:1812.01717}} (\bibinfo{year}{2018}).
\newblock


\bibitem[Voleti et~al\mbox{.}(2024)]%
        {voleti2024sv3d}
\bibfield{author}{\bibinfo{person}{Vikram Voleti}, \bibinfo{person}{Chun-Han Yao}, \bibinfo{person}{Mark Boss}, \bibinfo{person}{Adam Letts}, \bibinfo{person}{David Pankratz}, \bibinfo{person}{Dmitry Tochilkin}, \bibinfo{person}{Christian Laforte}, \bibinfo{person}{Robin Rombach}, {and} \bibinfo{person}{Varun Jampani}.} \bibinfo{year}{2024}\natexlab{}.
\newblock \showarticletitle{Sv3d: Novel multi-view synthesis and 3d generation from a single image using latent video diffusion}. In \bibinfo{booktitle}{\emph{European Conference on Computer Vision}}. Springer, \bibinfo{pages}{439--457}.
\newblock


\bibitem[Wang et~al\mbox{.}(2024a)]%
        {wang2024vistadream}
\bibfield{author}{\bibinfo{person}{Haiping Wang}, \bibinfo{person}{Yuan Liu}, \bibinfo{person}{Ziwei Liu}, \bibinfo{person}{Zhen Dong}, \bibinfo{person}{Wenping Wang}, {and} \bibinfo{person}{Bisheng Yang}.} \bibinfo{year}{2024}\natexlab{a}.
\newblock \showarticletitle{VistaDream: Sampling multiview consistent images for single-view scene reconstruction}.
\newblock \bibinfo{journal}{\emph{arXiv preprint arXiv:2410.16892}} (\bibinfo{year}{2024}).
\newblock


\bibitem[Wang et~al\mbox{.}(2024b)]%
        {wang2024customizing}
\bibfield{author}{\bibinfo{person}{Hai Wang}, \bibinfo{person}{Xiaoyu Xiang}, \bibinfo{person}{Yuchen Fan}, {and} \bibinfo{person}{Jing-Hao Xue}.} \bibinfo{year}{2024}\natexlab{b}.
\newblock \showarticletitle{Customizing 360-degree panoramas through text-to-image diffusion models}. In \bibinfo{booktitle}{\emph{Proceedings of the IEEE/CVF Winter Conference on Applications of Computer Vision}}. \bibinfo{pages}{4933--4943}.
\newblock


\bibitem[Wang et~al\mbox{.}(2025)]%
        {wang2024generative}
\bibfield{author}{\bibinfo{person}{Xiaojuan Wang}, \bibinfo{person}{Boyang Zhou}, \bibinfo{person}{Brian Curless}, \bibinfo{person}{Ira Kemelmacher-Shlizerman}, \bibinfo{person}{Aleksander Holynski}, {and} \bibinfo{person}{Steven~M Seitz}.} \bibinfo{year}{2025}\natexlab{}.
\newblock \showarticletitle{Generative Inbetweening: Adapting Image-to-Video Models for Keyframe Interpolation}.
\newblock  (\bibinfo{year}{2025}).
\newblock


\bibitem[Wang et~al\mbox{.}(2004)]%
        {wang2004image}
\bibfield{author}{\bibinfo{person}{Zhou Wang}, \bibinfo{person}{Alan~C Bovik}, \bibinfo{person}{Hamid~R Sheikh}, {and} \bibinfo{person}{Eero~P Simoncelli}.} \bibinfo{year}{2004}\natexlab{}.
\newblock \showarticletitle{Image quality assessment: from error visibility to structural similarity}.
\newblock \bibinfo{journal}{\emph{IEEE transactions on image processing}} \bibinfo{volume}{13}, \bibinfo{number}{4} (\bibinfo{year}{2004}), \bibinfo{pages}{600--612}.
\newblock


\bibitem[Wong et~al\mbox{.}(2025)]%
        {wong2025survey}
\bibfield{author}{\bibinfo{person}{Lik Hang~Kenny Wong}, \bibinfo{person}{Xueyang Kang}, \bibinfo{person}{Kaixin Bai}, {and} \bibinfo{person}{Jianwei Zhang}.} \bibinfo{year}{2025}\natexlab{}.
\newblock \showarticletitle{A Survey of Robotic Navigation and Manipulation with Physics Simulators in the Era of Embodied AI}.
\newblock \bibinfo{journal}{\emph{arXiv preprint arXiv:2505.01458}} (\bibinfo{year}{2025}).
\newblock


\bibitem[Wu et~al\mbox{.}(2023a)]%
        {wu2023tune}
\bibfield{author}{\bibinfo{person}{Jay~Zhangjie Wu}, \bibinfo{person}{Yixiao Ge}, \bibinfo{person}{Xintao Wang}, \bibinfo{person}{Stan~Weixian Lei}, \bibinfo{person}{Yuchao Gu}, \bibinfo{person}{Yufei Shi}, \bibinfo{person}{Wynne Hsu}, \bibinfo{person}{Ying Shan}, \bibinfo{person}{Xiaohu Qie}, {and} \bibinfo{person}{Mike~Zheng Shou}.} \bibinfo{year}{2023}\natexlab{a}.
\newblock \showarticletitle{Tune-a-video: One-shot tuning of image diffusion models for text-to-video generation}. In \bibinfo{booktitle}{\emph{Proceedings of the IEEE/CVF International Conference on Computer Vision}}. \bibinfo{pages}{7623--7633}.
\newblock


\bibitem[Wu et~al\mbox{.}(2024)]%
        {wu2024unique3d}
\bibfield{author}{\bibinfo{person}{Kailu Wu}, \bibinfo{person}{Fangfu Liu}, \bibinfo{person}{Zhihan Cai}, \bibinfo{person}{Runjie Yan}, \bibinfo{person}{Hanyang Wang}, \bibinfo{person}{Yating Hu}, \bibinfo{person}{Yueqi Duan}, {and} \bibinfo{person}{Kaisheng Ma}.} \bibinfo{year}{2024}\natexlab{}.
\newblock \bibinfo{title}{Unique{3D}: High-Quality and Efficient {3D} Mesh Generation from a Single Image}.
\newblock
\newblock
\showeprint[arxiv]{2405.20343}~[cs.CV]


\bibitem[Wu et~al\mbox{.}(2023b)]%
        {wu2023panodiffusion}
\bibfield{author}{\bibinfo{person}{Tianhao Wu}, \bibinfo{person}{Chuanxia Zheng}, {and} \bibinfo{person}{Tat-Jen Cham}.} \bibinfo{year}{2023}\natexlab{b}.
\newblock \showarticletitle{PanoDiffusion: 360-degree Panorama Outpainting via Diffusion}. In \bibinfo{booktitle}{\emph{The Twelfth International Conference on Learning Representations}}.
\newblock


\bibitem[Wu et~al\mbox{.}(2023c)]%
        {wu2023ipoldm}
\bibfield{author}{\bibinfo{person}{Tianhao Wu}, \bibinfo{person}{Chuanxia Zheng}, {and} \bibinfo{person}{Tat-Jen Cham}.} \bibinfo{year}{2023}\natexlab{c}.
\newblock \bibinfo{title}{PanoDiffusion: Depth-aided 360-degree Indoor RGB Panorama Outpainting via Latent Diffusion Model}.
\newblock
\newblock
\showeprint[arxiv]{2307.03177}~[cs.CV]


\bibitem[Xia et~al\mbox{.}(2017)]%
        {xia2017inception}
\bibfield{author}{\bibinfo{person}{Xiaoling Xia}, \bibinfo{person}{Cui Xu}, {and} \bibinfo{person}{Bing Nan}.} \bibinfo{year}{2017}\natexlab{}.
\newblock \showarticletitle{Inception-v3 for flower classification}. In \bibinfo{booktitle}{\emph{2017 2nd international conference on image, vision and computing (ICIVC)}}. IEEE, \bibinfo{pages}{783--787}.
\newblock


\bibitem[Xu et~al\mbox{.}(2024)]%
        {xu2024instantmesh}
\bibfield{author}{\bibinfo{person}{Jiale Xu}, \bibinfo{person}{Weihao Cheng}, \bibinfo{person}{Yiming Gao}, \bibinfo{person}{Xintao Wang}, \bibinfo{person}{Shenghua Gao}, {and} \bibinfo{person}{Ying Shan}.} \bibinfo{year}{2024}\natexlab{}.
\newblock \showarticletitle{InstantMesh: Efficient {3D} Mesh Generation from a Single Image with Sparse-view Large Reconstruction Models}.
\newblock \bibinfo{journal}{\emph{arXiv preprint arXiv:2404.07191}} (\bibinfo{year}{2024}).
\newblock


\bibitem[Xu et~al\mbox{.}(2021)]%
        {Xu_2021_CVPR}
\bibfield{author}{\bibinfo{person}{Jiale Xu}, \bibinfo{person}{Jia Zheng}, \bibinfo{person}{Yanyu Xu}, \bibinfo{person}{Rui Tang}, {and} \bibinfo{person}{Shenghua Gao}.} \bibinfo{year}{2021}\natexlab{}.
\newblock \showarticletitle{Layout-Guided Novel View Synthesis From a Single Indoor Panorama}. In \bibinfo{booktitle}{\emph{Proceedings of the IEEE/CVF Conference on Computer Vision and Pattern Recognition (CVPR)}}.
\newblock


\bibitem[Yang et~al\mbox{.}(2024b)]%
        {depth_anything_v2}
\bibfield{author}{\bibinfo{person}{Lihe Yang}, \bibinfo{person}{Bingyi Kang}, \bibinfo{person}{Zilong Huang}, \bibinfo{person}{Zhen Zhao}, \bibinfo{person}{Xiaogang Xu}, \bibinfo{person}{Jiashi Feng}, {and} \bibinfo{person}{Hengshuang Zhao}.} \bibinfo{year}{2024}\natexlab{b}.
\newblock \showarticletitle{Depth Anything V2}.
\newblock \bibinfo{journal}{\emph{arXiv:2406.09414}} (\bibinfo{year}{2024}).
\newblock


\bibitem[Yang et~al\mbox{.}(2024a)]%
        {yang2024direct}
\bibfield{author}{\bibinfo{person}{Shiyuan Yang}, \bibinfo{person}{Liang Hou}, \bibinfo{person}{Haibin Huang}, \bibinfo{person}{Chongyang Ma}, \bibinfo{person}{Pengfei Wan}, \bibinfo{person}{Di Zhang}, \bibinfo{person}{Xiaodong Chen}, {and} \bibinfo{person}{Jing Liao}.} \bibinfo{year}{2024}\natexlab{a}.
\newblock \showarticletitle{Direct-a-Video: Customized Video Generation with User-Directed Camera Movement and Object Motion}.
\newblock \bibinfo{journal}{\emph{arXiv preprint arXiv:2402.03162}} (\bibinfo{year}{2024}).
\newblock


\bibitem[Yang et~al\mbox{.}(2024d)]%
        {yang2024layerpano3dlayered3dpanorama}
\bibfield{author}{\bibinfo{person}{Shuai Yang}, \bibinfo{person}{Jing Tan}, \bibinfo{person}{Mengchen Zhang}, \bibinfo{person}{Tong Wu}, \bibinfo{person}{Yixuan Li}, \bibinfo{person}{Gordon Wetzstein}, \bibinfo{person}{Ziwei Liu}, {and} \bibinfo{person}{Dahua Lin}.} \bibinfo{year}{2024}\natexlab{d}.
\newblock \bibinfo{title}{LayerPano{3D}: Layered {3D} Panorama for Hyper-Immersive Scene Generation}.
\newblock
\newblock
\showeprint[arxiv]{2408.13252}~[cs.CV]
\urldef\tempurl%
\url{https://arxiv.org/abs/2408.13252}
\showURL{%
\tempurl}


\bibitem[Yang et~al\mbox{.}(2024c)]%
        {yang2024scenecraft}
\bibfield{author}{\bibinfo{person}{Xiuyu Yang}, \bibinfo{person}{Yunze Man}, \bibinfo{person}{Jun-Kun Chen}, {and} \bibinfo{person}{Yu-Xiong Wang}.} \bibinfo{year}{2024}\natexlab{c}.
\newblock \showarticletitle{SceneCraft: Layout-Guided {3D} Scene Generation}.
\newblock \bibinfo{journal}{\emph{arXiv preprint arXiv:2410.09049}} (\bibinfo{year}{2024}).
\newblock


\bibitem[Yang et~al\mbox{.}(2024e)]%
        {yang2024cogvideox}
\bibfield{author}{\bibinfo{person}{Zhuoyi Yang}, \bibinfo{person}{Jiayan Teng}, \bibinfo{person}{Wendi Zheng}, \bibinfo{person}{Ming Ding}, \bibinfo{person}{Shiyu Huang}, \bibinfo{person}{Jiazheng Xu}, \bibinfo{person}{Yuanming Yang}, \bibinfo{person}{Wenyi Hong}, \bibinfo{person}{Xiaohan Zhang}, \bibinfo{person}{Guanyu Feng}, {et~al\mbox{.}}} \bibinfo{year}{2024}\natexlab{e}.
\newblock \showarticletitle{CogVideoX: Text-to-Video Diffusion Models with An Expert Transformer}.
\newblock \bibinfo{journal}{\emph{arXiv preprint arXiv:2408.06072}} (\bibinfo{year}{2024}).
\newblock


\bibitem[Yu et~al\mbox{.}(2024a)]%
        {yu2024wonderjourney}
\bibfield{author}{\bibinfo{person}{Hong-Xing Yu}, \bibinfo{person}{Haoyi Duan}, \bibinfo{person}{Junhwa Hur}, \bibinfo{person}{Kyle Sargent}, \bibinfo{person}{Michael Rubinstein}, \bibinfo{person}{William~T Freeman}, \bibinfo{person}{Forrester Cole}, \bibinfo{person}{Deqing Sun}, \bibinfo{person}{Noah Snavely}, \bibinfo{person}{Jiajun Wu}, {et~al\mbox{.}}} \bibinfo{year}{2024}\natexlab{a}.
\newblock \showarticletitle{Wonderjourney: Going from anywhere to everywhere}. In \bibinfo{booktitle}{\emph{Proceedings of the IEEE/CVF Conference on Computer Vision and Pattern Recognition}}. \bibinfo{pages}{6658--6667}.
\newblock


\bibitem[Yu et~al\mbox{.}(2023)]%
        {yu2023long}
\bibfield{author}{\bibinfo{person}{Jason~J Yu}, \bibinfo{person}{Fereshteh Forghani}, \bibinfo{person}{Konstantinos~G Derpanis}, {and} \bibinfo{person}{Marcus~A Brubaker}.} \bibinfo{year}{2023}\natexlab{}.
\newblock \showarticletitle{Long-term photometric consistent novel view synthesis with diffusion models}. In \bibinfo{booktitle}{\emph{Proceedings of the IEEE/CVF International Conference on Computer Vision}}. \bibinfo{pages}{7094--7104}.
\newblock


\bibitem[Yu et~al\mbox{.}(2024b)]%
        {yu2024viewcrafter}
\bibfield{author}{\bibinfo{person}{Wangbo Yu}, \bibinfo{person}{Jinbo Xing}, \bibinfo{person}{Li Yuan}, \bibinfo{person}{Wenbo Hu}, \bibinfo{person}{Xiaoyu Li}, \bibinfo{person}{Zhipeng Huang}, \bibinfo{person}{Xiangjun Gao}, \bibinfo{person}{Tien-Tsin Wong}, \bibinfo{person}{Ying Shan}, {and} \bibinfo{person}{Yonghong Tian}.} \bibinfo{year}{2024}\natexlab{b}.
\newblock \showarticletitle{ViewCrafter: Taming Video Diffusion Models for High-fidelity Novel View Synthesis}.
\newblock \bibinfo{journal}{\emph{arXiv preprint arXiv:2409.02048}} (\bibinfo{year}{2024}).
\newblock


\bibitem[Zhang et~al\mbox{.}(2024b)]%
        {zhang2024taming}
\bibfield{author}{\bibinfo{person}{Cheng Zhang}, \bibinfo{person}{Qianyi Wu}, \bibinfo{person}{Camilo~Cruz Gambardella}, \bibinfo{person}{Xiaoshui Huang}, \bibinfo{person}{Dinh Phung}, \bibinfo{person}{Wanli Ouyang}, {and} \bibinfo{person}{Jianfei Cai}.} \bibinfo{year}{2024}\natexlab{b}.
\newblock \showarticletitle{Taming Stable Diffusion for Text to 360 Panorama Image Generation}. In \bibinfo{booktitle}{\emph{Proceedings of the IEEE/CVF Conference on Computer Vision and Pattern Recognition}}. \bibinfo{pages}{6347--6357}.
\newblock


\bibitem[Zhang et~al\mbox{.}(2024a)]%
        {zhang2024raydiffusion}
\bibfield{author}{\bibinfo{person}{Jason~Y Zhang}, \bibinfo{person}{Amy Lin}, \bibinfo{person}{Moneish Kumar}, \bibinfo{person}{Tzu-Hsuan Yang}, \bibinfo{person}{Deva Ramanan}, {and} \bibinfo{person}{Shubham Tulsiani}.} \bibinfo{year}{2024}\natexlab{a}.
\newblock \showarticletitle{Cameras as Rays: Pose Estimation via Ray Diffusion}. In \bibinfo{booktitle}{\emph{International Conference on Learning Representations (ICLR)}}.
\newblock


\bibitem[Zhang et~al\mbox{.}(2018)]%
        {zhang2018unreasonable}
\bibfield{author}{\bibinfo{person}{Richard Zhang}, \bibinfo{person}{Phillip Isola}, \bibinfo{person}{Alexei~A Efros}, \bibinfo{person}{Eli Shechtman}, {and} \bibinfo{person}{Oliver Wang}.} \bibinfo{year}{2018}\natexlab{}.
\newblock \showarticletitle{The unreasonable effectiveness of deep features as a perceptual metric}. In \bibinfo{booktitle}{\emph{Proceedings of the IEEE conference on computer vision and pattern recognition}}. \bibinfo{pages}{586--595}.
\newblock


\bibitem[Zhang et~al\mbox{.}(2023)]%
        {zhang2023controlvideo}
\bibfield{author}{\bibinfo{person}{Yabo Zhang}, \bibinfo{person}{Yuxiang Wei}, \bibinfo{person}{Dongsheng Jiang}, \bibinfo{person}{Xiaopeng Zhang}, \bibinfo{person}{Wangmeng Zuo}, {and} \bibinfo{person}{Qi Tian}.} \bibinfo{year}{2023}\natexlab{}.
\newblock \showarticletitle{Controlvideo: Training-free controllable text-to-video generation}.
\newblock \bibinfo{journal}{\emph{arXiv preprint arXiv:2305.13077}} (\bibinfo{year}{2023}).
\newblock


\bibitem[Zhou et~al\mbox{.}(2018)]%
        {zhou2018stereo}
\bibfield{author}{\bibinfo{person}{Tinghui Zhou}, \bibinfo{person}{Richard Tucker}, \bibinfo{person}{John Flynn}, \bibinfo{person}{Graham Fyffe}, {and} \bibinfo{person}{Noah Snavely}.} \bibinfo{year}{2018}\natexlab{}.
\newblock \showarticletitle{Stereo magnification: Learning view synthesis using multiplane images}.
\newblock \bibinfo{journal}{\emph{arXiv preprint arXiv:1805.09817}} (\bibinfo{year}{2018}).
\newblock


\end{thebibliography}

\newpage
% \section*{Supplementary Material: Additional Experiments}
% \clearpage
\setcounter{page}{1}

\twocolumn[
\begin{@twocolumnfalse}
\centering
{\huge \textbf{Appendix for Look Beyond: Two-Stage Scene View Generation via Panorama and Video Diffusion}}\par
\par
\vspace{1em}
\end{@twocolumnfalse}
]
% The demo and codes can be accessed from the anonymous GitHub link: 
Supplementary explanations and experimental results are provided beyond the main paper in the following.
% We provide a short demo video via the shared anonymous GitHub link as mentioned in the README (\href{https://anonymous.4open.science/r/acm-mm-pano2video-diffusion-2F3E}{anonymous repository for video demo}). Due to GitHub storage limitations, full-resolution demos and additional results will be available on the public project page upon acceptance.
%%%%%%%%% TITLE - PLEASE UPDATE% 
\setcounter{section}{0}
\section{Perspective View to Panorama View}
\begin{equation}
\mathbf{x}_k = \Pi_{\text{Pan}}^{\text{Per}}(\mathbf{X}_0, \theta_k, \phi_k, \beta, \gamma, H, W), \quad k = {1, \ldots, N^{*}},
\end{equation}
where $\mathbf{X}_0$ represents the equirectangular panorama. The yaw and pitch angles $(\theta_k, \phi_k)$ define the position of each perspective view $\mathbf{x}_k$ on the panorama. The horizontal and vertical fields of view are denoted as $\beta$ and $\gamma$, while $H$ and $W$ specify the height and width of the projected perspective images.

To obtain a perspective view $\mathbf{x}_k$ from the panorama $\mathbf{X}_0$, the projection process maps spherical coordinates to multiple perspective images, where the field of view determines the output dimensions:

\begin{equation}
\Pi_{\text{Pan}}^{\text{Per}} = \begin{bmatrix}
\cos(\theta)\cos(\phi) \\
\sin(\theta)\cos(\phi) \\
\sin(\phi)
\end{bmatrix}.
\end{equation}

Each perspective view undergoes a rotation transformation using the yaw and pitch angles:

\begin{equation}
\begin{aligned}
\mathbf{R}_k &= \mathbf{R}_y(\theta_k)^{-1} \mathbf{R}_z(\phi_k)^{-1}, \\
\mathbf{x}_k &= \mathbf{R}_k \Pi_{\text{Pan}}^{\text{Per}}.
\end{aligned}
\end{equation}

The perspective projection onto the image plane is then computed as:

\begin{equation}
(u_k, v_k) = \begin{cases}
\left(\frac{W\mathbf{x}_k^y}{2w_l\mathbf{x}_k^x} + \frac{W}{2}, \frac{-H\mathbf{x}_k^z}{2h_l\mathbf{x}_k^x} + \frac{H}{2} \right), & \text{if } \mathcal{M}_k = 1, \\
(0, 0), & \text{otherwise}.
\end{cases}
\end{equation}

Here, $\theta \in [-180^{\circ}, 180^{\circ}]$ and $\phi \in [-90^{\circ}, 90^{\circ}]$, while $w_l = \tan(\beta/2)$ and $h_l = \tan(\gamma/2)$ define the lens scaling factors based on the field of view angles.

A visibility mask $\mathcal{M}_k$ is introduced to ensure valid perspective projections:

\begin{equation}
\mathcal{M}_k = \begin{cases}
1, & \text{if } |\mathbf{x}_k^y| < w_l, \ |\mathbf{x}_k^z| < h_l, \text{ and } \mathbf{x}_k^x > 0, \\
0, & \text{otherwise}.
\end{cases}
\end{equation}

Finally, the perspective image of size $H \times W$ is generated through bilinear interpolation with projected pixels $(u_j, v_j), j \in H \times W$.

\section{Panorama View to Perspective View}
\textbf{Image Coordinates to Sphere Mapping.} We compute each pixel in the perspective image's corresponding 3D direction vector on the unit sphere. Let:
\begin{itemize}
    \item $\theta$ be the yaw (left-right) angle in radians, corresponding to longitude.
    \item $\phi$ be the pitch (up-down) angle in radians, corresponding to latitude.
    \item $\beta$ be the field of view (FOV) in degrees.
\end{itemize}

For a pixel $(u, v)$ in the perspective image $\mathbf{x}_i$ of dimensions $H_v \times W_v$, the direction vector $\mathbf{d}(u, v)$ is given by:

\begin{equation}
\mathbf{d}(u, v) = \begin{bmatrix}
1 \\
\frac{2(u - W_v/2)}{W_v} \cdot \tan\left(\frac{\beta}{2}\right) \\
\frac{2(v - H_v/2)}{H_v} \cdot \tan\left(\frac{\beta}{2}\right) \cdot \frac{H_v}{W_v}
\end{bmatrix}.
\end{equation}

\begin{figure*}[!th]
    \centering
    \includegraphics[width=0.98\textwidth]{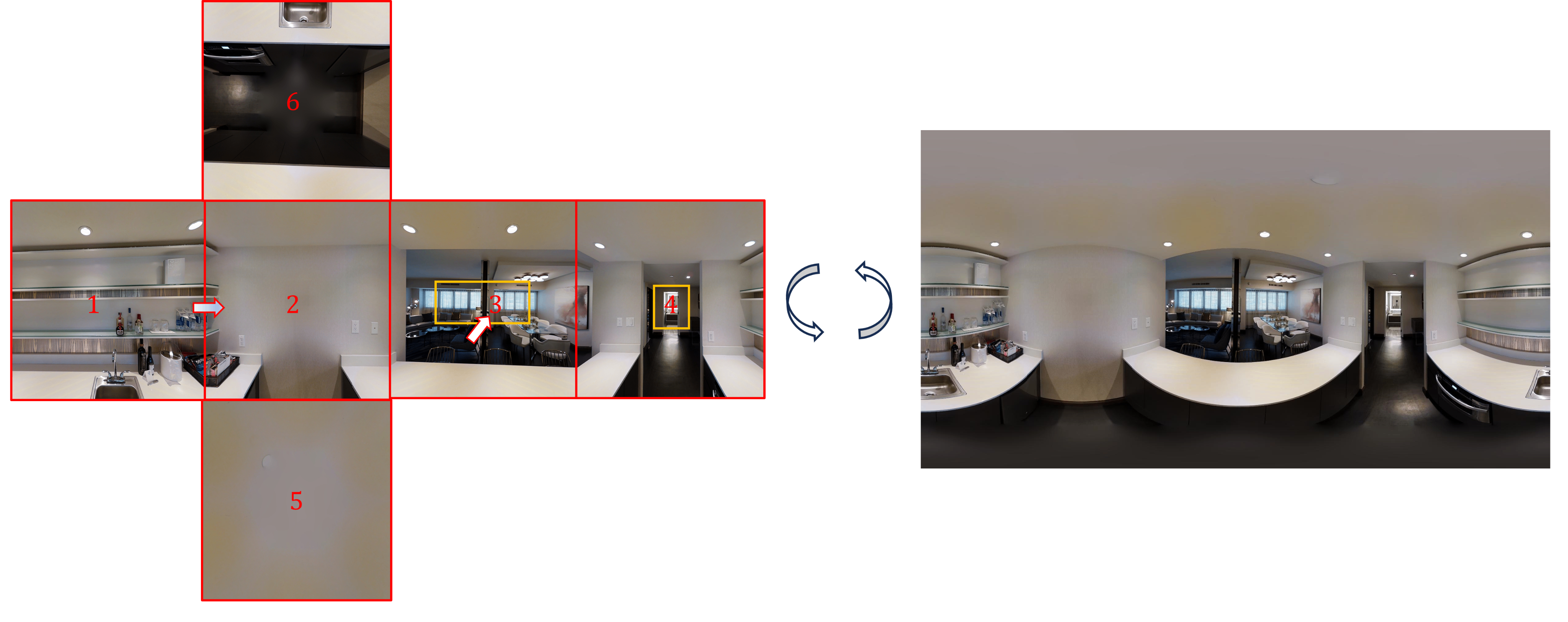}
    %\vspace{-5.5em}
    \caption{ Panorama from or to perspective view image. 
}
    \label{fig:pan-persp}
    \Description{Pano2persp.}
\end{figure*}

To transform $\mathbf{d}(u, v)$ based on the orientation of each view (front, right, back, left, top, bottom), we apply a rotation matrix $\mathbf{R}(\theta, \phi)$, where $\theta$ and $\phi$ define the yaw and pitch angles for each perspective view. The mapping between cube face images and the 360-degree panorama is illustrated in Figure \ref{fig:pan-persp}. 

The yaw and pitch rotation matrices are:

\begin{align}
\mathbf{R}_1(\theta) &= \begin{bmatrix}
\cos(\theta) & -\sin(\theta) & 0 \\
\sin(\theta) & \cos(\theta) & 0 \\
0 & 0 & 1
\end{bmatrix}, \\
\mathbf{R}_2(\phi) &= \begin{bmatrix}
\cos(\phi) & 0 & \sin(\phi) \\
0 & 1 & 0 \\
-\sin(\phi) & 0 & \cos(\phi)
\end{bmatrix}.
\end{align}

The transformed direction vector is then computed as:

\begin{equation}
\mathbf{d}'(u, v) = \mathbf{R}_2(\phi) \mathbf{R}_1(\theta) \mathbf{d}(u, v).
\end{equation}

Next, we project $\mathbf{d}'(u, v)$ back onto the equirectangular panorama $\mathbf{X}_P$. The latitude $\lambda$ and longitude $\gamma$ are obtained from the transformed vector's 3D coordinates $(x', y', z')$:

\begin{equation}
\lambda = \arcsin(z'),
\quad
\gamma = \arctan2(y', x').
\end{equation}

These values are then normalized to pixel coordinates $(u_P, v_P)$ in $\mathbf{X}_P$:

\begin{equation}
u_P = \left(\frac{\gamma}{2\pi} + 0.5\right) W_P,
\quad
v_P = \left(-\frac{\lambda}{\pi} + 0.5\right) H_P.
\end{equation}

Thus, for each pixel in the perspective image, we determine its corresponding pixel $(u_P, v_P)$ in the equirectangular panorama.

We generate multiple perspective views, typically six or eight, corresponding to different viewing directions. Each view $\mathbf{x}_i$ is associated with predefined yaw $\theta$ and pitch $\phi$ angles:

\begin{itemize}
    \item \textbf{Front}: $\theta = 0^\circ, \quad \phi = 0^\circ$
    \item \textbf{Right}: $\theta = 90^\circ, \quad \phi = 0^\circ$
    \item \textbf{Back}: $\theta = 180^\circ, \quad \phi = 0^\circ$
    \item \textbf{Left}: $\theta = 270^\circ, \quad \phi = 0^\circ$
    \item \textbf{Top}: $\theta = 0^\circ, \quad \phi = 90^\circ$
    \item \textbf{Bottom}: $\theta = 0^\circ, \quad \phi = -90^\circ$
\end{itemize}

Each view is generated by applying the corresponding rotations and mapping the transformed direction vectors back to the panorama.

% \subsection{Transformation Pipeline}
The transformation from the equirectangular panorama $\mathbf{X}_P$ to each perspective view $\mathbf{x}_i$ follows these steps:
\begin{enumerate}
    \item Compute the direction vector $\mathbf{d}(u, v)$ for each pixel in the perspective image.
    \item Apply the rotation matrices $\mathbf{R}_1(\theta)$ and $\mathbf{R}_2(\phi)$ to transform the direction vector.
    \item Compute the corresponding latitude $\lambda$ and longitude $\gamma$ from the transformed vector.
    \item Map $(\lambda, \gamma)$ to pixel coordinates $(u_P, v_P)$ in the equirectangular image $\mathbf{X}_P$.
    \item Extract the color value from $\mathbf{X}_P$ and assign it to the corresponding pixel in $\mathbf{x}_i$.
\end{enumerate}

% Given 3D Cartesian coordinates \( (x, y, z) \), the corresponding spherical coordinates (longitude \( \lambda \) and latitude \( \phi \)) are given by:

% \[
% \lambda = \text{atan2}(x, z)
% \]
% \[
% \phi = \arcsin(y)
% \]

% For a panorama with width \( W_{eq} \) and height \( H_{eq} \), the pixel coordinates \( (X, Y) \) corresponding to \( (\lambda, \phi) \) are:

% \[
% X = \left( \frac{\lambda}{2 \pi} + 0.5 \right) \cdot (W_{eq} - 1)
% \]
% \[
% Y = \left( \frac{\phi}{\pi} + 0.5 \right) \cdot (H_{eq} - 1)
% \]

% The camera intrinsic matrix \( K \) for a perspective image with field of view \( \text{FOV} \) and resolution \( (W, H) \) is:

% \[
% f = \frac{0.5 \cdot W}{\tan(0.5 \cdot \text{FOV})}
% \]
% \[
% K = \begin{bmatrix}
% f & 0 & c_x \\
% 0 & f & c_y \\
% 0 & 0 & 1
% \end{bmatrix}
% \]
% where \( c_x = \frac{W - 1}{2} \) and \( c_y = \frac{H - 1}{2} \).

% The yaw rotation matrix \( R_1 \) (around the y-axis) is:

% \[
% R_1 = \begin{bmatrix}
% \cos(\text{THETA}) & 0 & \sin(\text{THETA}) \\
% 0 & 1 & 0 \\
% -\sin(\text{THETA}) & 0 & \cos(\text{THETA})
% \end{bmatrix}
% \]

% The pitch rotation matrix \( R_2 \) (around the x-axis) is:

\begin{figure*}[!th]
    \centering
    \includegraphics[width=0.84\textwidth]{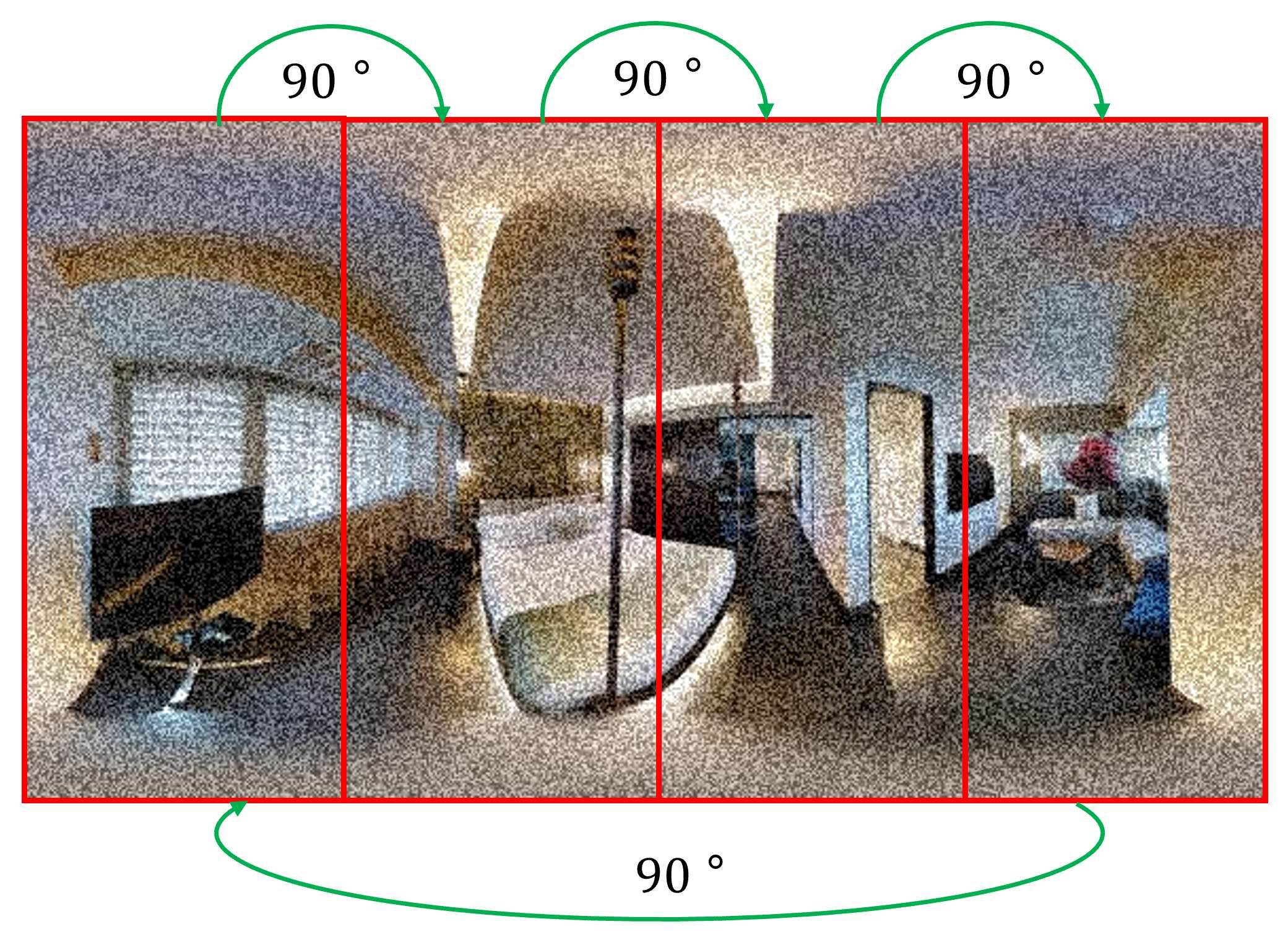}
    %\vspace{-5.5em}
    \caption{ Cycle consistency loss is enforced during the inference sampling stage by partitioning the panorama into four regions. Each region is shifted one-quarter to the right, with the rightmost region wrapped around to the leftmost position in the panorama image mask.
}
    \label{fig:cycle-loss}
\Description{Cycle consistency.}
\end{figure*}

Our demo video, showcasing the full transformation from a single image to a panoramic scene and a video walkthrough of the complete 360$^{\circ}$ environment, follows a structured trajectory. The trajectory planning adopts a star-shaped pattern centered at $O$, systematically exploring the scene. For each direction with sufficient depth—determined by comparing the average frustum depth to the nearest corner point distance—we implement a room exploration strategy: moving outward from the center, scanning the space, and returning before rotating to the next direction.  

Path generation follows two modes: (1) \textit{Room traversal}, where paths $\{l_1, l_2\}$ or $\{l_3, l_4\}$ are constructed using a combination of straight and curved segments while maintaining a safe distance from walls, and (2) \textit{Circular sector paths} $\{l_5, l_6, l_7\}$ for navigating open areas. Corner points $P_i$, where $i = 1, \dots, 8$, are extracted from undistorted depth maps at $\pi/4$ intervals to define the room structure. During training, we augment these patterns by introducing varied turn curvatures alongside ground truth trajectories.

\begin{figure*}[!th]
    \centering
    \includegraphics[width=0.98\textwidth]{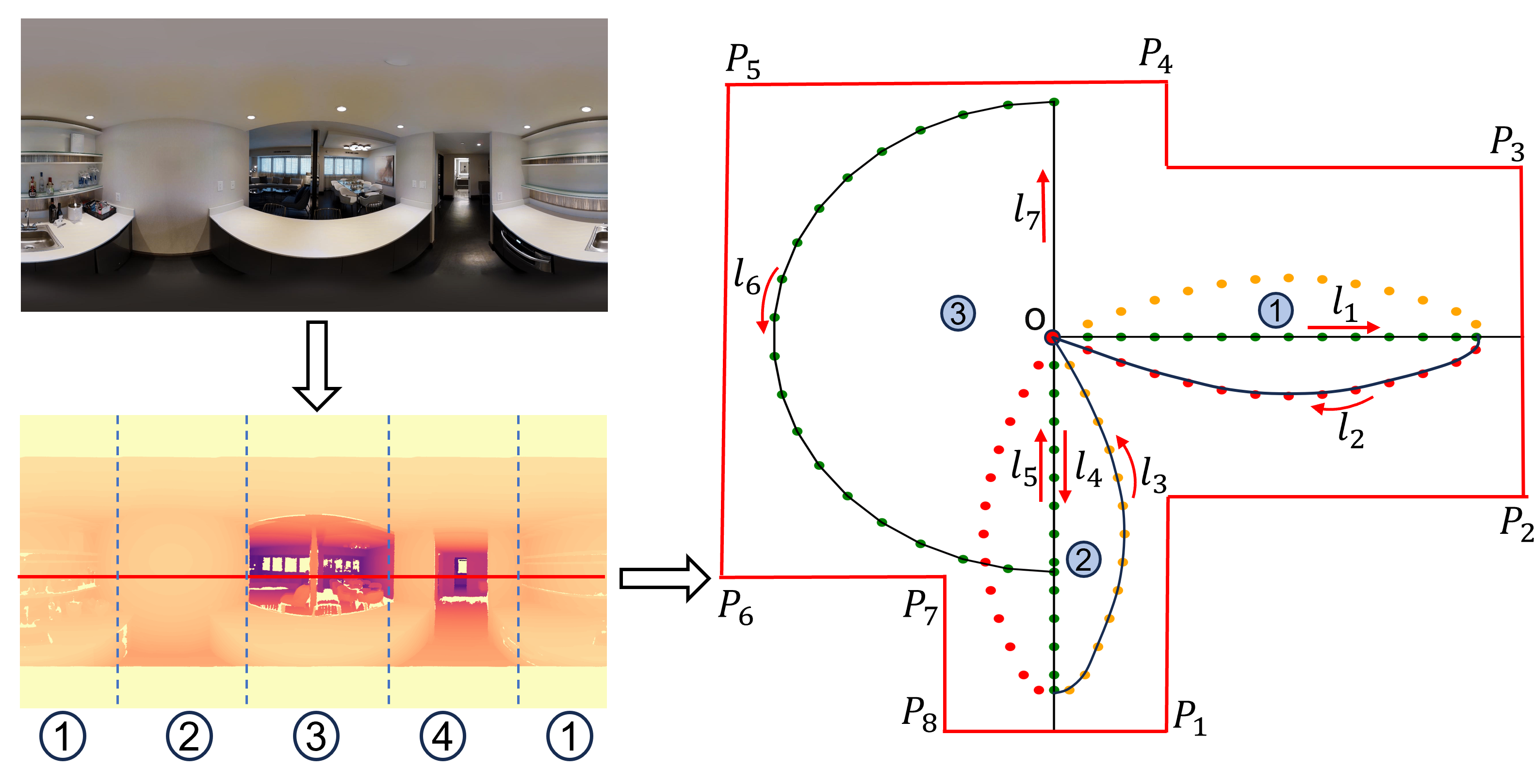}

\caption{Custom trajectory generation in panoramic scenes. Using depth maps from (a) panorama view and (b) depth estimation, we partition the scene into four sections and extract room structure for trajectory planning. Black lines indicate probing directions from camera position $O$, with possible paths shown in orange (left turn), red (right turn), and green (straight).}
    \label{fig:gen-traj}
\Description{Custom traj.}
\end{figure*}

\section{DiT Building Block}
\begin{figure}[!thbp]
  \centering
\includegraphics[width=\linewidth]{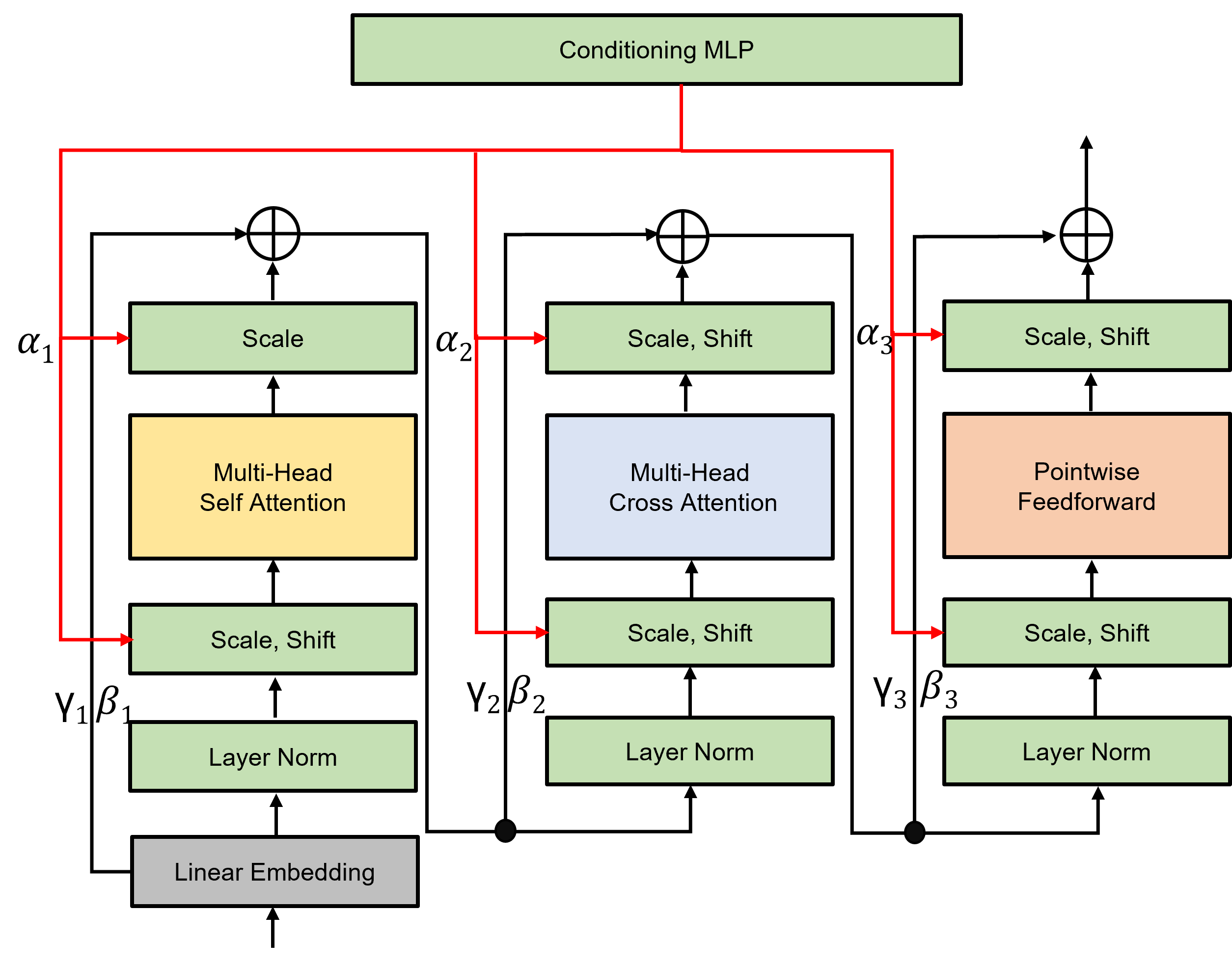}
  \centering
  \caption{Our panorama diffusion transformer block contains both self-attention and cross-attention blocks. }
  \vspace{-1.6em}
  \label{fig:DiT}
\Description{Pano self cross atten.}
\end{figure}

Our panorama diffusion model is mainly composed of several repeating Diffusion Transformer (DiT) blocks, where each block consists of three key cascaded modules as demonstrated in Figure \ref{fig:DiT}. The DiT block processes tokens of diffusion feature maps through a sequence of layers, beginning with Layer Normalization, followed by Multi-Head Self-Attention, Multi-Head Cross-Attention conditioned on CLIP image features, and a Pointwise Feedforward Network. The self-attention mechanism within the DiT block captures spatial dependencies, while the cross-attention module integrates semantic information of the input image. Outputs from each layer are modulated by distribution parameters via a conditioning Multi-Layer Perceptron (MLP) prediction, which applies the scale and shift parameters to the feature distribution.

\section{Outpainting Sampling}
The latent feature $z_{t-1} \in \mathbb{R}^ {H' \times W'\times C'}$ diffusion process consists of the following six steps:
\begin{itemize} 
\item \textbf{Forward Noise:} The top row illustrates the forward noising process, where known regions (white areas) remain unchanged while unknown regions (colored areas) progressively accumulate noise, resulting in a noisy representation in latent space. 
\item \textbf{Denoising:} The bottom row shows the denoising process, where the model gradually reduces noise in the unknown regions while preserving the integrity of the known regions, starting from the fully noised state. 
\item \textbf{outpainting in Latent Space:} All operations are conducted in latent feature space rather than pixel space, facilitating a more efficient and effective outpainting process. 
\item \textbf{Progressive Refinement:} Moving left to right in the bottom row, unknown regions become increasingly inpainted and consistent, aligning with known regions.
\item \textbf{Final Output:} The rightmost image represents the final denoised feature map, with previously unknown regions seamlessly filled by the known regions. 
\item \textbf{Latent to Pixel Space:} The final step, which is not shown, involves passing the denoised latent representation through the pre-trained VAE decoder to obtain the inpainted image in pixel space from the feature map space. 
\end{itemize}
\begin{figure*}[!th]
    \centering
    \includegraphics[width=0.98\textwidth]{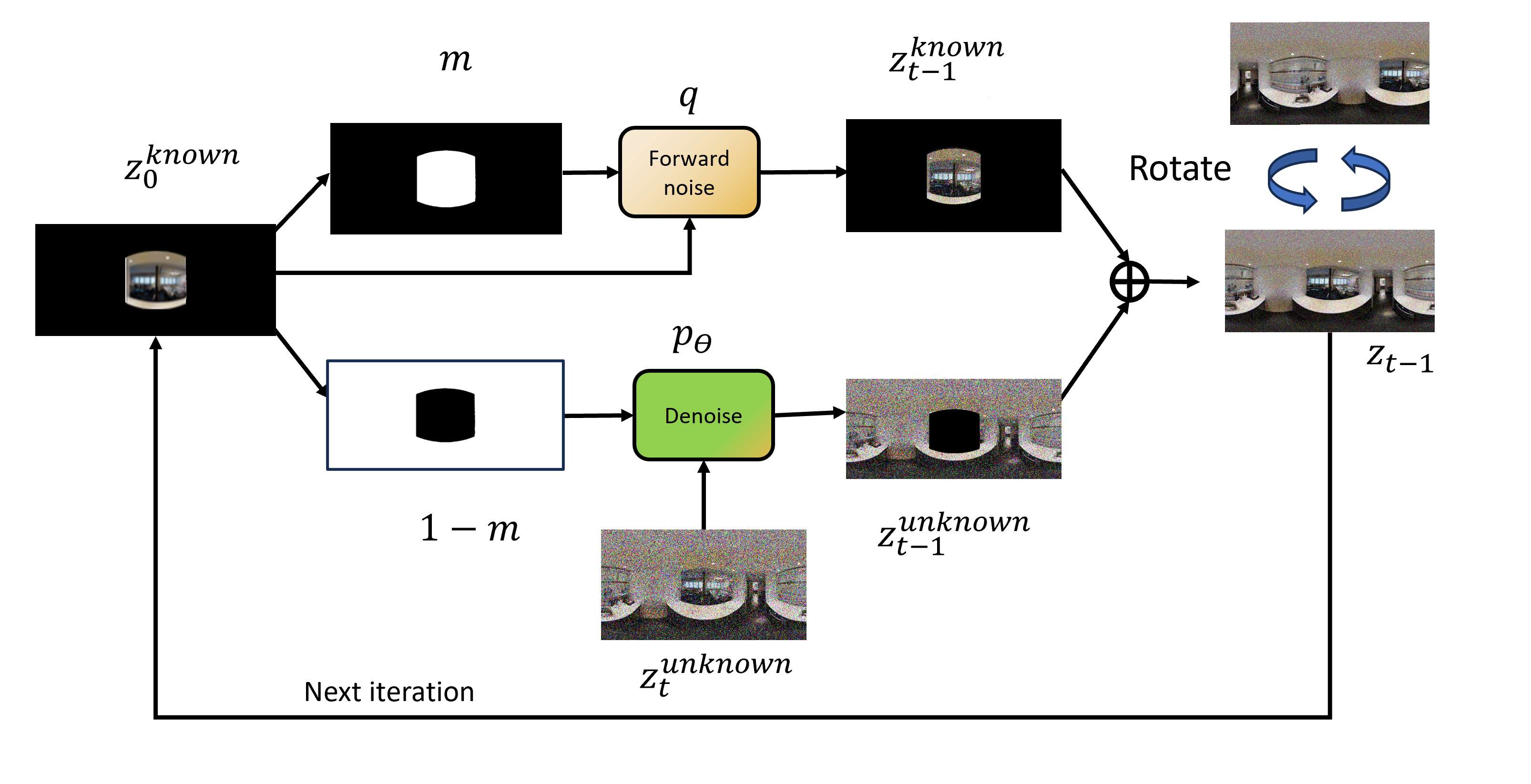}
    %\vspace{-5.5em}
    \caption{Outpainting diffusion process at each iteration on panorama image by fusion of known region through forward noising and unknown region through reverse denoising. We use the same cycle outpainting loss inspired by PanoDiffusion \cite{wu2023panodiffusion}.}
    \label{fig:outpainting}
    \Description{Outpainting.}
\end{figure*}

As shown in Figure \ref{fig:cycle-loss}, our panorama outpainting enforces 360$^{\circ}$ consistency through a cyclic shifting mechanism. The panorama is partitioned into four regions, each spanning a 90$^{\circ}$ field of view (FOV), and a consistency loss is applied during inference sampling. At each iteration, the rightmost region shifts to the leftmost position, while all other regions shift rightward by 90$^{\circ}$, ensuring seamless boundary alignment. This cyclic rotation strategy enforces global consistency across the entire 360$^{\circ}$ panorama, requiring generated content in overlapping regions to remain consistent under all possible circular shifts, and such cycle outpainting loss is inspired by PanoDiffusion \cite{wu2023panodiffusion}, which can be used in simulator \cite{wong2025survey}.

\section{Video Diffusion with Camera Control}
\begin{figure*}[!th]
    \centering
    \includegraphics[width=0.98\textwidth]{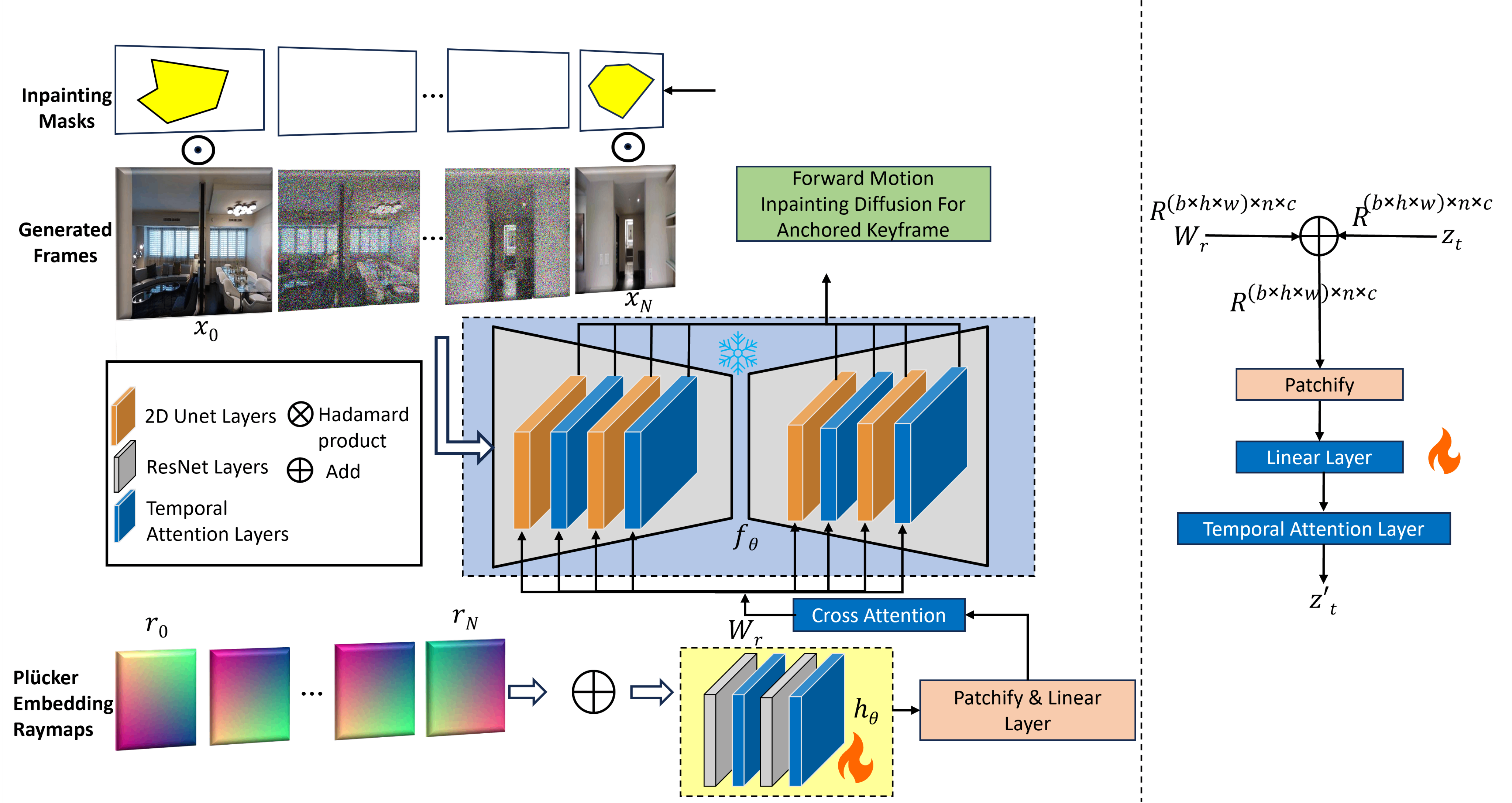}
    %\vspace{-5.5em}
    \caption{The video diffusion model $f_{\theta}$ is frozen, while the raymap conditioning encoder $h_{\theta}$ is learnable. For keyframe-anchored inpainting, each input frame $\mathbf{x}_t$ is masked by the Hadamard product $\mathbf{x}_t \odot \mathbf{m}_t$ to guide generation in occluded regions. The corresponding raymap embeddings $\{\mathbf{r}_0, \dots, \mathbf{r}_N\}$, derived from Plücker-embedded camera rays via intrinsic $\mathbf{K}$ and essential matrix $\mathbf{E}$, are fused across time and channels into the feature volume $\mathbf{W}_r$. These features are processed by 2D CNNs and temporal attention layers in $h_{\theta}$ to produce the conditioning input. Cross-attention then injects $\mathbf{W}_r$ into $f_{\theta}$ during denoising. The right block shows the "Patchify \& Linear Layer" module used in both branches, where features $\mathbf{z}_t$ are further refined by temporal attention to yield $\mathbf{z}'_t$.}
    \label{fig:video-diffusion-condition}
\Description{Video diffusion.}
\end{figure*}

To achieve camera-aware video generation, we propose a bi-directional diffusion framework, as illustrated in Figure \ref{fig:video-diffusion-condition}. The trained video diffusion backbone $f_{\theta}$ remains frozen while a learnable raymap conditioning encoder is introduced. The framework synthesizes intermediate frames between the source $\mathbf{x}_0$ and target $\mathbf{x}_N$ using bi-directional score fusion, guided by camera pose information.  

Each input frame $\mathbf{x}_i$ is multiplied by a corresponding mask $\mathbf{m}_i$ to label source and target images as non-diffusion inputs (black mask). When generating target images via a walk-in motion, peripheral regions undergo outpainting, where missing content is synthesized through video diffusion using an outpainting noise score.  

For each frame, we compute raymap embeddings $\{\mathbf{r}_0, \dots, \mathbf{r}_N\}$ from camera parameters $(\mathbf{E}, \mathbf{K})$ and concatenate them into a spatio-temporal feature volume $\mathbf{W}_r$. This volume is compressed via 1D convolution into a 4-channel representation, then processed through alternating 2D CNN and temporal attention layers.  

% A key innovation in our approach is the attention-flipping mechanism: attention maps extracted from $f_{\theta}$ undergo horizontal and vertical flips, generating $f_{\theta'}$, which is fused with the original features. The processing pipeline transforms input features via patchification and linear projection ($\mathbf{z}_t$), followed by temporal attention to produce the final representation $\mathbf{z}'_t$.  

During generation, binary masks indicate frames requiring denoising. Source and target frames are weighted by zero masks to exclude them from the diffusion process. However, for target images generated via walk-in motion synthesis, peripheral image regions require inpainting, as they contain unknown content inferred from the known region.

For non-anchor timesteps, the frames are fully generated by the model from the noise vector $x_N \sim \mathcal{N}(0, I)$ down to the final reconstruction $x_0$. The inpainting masks are either omitted (i.e., set to all-ones) or are identity operations, ensuring that generation is unconstrained.

This strategy allows the model to leverage explicit motion priors and view-dependent features while anchoring reconstruction around keyframes using ray-guided geometric consistency. The Hadamard fusion also ensures spatial localization in both training and inference, yielding sharper synthesis and improved temporal coherence.

The complete frame update pipeline integrates both spatial cues from the Plücker raymap embedding $r_t$ and learned visual features from the masked keyframe $x_t \otimes m_t$, processed through cross-attentional and temporal modules:
\begin{align}
    W_t &= h_\theta(r_t \oplus (x_t \otimes m_t)), \\
    Z_t &= f_\theta(x_t), \\
    Z'_t &= \text{TempAttn}(\text{Linear}(\text{Patchify}(W_r \oplus Z_t)))\,.
\end{align}
Here, $W_r$ refers to the learned embedding of ray-guided feature tokens across timesteps, and $Z'_t$ is the temporally integrated latent used for subsequent denoising steps in the diffusion trajectory.

Specifically, Attention Maps: The rotated temporal self-attention maps 
$\mathbf{A}'_{i}$, which are directly fed into the backward prediction, could be cached after one pass through the 3D UNet. This would allow you to reuse these maps instead of recomputing them in the forward branch.
Output Projections: The final output projection matrices, and $W_{v_o}$, if cached, could reduce the need for recalculating these for each pass, since they are consistent across both branches.

\noindent\textbf{Spatial Encoding}: Each video frame \( \mathbf{x}_i \) is processed through a pretrained ResNet-50 model to extract spatial features. Here, \( i \) represents the sequence index, and \( t \) denotes the frame within that sequence. The spatial encoder maps each frame to a feature vector:
   \[
   \mathbf{f}_{i,t} = \text{ResNet50}(\mathbf{x}_i),
   \]
   where \( \mathbf{f}_{i,t} \in \mathbb{R}^{2048} \) represents the spatial feature vector of frame \( t \) in sequence \( i \).

\noindent\textbf{Temporal Encoding}: The spatial feature vectors from each sequence are stacked to form a sequence volume \( \mathbf{F}_i \):
   \[
   \mathbf{F}_i = [\mathbf{f}_{i,1}, \mathbf{f}_{i,2}, \dots, \mathbf{f}_{i,T}],
\]
   where \( T \) is the number of frames in each sequence, and \( \mathbf{F}_i \in \mathbb{R}^{T \times 2048} \). This stacked feature volume \( \mathbf{F}_i \) is then passed through a 3D-UNet-based temporal encoder to capture camera motion across frames:
   \[
   \mathbf{z}_i = \Phi(\mathbf{F}_i),
   \]
   where \( \mathbf{z}_i \in \mathbb{R}^{512} \) represents the temporal feature vector for the \( i \)-th sequence, encoding both spatial and temporal information.
\section{Raymap Embedding}
For the second stage of video diffusion, 
the source and target camera orientations are represented as quaternions $\mathbf{q}_k$ and $\mathbf{q}_{k+1}$.\{($\mathbf{R}_i, \mathbf{T}_i$), (i = 1, \ldots, N)\} %The extrinsic poses are used to condition each view frame generation. %We map the extrinsic camera pose ($\mathbf{R}_i, \mathbf{T}_i$) to a feature map  $\mathbf{r}_i \in \mathbb{R}^{ H \times W \times C}$, where diffusion feature channel $C$ is 6.

\noindent\textbf{Raymap Feature.} We employ the Plücker embedding \cite{zhang2024raydiffusion} based raymap. Raymaps of all the trajectories posed are concatenated to be used in volume feature encoding.
Each feature per pixel is to represent a 3D ray in 3D using Plücker embedding in the same world frame, which is a 6D representation of a pixel ray, composed of a ray direction vector and a moment vector, orthogonal to the ray direction. The direction vector $\mathbf{d}_k$ of a pixel ray passing through each pixel in an image is calculated from the projection of pixel $u_k$ via intrinsic matrix $\mathbf{R}^{T}_{i}\mathbf{K}_{i}^{-1}u_k, k \in\mathbb{R}^{ H \times W}$.
Each ray moment $\mathbf{m}_k$ is calculated as the cross product of the extrinsic essential matrix ($-\mathbf{R}_i^{T}\mathbf{T}_i$) and the direction vector $\mathbf{d}$:
\begin{equation}    
    \mathbf{r}_k = \begin{bmatrix} \mathbf{m}_k \\ \mathbf{d}_k \end{bmatrix} = \begin{bmatrix}  (-\mathbf{R}_i^{T}\mathbf{T}_i)\times \mathbf{R}_i^{T}\mathbf{K}_{i}^{-1}u_{k} \\ \mathbf{R}_i^{T}\mathbf{K}_{i}^{-1}u_{k}. \end{bmatrix}
\end{equation}
% We also compared the results of using Plücker embedding with other types of embedding-based raymap, including quaternion embedding, ray direction embedding, and Fourier embedding in the following experiment section. 
% Given an extrinsic camera matrix $\mathbf{T} \in \mathbb{R}^{4 \times 4}$, the Plücker embedding of rays can be derived as follows:
% The function outputs an array of Plücker embedding coordinates for each ray, reshaped to match the dimensions of the image. This results in a 6D Plücker embedding $\mathbf{P} \in \mathbb{R}^{h \times w \times 6}$ for the input image.
% Quaternion embedding is another technique for encoding the orientation of rays. A quaternion is a 4D number often used to represent rotations in 3D space, and it can be derived from a ray direction vector. Given the extrinsic matrix $\mathbf{T} \in \mathbb{R}^{4 \times 4}$, the process for quaternion embedding is to compute the ray directions in world coordinates for each pixel. 
For each ray direction vector $\mathbf{d} = [d_x, d_y, d_z]^\top$, construct the corresponding quaternion $\mathbf{q}$. In this simplified representation, the quaternion consists of a zero scalar part and the ray direction vector as the imaginary components:
    \begin{equation}        
    \mathbf{q} = [ 0, d_x, d_y, d_z ]
    \end{equation}
This quaternion represents a rotation along the ray's direction.
The function outputs an array of quaternions for each ray, reshaped to match the dimensions of the image.

This results in a 4D quaternion embedding $\mathbf{q} \in \mathbb{R}^{h \times w \times 4}$ for the input image.
Given a pair of keyframes \(I_0\) and \(I_{N-1}\), 
Our diffusion video goal is to generate a video \(\{\mathbf{x}_0, \mathbf{x}1, \dots, \mathbf{x}{N-1}\}\) that begins with frame \(\mathbf{x}_0\) and ends with frame \(\mathbf{x}{N-1}\), leveraging the pre-trained image-to-video Stable Video Diffusion (SVD) model.
The previously mentioned camera poses are upsampled along the temporal dimension if the pose change of two adjacent frames is larger than the interval, and the camera pose of the new frame is 
linearly interpolated from two neighboring camera pose \( (\mathbf{R}_k, \mathbf{T}_k) \) and \( (\mathbf{R}_{k+1}, \mathbf{T}_{k+1}) \). 
defined as:
\begin{equation}
\text{SLERP}(\mathbf{q}_k^{}, \mathbf{q}_{k+1}^{}, \lambda) = \frac{\sin((1 - \lambda) \theta)}{\sin(\theta)} \mathbf{q}_k^{} + \frac{\sin(\lambda \theta)}{\sin(\theta)} \mathbf{q}_{k+1}^{},
\label{eq:quaternion-interp}
\end{equation}
Where \( \theta \) is the cosine angle between the two quaternions, which can be computed using the dot product $\cos(\theta) = \mathbf{q}_{k}^{} \cdot \mathbf{q}_{k+1}^{}$. \( \lambda \in [0, 1] \) is an interpolation parameter proportional to $
(t - t_k)/(t_{k+1} - t_k)$, where $t$ is any time between $t_{k}$ and $t_{k+1}$. Additionally, to ensure that the interpolation follows the shortest path on the quaternion unit sphere, if \( \cos(\theta) < 0 \), we flip \( q \) to \(-q\). 
The spherical interpolation formula now ensures that the rotation is smoothly interpolated along the shortest path between \( q_0 \) and \( q_1 \).

Linear quaternion interpolation (LERP) works well for small-angle rotations. However, for larger rotations, the interpolation path may deviate from the unit quaternion space. To avoid this issue, spherical linear interpolation (SLERP) is typically preferred. However, since the problem specifies linear interpolation, we employ LERP here.
% \textbf{Normalization} is a crucial step to ensure quaternions maintain unit length and should not be omitted.

The source and target camera orientations are converted to quaternions $\mathbf{q}_k^{}$ and $\mathbf{q}_{k+1}^{}$ respectively, thus the Spherical Linear Interpolation (SLERP) of quaternions is normally used. When the angle difference \( \theta \) between adjacent frames is small, SLERP can be approximated by linear interpolation (LERP): 
\begin{equation}    
\text{LERP}(\mathbf{q}_k^{}, \mathbf{q}_{k+1}^{}, \lambda) = (1 - \lambda) \mathbf{q}_k^{} + \lambda \mathbf{q}_{k+1}^{}.
\label{eq:linear-lerp}
\end{equation}
% and $H$ and $W$ are the spatial latent dimensions. The spatial
% layers interpret the video as a batch of independent images
% (by shifting the temporal axis into the batch dimension), and
% for each temporal mixing layer $\ell^{\varphi}_\theta$, we reduce back to video
% dimensions as follows (using \texttt{einops} [64] notation):
% \[
% z' \leftarrow \texttt{rearrange}(z, (b\ t\ c\ h\ w) \rightarrow (b\ c\ t\ h\ w)),
% \]
% \[
% z' \leftarrow \ell^{\varphi}_\theta(z', c),
% \]
% \[
% z' \leftarrow \texttt{rearrange}(z', (b\ c\ t\ h\ w) \rightarrow (b\ t\ c\ h\ w)),
% \]
% where we added the batch dimension $b$ for clarity. In other
% words, the spatial layers treat all $B\cdot T$ encoded video frames
% independently in the batch dimension, while the temporal
% mixing layers $\ell^{\varphi}_\theta, c$ process entire videos in a new temporal
% dimension $c$. Furthermore, $c$ is optional conditioning
% information such as a text prompt. After each temporal layer,
% the output $z'$ is combined with a mix of $\alpha^{\varphi}_\theta z' + (1-\alpha^{\varphi}_\theta) z'; \alpha^{\varphi}_\theta \in [0, 1]$ denotes a (learnable) parameter (also Appendix D).

\section{Implementation Details}
\noindent\textbf{Matterport3D Rendering Views:} The perspective views are rendered from an embodied agent deployed in the Habitat simulator~\cite{savva2019habitat}, paired with a panoramic image.

Our efficient Fréchet Video Distance (FVD)~\cite{unterthiner2018towards} implementation processes each image by first passing it through an encoder for feature embedding, followed by fusing temporal information along the stacked sequence dimension.  
To compute the Fréchet Video Distance (FVD), we apply both spatial and temporal encoding to video frames, generating a spatiotemporal feature embedding from which distribution statistics are calculated.

\noindent\textbf{Mean and Covariance Calculation}: After obtaining temporal features \( \mathbf{z}_i \) for each sequence, we compute the mean \( \mu \) and covariance \( \Sigma \) across all sequences in the dataset. Let \( N \) be the total number of sequences. The mean and covariance are given by:
   \[
   \mu = \frac{1}{N} \sum_{i=1}^{N} \mathbf{z}_i,
   \]
   \[
   \Sigma = \frac{1}{N - 1} \sum_{i=1}^{N} (\mathbf{z}_i - \mu)(\mathbf{z}_i - \mu)^\top,
   \]
   where \( \mu \in \mathbb{R}^{512} \) is the mean vector, and \( \Sigma \in \mathbb{R}^{512 \times 512} \) is the covariance matrix of the spatiotemporal feature embeddings.

\noindent \textbf{FVD Calculation}: Using these mean and covariance statistics, the FVD score between real and generated videos is computed as:
   \[
   \text{FVD} = \|\mu_r - \mu_g\|^2 + \text{Tr}(\Sigma_r + \Sigma_g - 2 \sqrt{\Sigma_r \Sigma_g}),
   \]
   Where \( \mu_r, \Sigma_r \) represent the mean and covariance of real video features, while \( \mu_g, \Sigma_g \) correspond to generated video features. The term \( \|\mu_r - \mu_g\|^2 \) denotes the squared difference between means, and \( \text{Tr}(\cdot) \) represents the trace operator. The matrix square root \( \sqrt{\Sigma_r \Sigma_g} \) is computed for the covariance product. Finally, the normalized features after encoding are globally scaled back to align with the original data distribution.

\section{Experiments and Results}
We provide additional ablation study analysis and additional qualitative comparison results in the following.
\subsection{Ablation Study}
\begin{figure}[!tbp]
  \centering
\includegraphics[width=\linewidth]{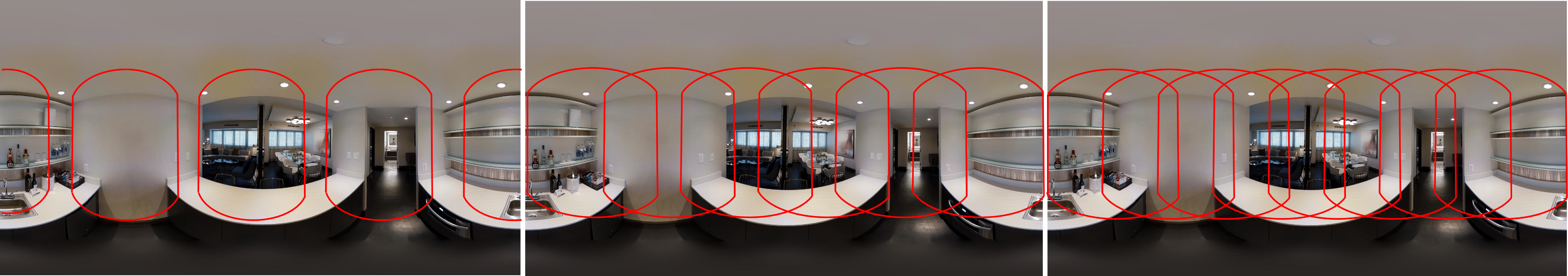}
  \centering
  \caption{Perspective keyframe extracted from the panorama view image through neighboring keyframes with various overlapping ratios of 4, 6, and 8 keyframes.}
  \vspace{-1.2em}
  \label{fig:neighbor-setup}
  \Description{Persp from pano.}
\end{figure}

\begin{figure}[!tbp]
  \centering
\includegraphics[width=\linewidth]{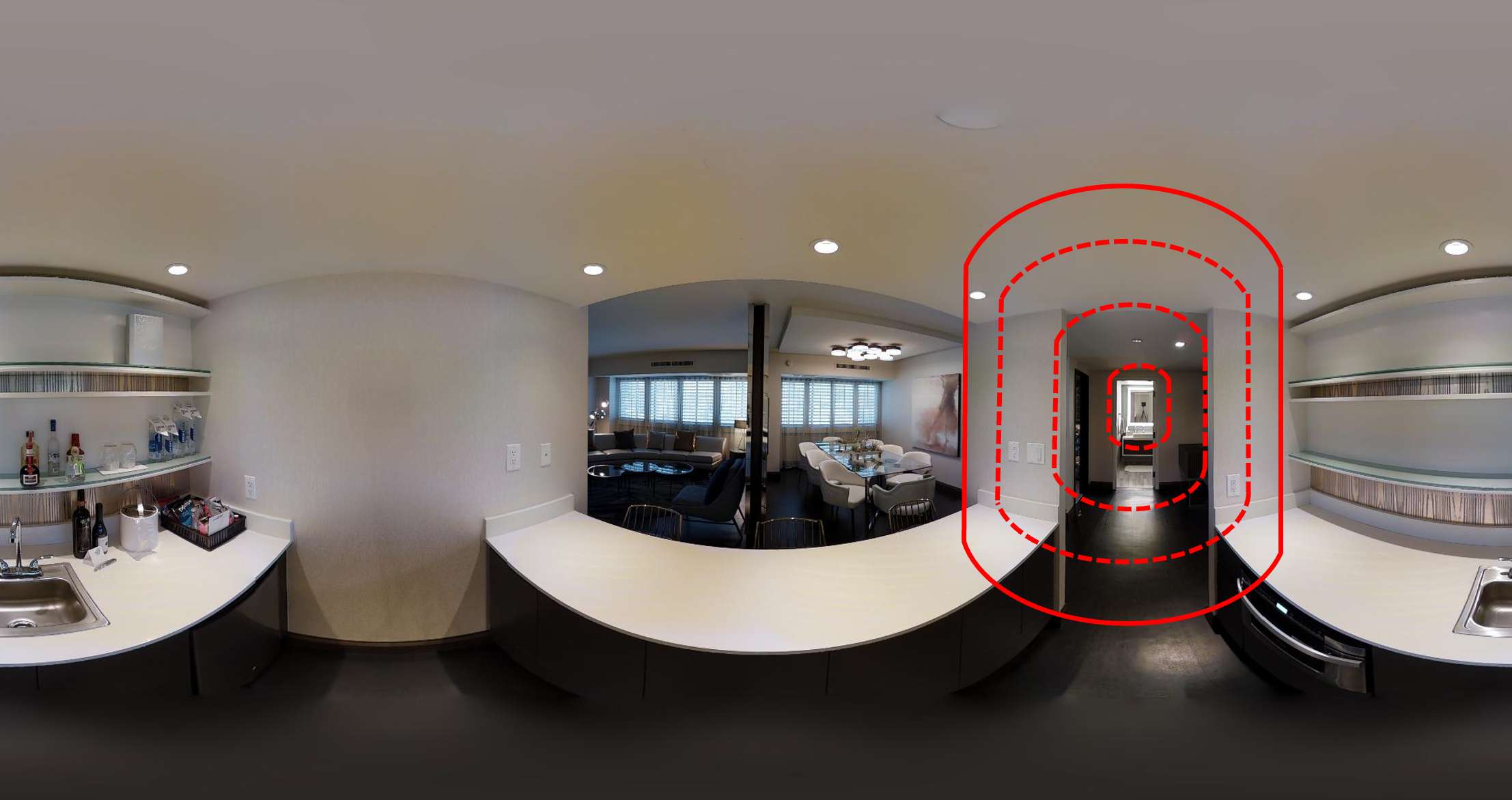}
  \centering
  \caption{Keyframe extracted from the panorama view image through walk-in motion within a valid room view on the panorama image with various walk-in motion factors.}
  \vspace{-1.2em}
  \label{fig:walkin-setup}
\Description{keyframe from pano walk-in.}
\end{figure}

This figure demonstrates the impact of varying neighboring frame overlaps on the generation of panoramic sequences. As the overlap between keyframes increases (progressing from "No Overlap" to "2/3 Overlap"), the visual consistency of the generated sequences improves. In cases of minimal or no overlap (first two rows), noticeable mismatches are evident in the synthesized frames, particularly in dynamic regions (indicated by red circles). By incorporating more keyframes and increasing the overlap, transitions between frames become smoother, leading to improved alignment and reduced artifacts. The ground truth (GT) row provides a reference for the ideal appearance of the sequence. The setup for this neighboring keyframe mode is demonstrated in Figure \ref{fig:neighbor-setup}.

\begin{figure}[!tbp]
  \centering
\includegraphics[width=\linewidth]{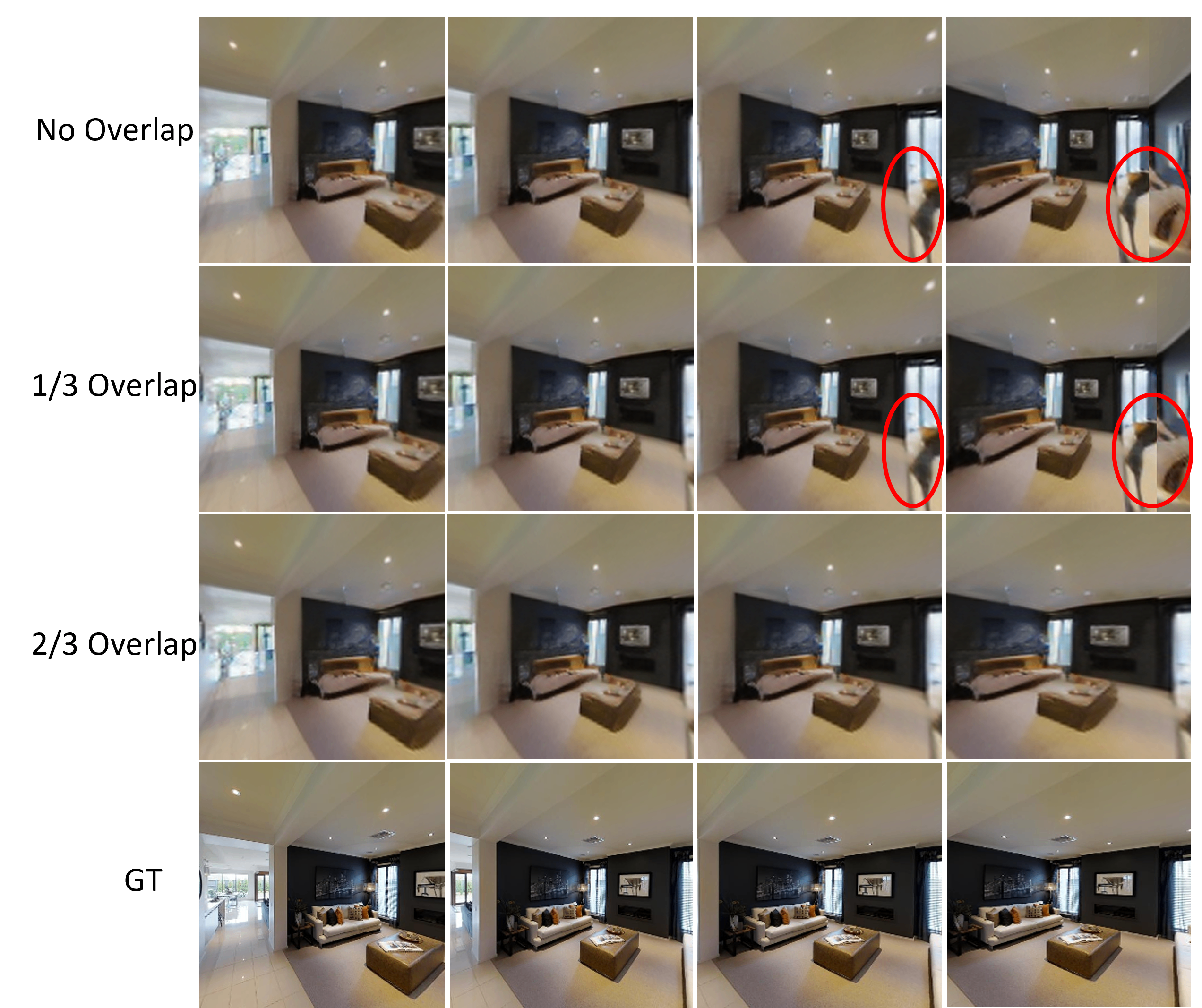}
  \centering
  \caption{Video generation of pure rotation based on the two neighboring keyframes extracted from the panorama image, with various overlapping ratios, ranging from zero to one-half. The last row is the results of the GT sequence images. The red circle highlights the mismatch region on the generated image. }
  \vspace{-1.6em}
  \label{fig:rotation-cmps}
  \Description{keyframe from pano by rotation in-place.}
\end{figure}

% The figure illustrates the effect of straight-line distance on synthesized image quality across different distance ratios. As the walk-in motion factor increases (corresponding to smaller distance ratios), the scene appears closer to the camera, simulating a shorter forward step. Conversely, smaller walk-in motion factors simulate farther distances, resulting in longer straight-line motion. Notably, image quality deteriorates as the distance increases from the source camera pose, with sharper details in closer views and increased blur or distortion in farther ones. This highlights the challenges of maintaining high-fidelity synthesis over long sequences, particularly when emulating distant viewpoints. The setup for generating a keyframe through this walk-in motion is demonstrated in Figure \ref{fig:walkin-setup}.

% \begin{figure}[!tbp]
%   \centering
% \includegraphics[width=\linewidth]{figures/zoomin-cmps.png}
%   \centering
%   \caption{Video generation by a walk-in motion within the valid room view window for straight line walk, with various walk-in motion ratios $c = \hat{d} / d_{\text{Max}}$ ranging from 0.2, 0.4 to 0.8. The last row is the results of the GT sequence images.}
%   \vspace{-1.6em}
%   \label{fig:walkin-cmps}
% \end{figure}

\begin{figure}[!th]
  \centering
\includegraphics[width=\linewidth]{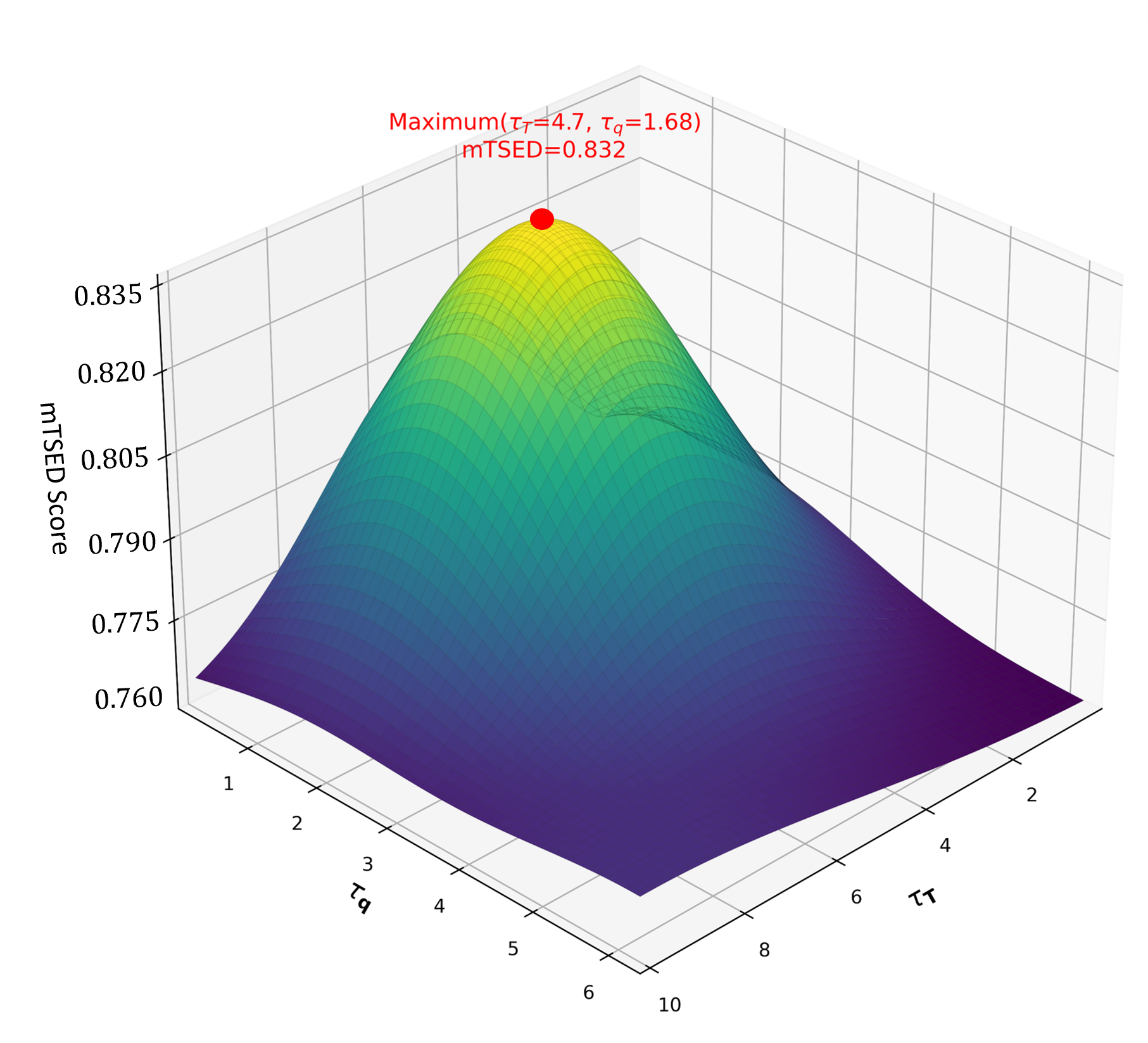}
  \centering
  \caption{The plot of mTSEd as a function of temperature parameter $\tau_{\mathbf{T}}$ in meter and $\tau_{\mathbf{q}}$ in radian. The red dot indicated the maximum value tested on the RealEstate10K subset. The grid size for the temperature parameter tuning is set to 0.1. }
  \vspace{-1.6em}
  \label{fig:temp-param}
  \Description{Plot of temperature.}
\end{figure}

Figure \ref{fig:temp-param} presents a 3D visualization of mTSED scores across varying temperature parameters, $\tau_T$ and $\tau_q$. The mTSED score, which reflects the effectiveness of temporal-spatial encoding, reaches its maximum value of 0.832 at the specific combination of parameters $\tau_T = 4.7$ meters and $\tau_q = 1.68$ radians. The highlighted peak, marked by a red dot, emphasizes the optimal settings for these parameters. The gradual gradient across the surface demonstrates the sensitivity of mTSED scores to parameter variations, revealing a clear trend that indicates regions of high performance. This analysis facilitates the identification of parameter configurations that maximize encoding performance.

To evaluate the effectiveness of panorama-guided conditioning, we conduct an ablation study on the Matterport3D dataset. As shown in Table~\ref{tab:ab-pano-keyframe}, performance improves consistently with richer keyframe conditioning. Using only a single input view yields the weakest results across all metrics, while introducing multiple keyframes extracted from the panorama significantly enhances both spatial and temporal consistency. Further gains are achieved by augmenting panorama keyframes with walk-in warped keyframes, demonstrating the benefit of integrating both global scene context and local view alignment. The full configuration with six panoramas and six walk-in keyframes achieves the best overall performance, confirming the importance of comprehensive multi-view conditioning for high-quality video interpolation.
\begin{table}[!th]
\vspace{-0.4em}
\centering
    \caption{Ablation study on the impact of panorama-guided conditioning for video diffusion, and the test is on Matterport3D. 
    \colorbox{orange!50}{\textbf{Orange}} indicates the best result.}
    \vspace{-0.8em}
\begin{adjustbox}{width=\columnwidth}
     \begin{tabular}{l c c c c c c }
        \toprule
        Conditioning Strategy & LPIPS $\downarrow$ & PSNR $\uparrow$ & SSIM $\uparrow$ & FID $\downarrow$ & FVD $\downarrow$ & mTSED $\uparrow$\\
        \hline \hline
        Single Input View & 0.495 & 9.361 & 0.436 & 205.124 & 257.83 & 0.573 \\
        3 Keyframes (Panorama Only) & 0.378 & 12.803 & 0.486 & 151.254 & 183.47 & 0.685 \\
        3 + 3 Keyframes (Panorama + Walk-in) & 0.402 & 14.312 & 0.541 & 80.382 & 83.261 & 0.850 \\
        6 + 6 Keyframes (Panorama + Walk-in) & \cellcolor{orange!50} 0.321 & \cellcolor{orange!50} 14.698 & \cellcolor{orange!50} 0.548 & \cellcolor{orange!50} 77.402 & \cellcolor{orange!50} 79.14 & \cellcolor{orange!50} 0.882 \\
        \bottomrule
    \end{tabular}
\end{adjustbox}
\vspace{-0.8em}
\label{tab:ab-pano-keyframe}
\end{table}
\subsection{Baseline Comparisons}
\begin{figure*}[!tbp]
    \centering
    \includegraphics[width=1.0\textwidth,  trim=6pt 4pt 25pt 0pt, clip]{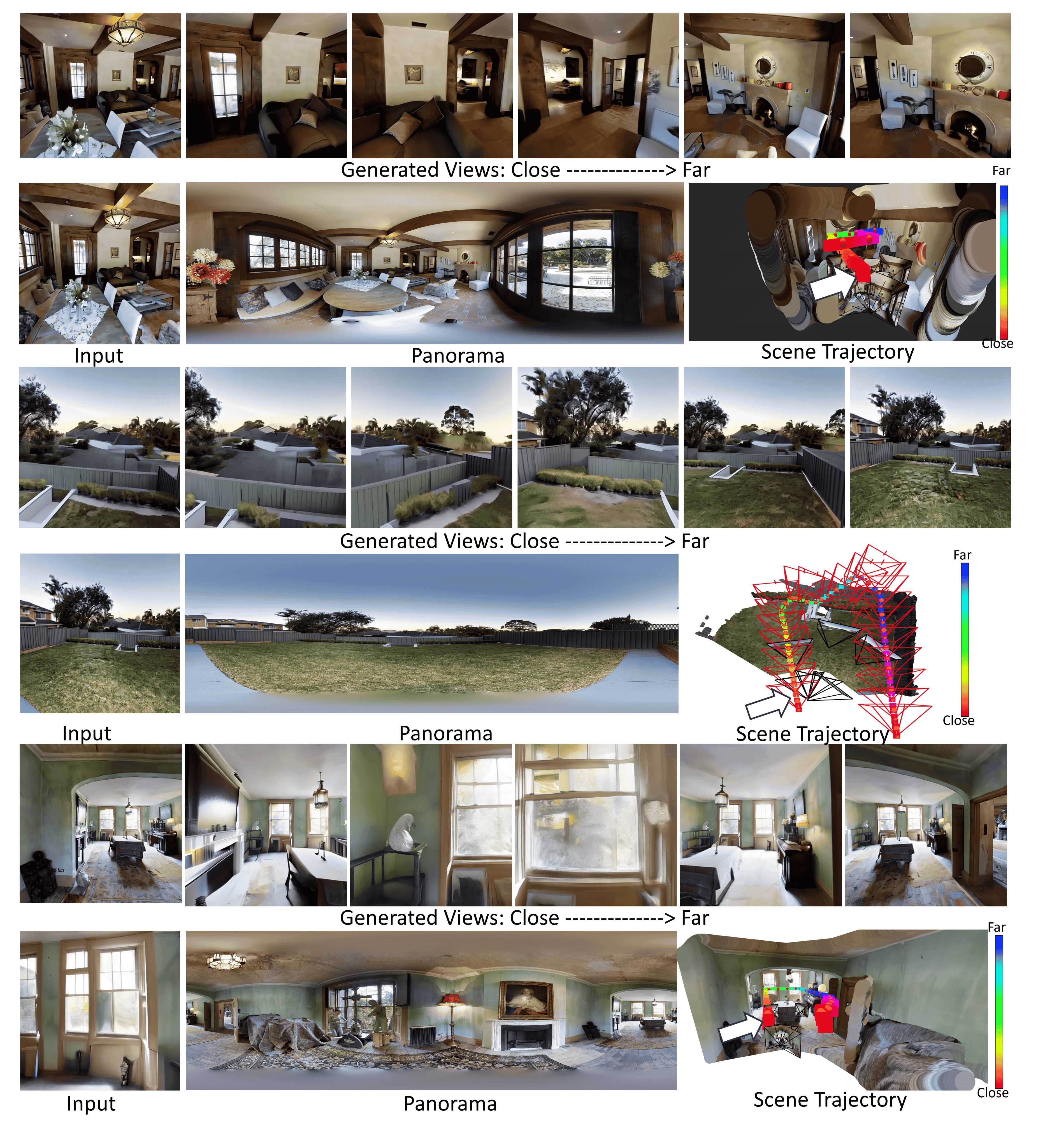}
    \vspace{-1.5em}
    \caption{Additional qualitative results of our model, showing generated views (odd rows) along a looped trajectory. The leftmost image in each even row is the input view, marked in red on the adjacent scene panorama. The rightmost image in each even row displays the reconstructed scene with the overlaid query trajectory; the white arrow indicates the starting point, and the trajectory color encodes distance from the input view.}
    \label{fig:our-more}
    \vspace{-1.0em}
    \Description{Image number.}
\end{figure*}

We first present additional qualitative results of our model in Figure~\ref{fig:our-more}, showing generated views along a looping trajectory. These results demonstrate that our approach maintains both scene and long-term view consistency, even after a full loop. In contrast, baseline models shown in the main paper struggle to preserve scene coherence after extended camera movement.

\begin{figure*}[!tbp]
    \centering
    \includegraphics[width=1.0\textwidth,  trim=6pt 4pt 25pt 0pt, clip]{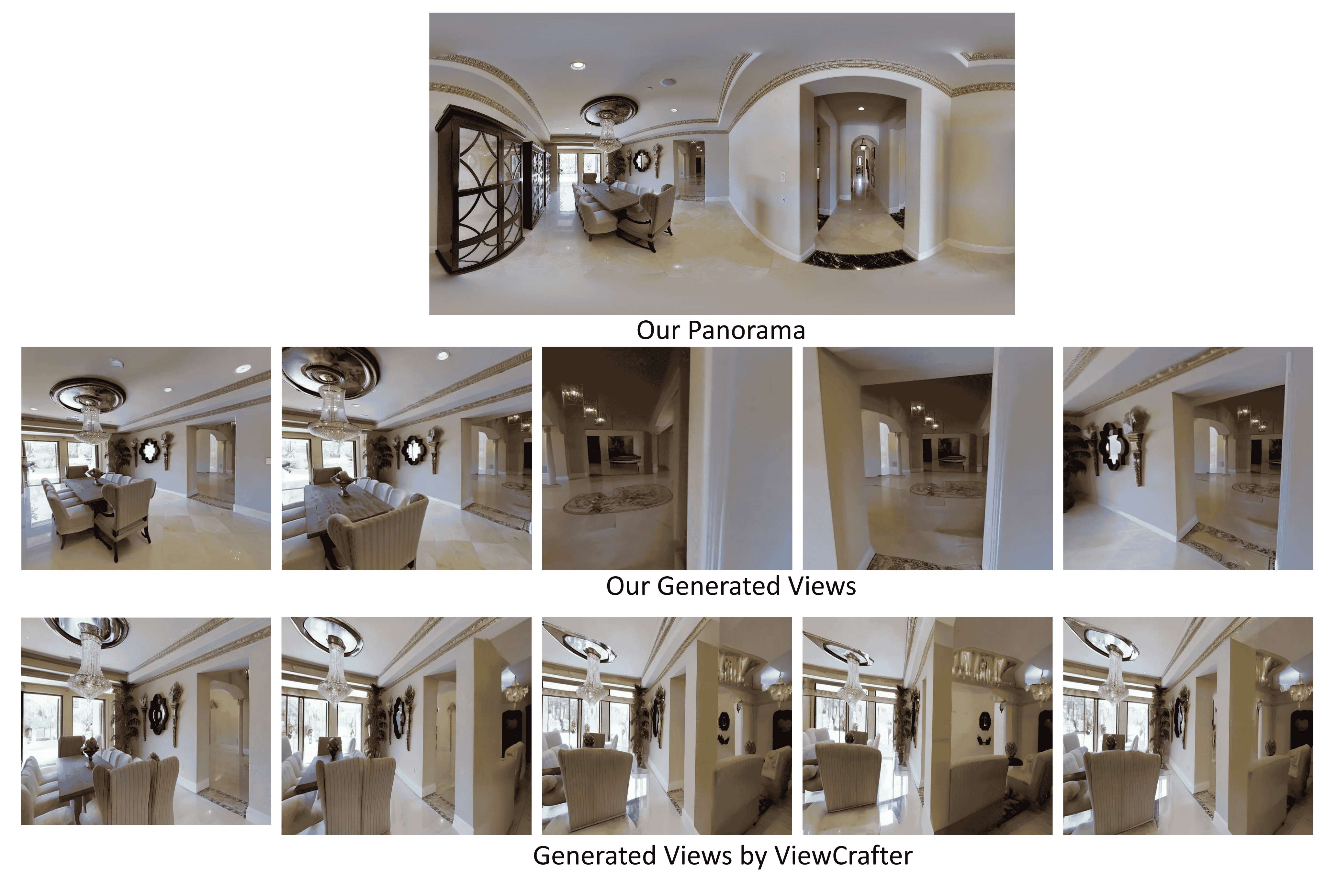}
    \vspace{-1.5em}
    \caption{Comparison with ViewCrafter. The first row shows our generated panorama; the second row presents our synthesized novel views; the third row displays ViewCrafter results.}
    \label{fig:supple-flip}
\Description{Qualitative.}
    \vspace{-1.0em}
\end{figure*}

Then, we highlight a key limitation of the current leading baseline, ViewCrafter. Although it generates locally consistent views, it fails to maintain contextual coherence when the camera rotates around corners. As shown in the last row of Figure~\ref{fig:supple-flip}, a corridor is incorrectly hallucinated into a different room, and an existing chair is replaced with a sofa. In comparison, our model preserves scene integrity, thanks to the two-stage conditioning on both the panoramic scene and anchor keyframes.

Next, we present results from the WonderJourney~\cite{yu2024wonderjourney} baseline, generated along a predefined trajectory, to illustrate the hallucination issue inherent in Large Language Models. Specifically, the model gradually transitions from indoor scenes to outdoor landscapes, revealing a lack of scene coherence.

Additional qualitative results on the two datasets discussed in the paper are provided in the following sections.
The following pages, including Ground Truth (GT) Views along the GT short trajectory, usually like a straight line walk or left turn, right turn, to show our model's performance of better view alignment to the GT viewpoints.

\begin{figure}[!thbp]
  \centering
\includegraphics[width=\linewidth]{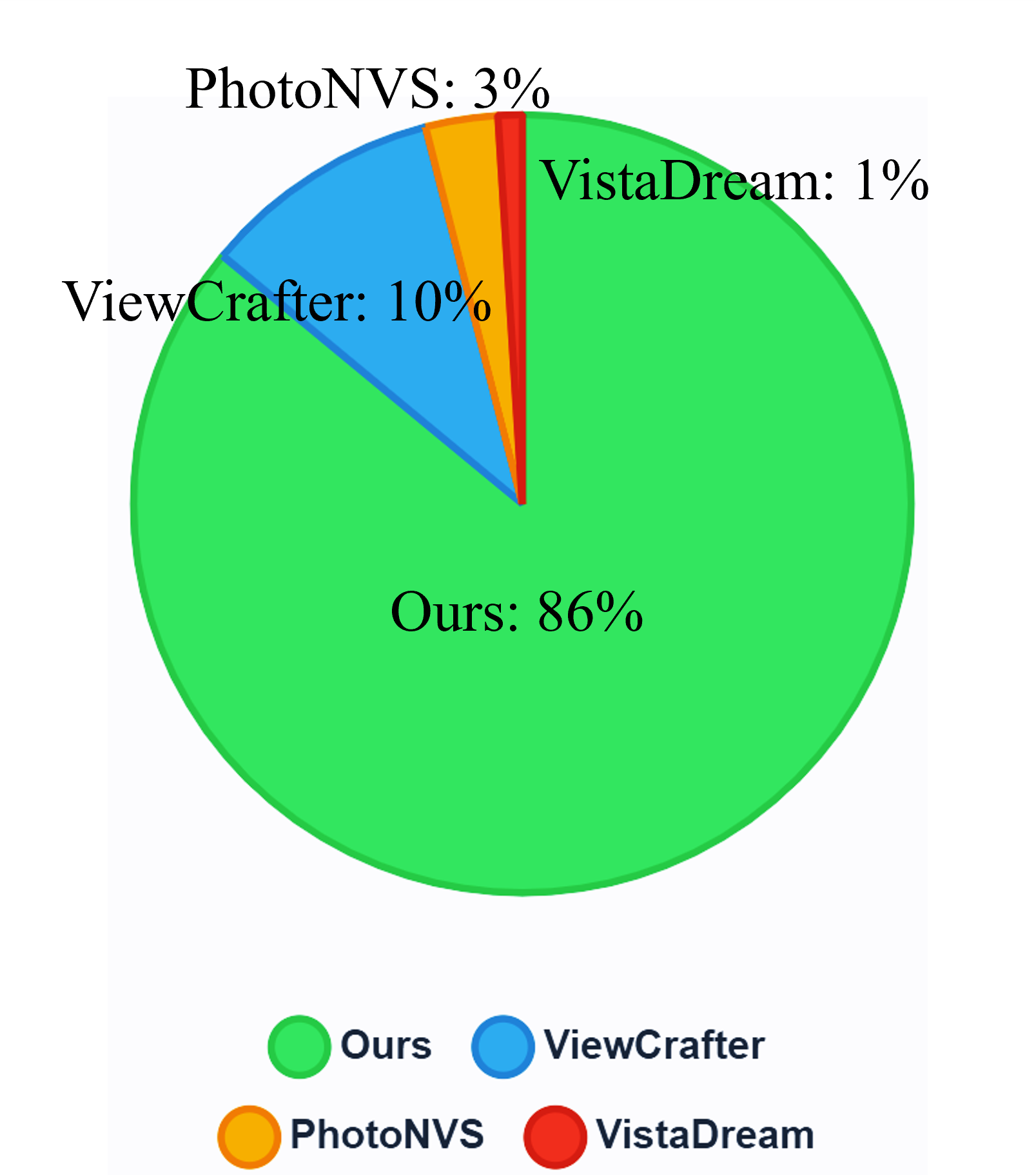}
  \centering
  \caption{User study results showing method preference based on perceptual quality and realism. Our method was preferred in 86\% of 5,000 total comparisons (50 participants $\times$ 100 comparisons each).}
  \vspace{-1.6em}
  \label{fig:survey-pref}
  \Description{Survey.}
\end{figure}

Finally, the pie chart in Figure~\ref{fig:survey-pref} presents subjective evaluation results from our user study conducted during the rebuttal phase. We recruited 50 participants from various backgrounds, each evaluating 100 sets of video clip comparisons (5,000 total comparisons), where all methods generated videos from the same input view. Participants selected their preferred method based on perceptual quality and realism. The results demonstrate a strong preference for our approach, which was chosen in 86\% of cases (shown in green), significantly outperforming baseline methods: ViewCrafter received 10\% preference (blue), PhotoNVS 3\% (orange), and VistaDream 1\% (red). This substantial margin indicates our method's superior performance in generating high-quality, realistic video content compared to existing state-of-the-art approaches.

\begin{figure*}[!tbp]
    \centering
    \includegraphics[width=1.0\textwidth,  trim=6pt 4pt 25pt 0pt, clip]{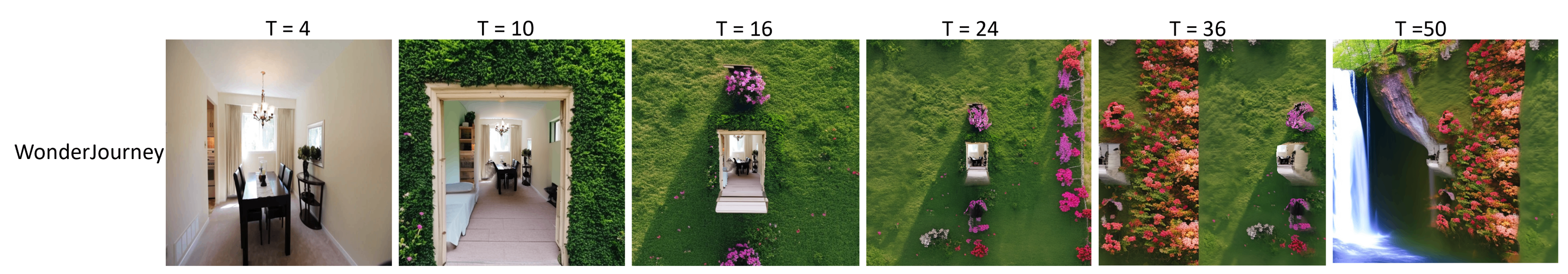}
    \vspace{-1.5em}
    \caption{WonderJouney sample results on a single image input.}
    \label{fig:wonder-journey}
\Description{Wonderjourney results.}
    \vspace{-1.0em}
\end{figure*}

\begin{figure*}[!tbp]
    \centering
    \includegraphics[width=1.0\textwidth,  trim=6pt 4pt 25pt 0pt, clip]{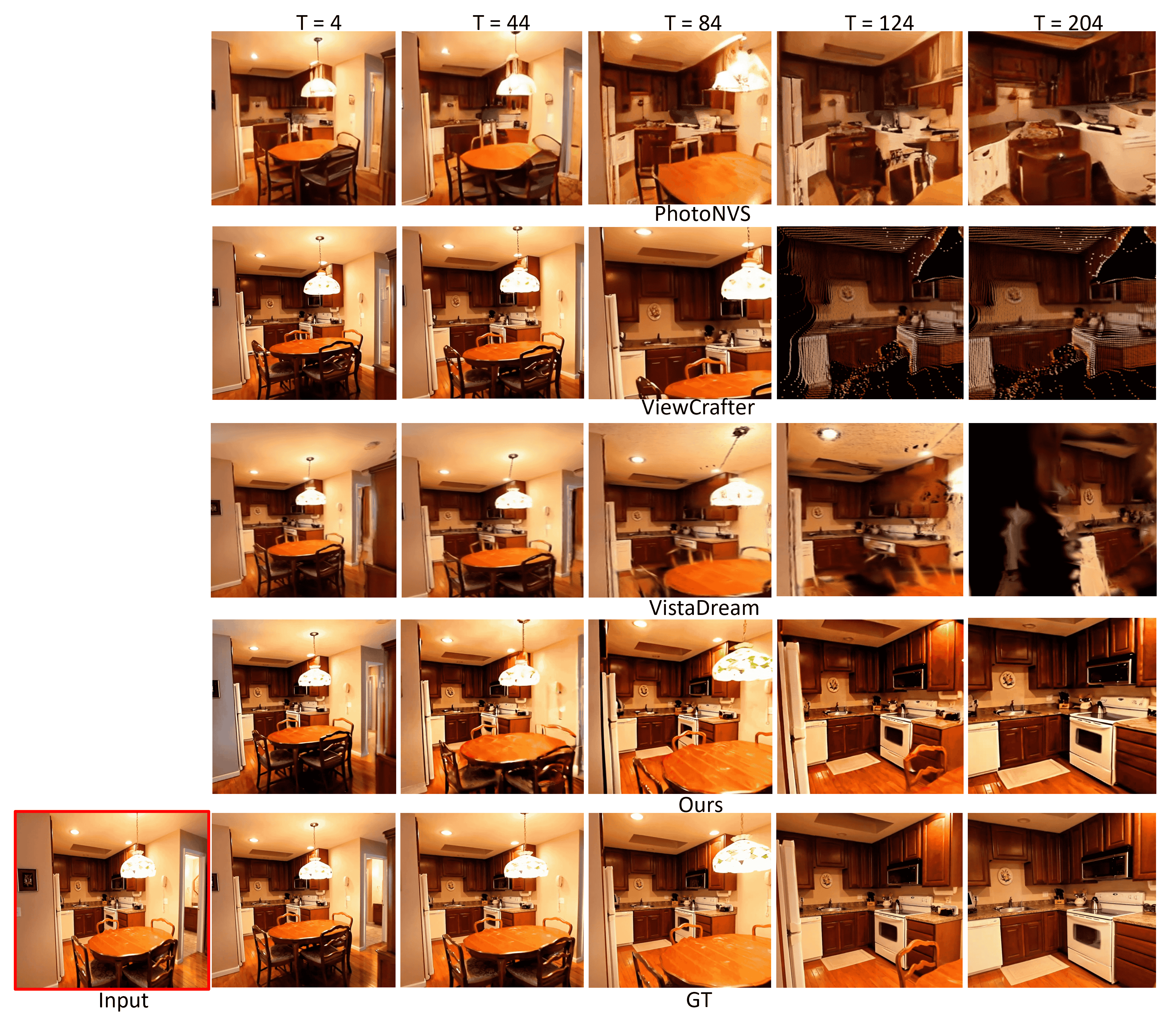}
    \vspace{-1.5em}
    \caption{Comparison of baseline models and our approach on the RealEstate10K~\cite{zhou2018stereo} dataset (each row shows a sequence of generated frames for a specific model corresponding to increasing timestamps). The input view image, highlighted in red, is shown in the bottom left. }
    \label{fig:supple-re1}
\Description{Qualitative realestate10k 1.}
    \vspace{-1.0em}
\end{figure*}

\begin{figure*}[!tbp]
    \centering
    \includegraphics[width=1.0\textwidth,  trim=6pt 4pt 25pt 0pt, clip]{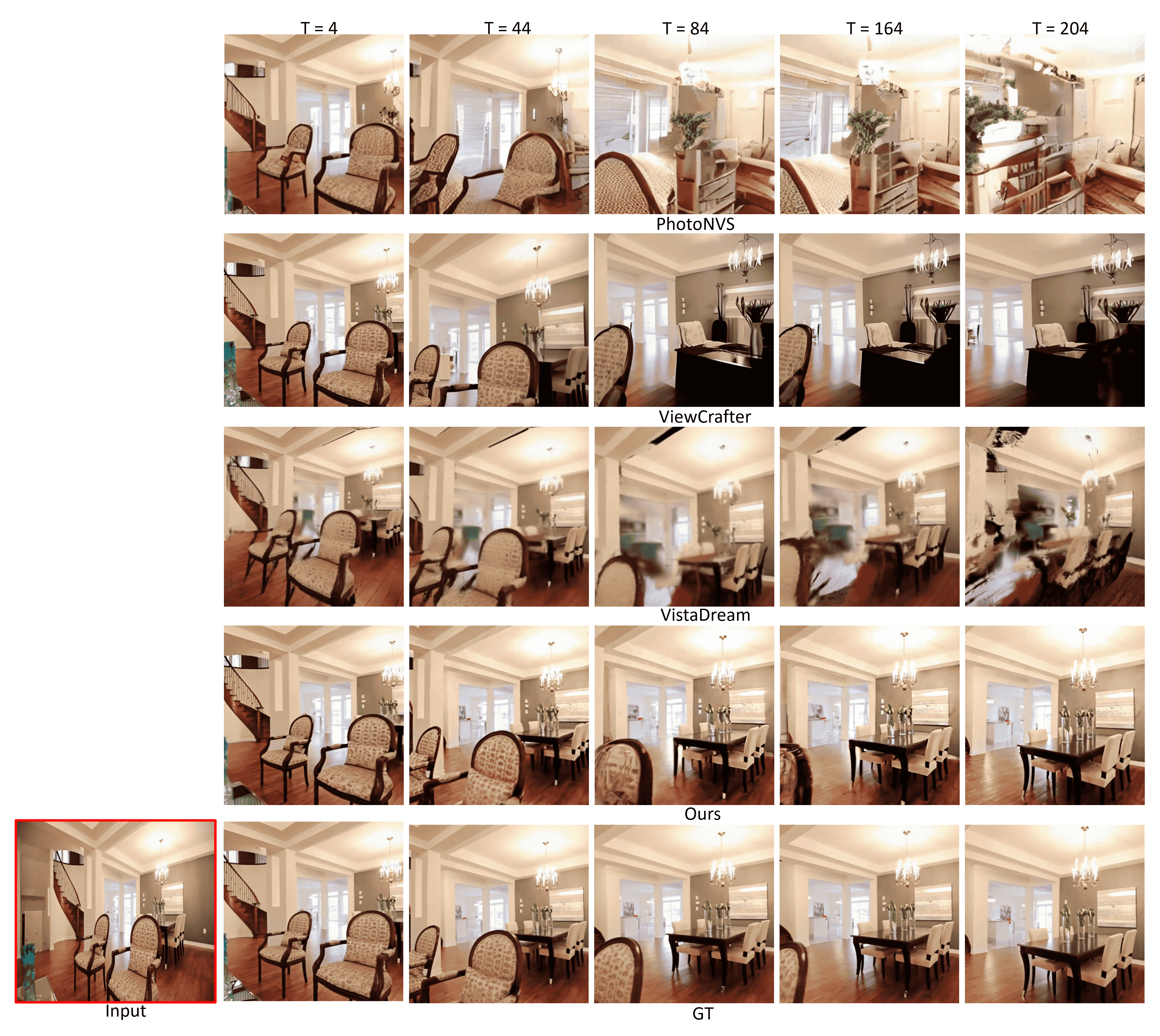}
    \caption{Comparison of baseline models and our approach on the RealEstate10K~\cite{zhou2018stereo} dataset (each row shows a sequence of generated frames for a specific model corresponding to increasing timestamps). The input view image, highlighted in red, is shown in the bottom left. }
    \label{fig:supple-re2}
\Description{Qualitative results of realestate10K 2.}
    \vspace{-1.0em}
\end{figure*}

% \begin{figure*}[!tbp]
%     \centering
%     \includegraphics[width=1.0\textwidth,  trim=6pt 4pt 25pt 0pt, clip]{figures/re-baseline5.png}
%     \vspace{-1.5em}
%     \caption{Comparison of baseline models and our approach on the RealEstate10K~\cite{zhou2018stereo} dataset (each row shows a sequence of generated frames for a specific model corresponding to increasing timestamps). The input view image, highlighted in red, is shown in the bottom left. We also present the outpainted panorama generated by our model from a single perspective view input, shown in the second-to-last row.  }
%     \label{fig:supple-re5}
%     \vspace{-1.0em}
% \end{figure*}

\begin{figure*}[!tbp]
    \centering
    \includegraphics[width=1.0\textwidth,  trim=6pt 4pt 25pt 0pt, clip]{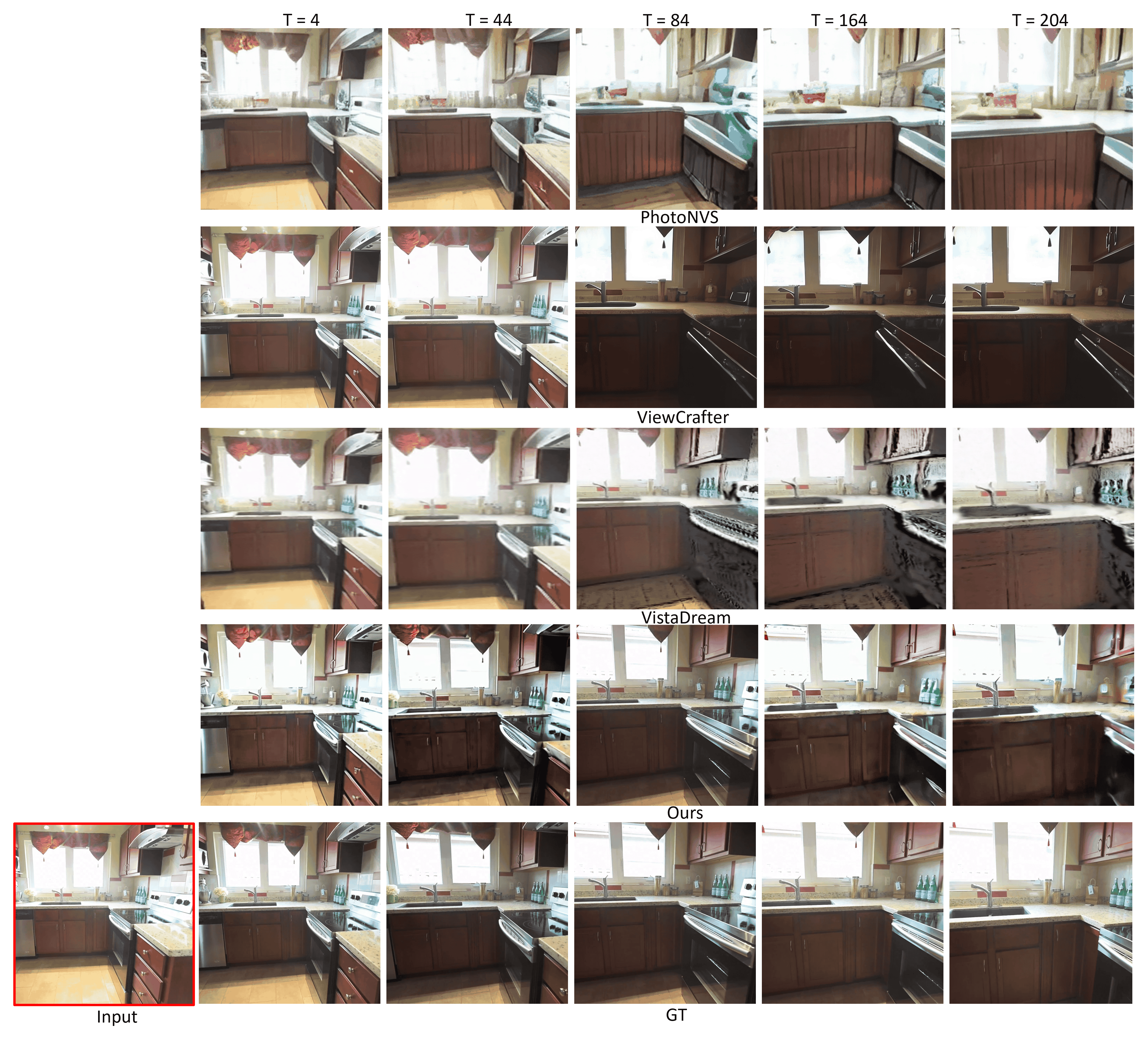}
    \vspace{-1.5em}
    \caption{Comparison of baseline models and our approach on the RealEstate10K~\cite{zhou2018stereo} dataset (each row shows a sequence of generated frames for a specific model corresponding to increasing timestamps). The input view image, highlighted in red, is shown in the bottom left. }
    \label{fig:supple-re3}
    \vspace{-1.0em}
\Description{Qualitative results of  Reaestate10K 3.}
\end{figure*}

% \begin{figure*}[!tbp]
%     \centering
%     \includegraphics[width=1.0\textwidth,  trim=6pt 4pt 25pt 0pt, clip]{figures/mp-baseline1.png}
%     \vspace{-1.5em}
%     \caption{Comparison of baseline models and our approach on the Matterport3D~\cite{Matterport3D} dataset (each row shows a sequence of generated frames for a specific model corresponding to increasing timestamps). The input view image, highlighted in red, is shown in the bottom left. We also present the outpainted panorama generated by our model from a single perspective view input, shown in the second-to-last row.}
%     \label{fig:supple-mp1}
%     \vspace{-1.0em}
% \end{figure*}

\begin{figure*}[!tbp]
    \centering
    \includegraphics[width=1.0\textwidth,  trim=6pt 4pt 25pt 0pt, clip]{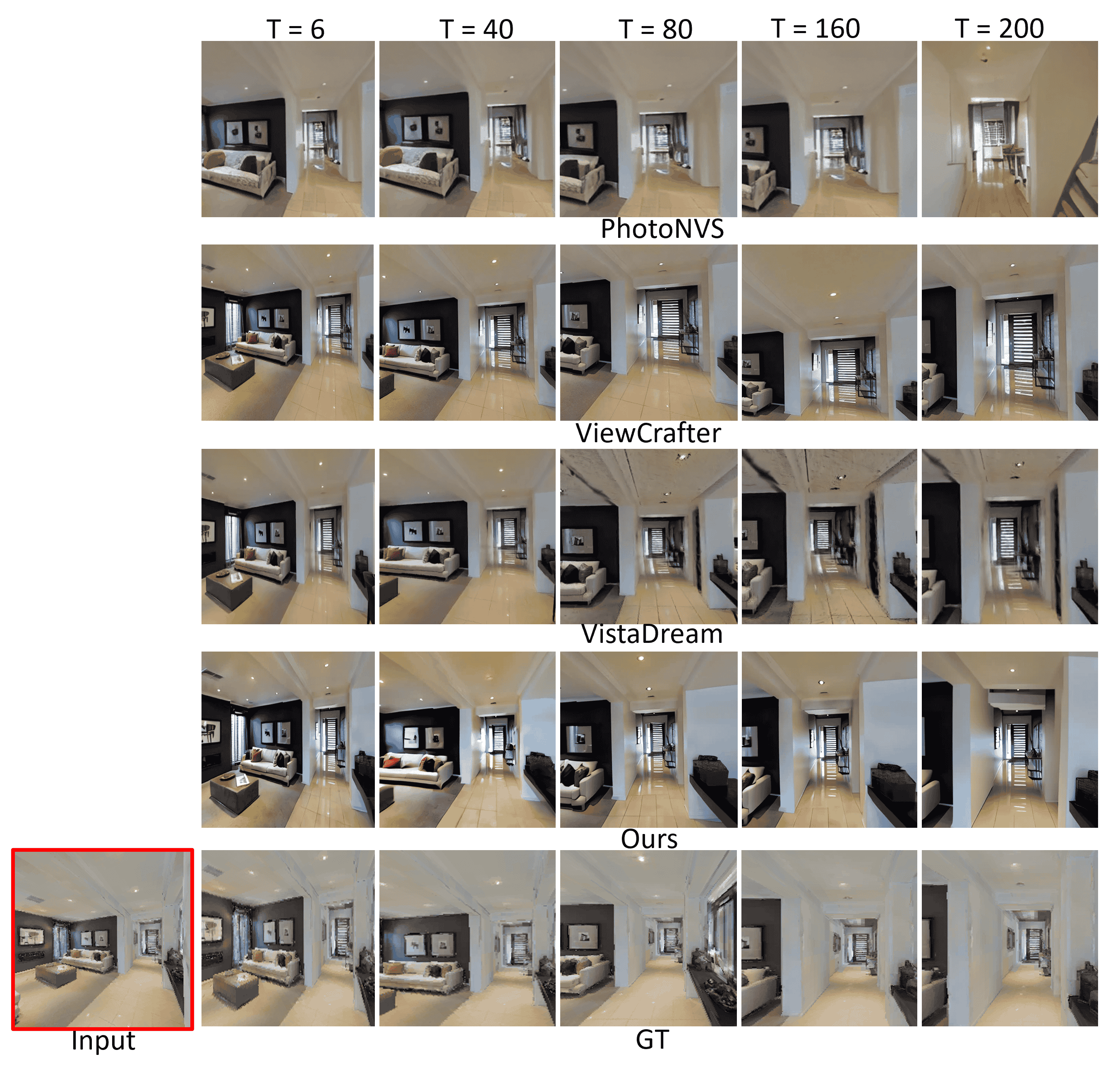}
    \vspace{-1.5em}
    \caption{Comparison of baseline models and our approach on the Matterport3D~\cite{Matterport3D} dataset (each row shows a sequence of generated frames for a specific model corresponding to increasing timestamps). The input view image, highlighted in red, is shown in the bottom left. }
    \label{fig:supple-mp1}
\Description{Qualitative results of Matterport3D 1.}
    \vspace{-1.0em}
\end{figure*}

\begin{figure*}[!tbp]
    \centering
    \includegraphics[width=1.0\textwidth,  trim=6pt 4pt 25pt 0pt, clip]{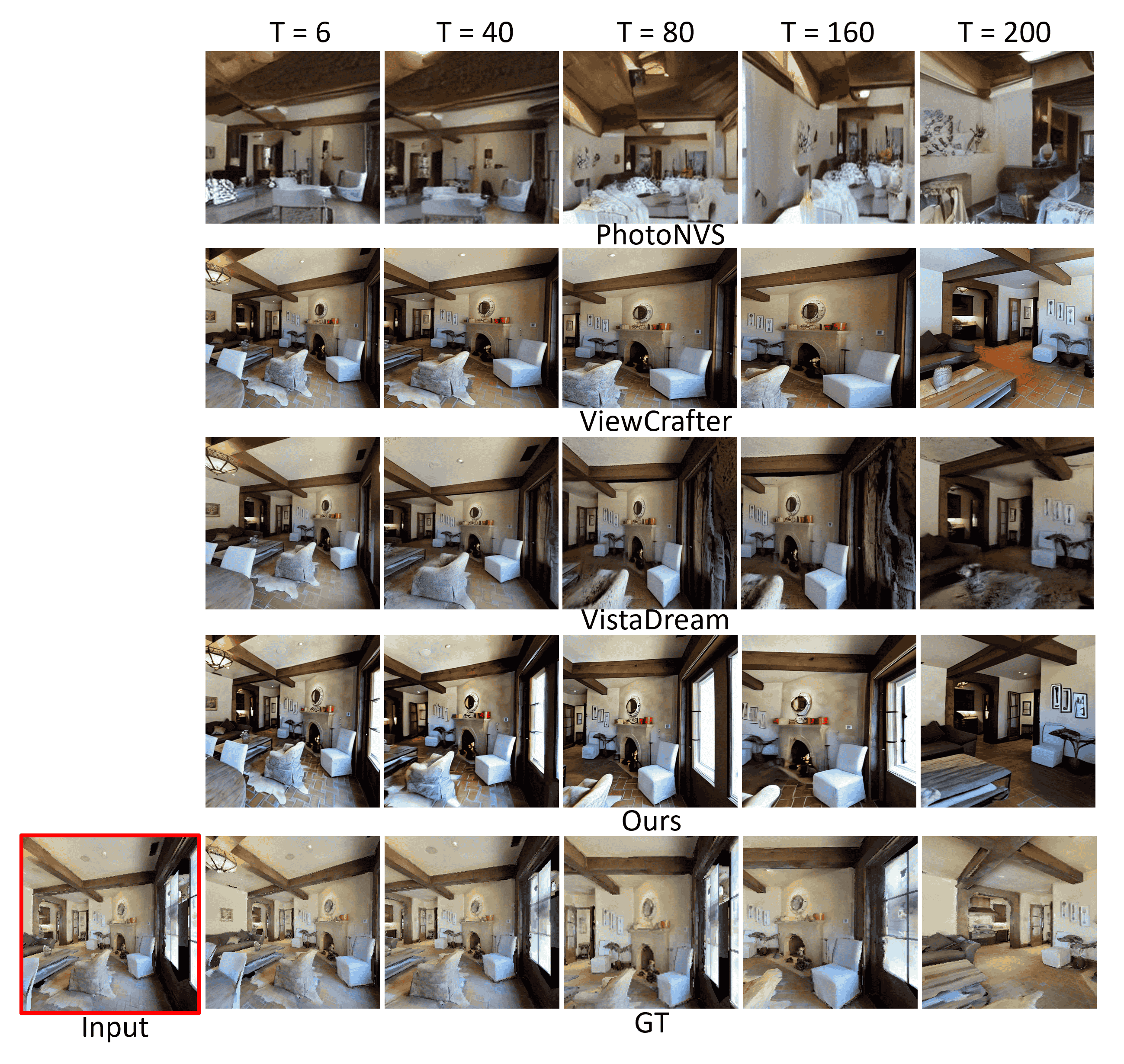}
    \vspace{-1.5em}
    \caption{Comparison of baseline models and our approach on the Matterport3D~\cite{Matterport3D} dataset (each row shows a sequence of generated frames for a specific model corresponding to increasing timestamps). The input view image, highlighted in red, is shown in the bottom left. }
    \label{fig:supple-mp2}
    \vspace{-1.0em}
\Description{Qualitative results of Matterport3D 2.}
\end{figure*}

\begin{figure*}[!tbp]
    \centering
    \includegraphics[width=1.0\textwidth,  trim=6pt 4pt 25pt 0pt, clip]{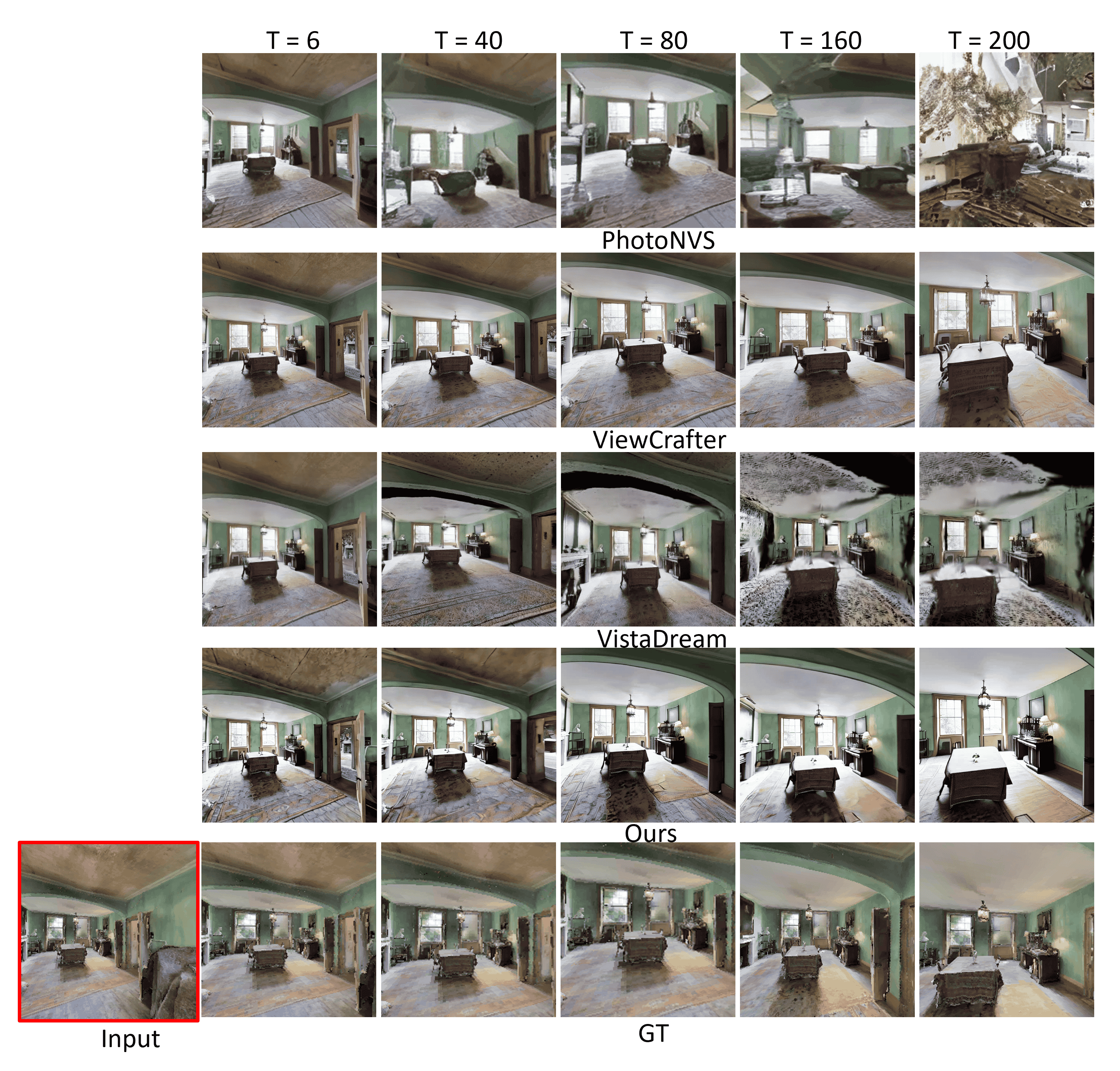}
    \caption{Comparison of baseline models and our approach on the Matterport3D~\cite{Matterport3D} dataset (each row shows a sequence of generated frames for a specific model corresponding to increasing timestamps). The input view image, highlighted in red, is shown in the bottom left. }
    \label{fig:supple-mp3}
\Description{Qualitative results of Matterport3D 3.}
    \vspace{-1.0em}
\end{figure*}

% \begin{figure*}[!tbp]
%     \centering
%     \includegraphics[width=1.0\textwidth,  trim=6pt 4pt 25pt 0pt, clip]{figures/mp-baseline5.png}
%     \vspace{-1.5em}
%     \caption{Comparison of baseline models and our approach on the Matterport3D~\cite{Matterport3D} dataset (each row shows a sequence of generated frames for a specific model corresponding to increasing timestamps). The input view image, highlighted in red, is shown in the bottom left. We also present the outpainted panorama generated by our model from a single perspective view input, shown in the second-to-last row.  }
%     \label{fig:supple-mp5}
%     \vspace{-1.0em}
% \end{figure*}
% \balance
% \bibliographystyle{ACM-Reference-Format}
% \bibliography{samples/kxy}

\end{document}